\documentclass[10pt,twocolumn,letterpaper]{article}

\usepackage{cvpr}              % To produce the CAMERA-READY version
\usepackage[dvipsnames]{xcolor}

\usepackage{times}
\usepackage{epsfig}
\usepackage{graphicx}
\usepackage{amsmath}
\usepackage{amssymb}
\usepackage{enumitem}
\usepackage[export]{adjustbox}
\usepackage{tikz}
\usepackage{bm}
\usepackage{bbm}
\usepackage{booktabs}

\definecolor{cvprblue}{rgb}{0.21,0.49,0.74}
\usepackage[pagebackref,breaklinks,colorlinks,citecolor=cvprblue]{hyperref}

\def\paperID{169} % *** Enter the Paper ID here
\def\confName{3DV\xspace}
\def\confYear{2024\xspace}

\title{A Local Appearance Model for Volumetric Capture of Diverse Hairstyles}

\author{
Ziyan Wang$^{1,2}$~~~~
Giljoo Nam$^{2}$~~~~
Aljaz Bozic$^{2}$~~~~
Chen Cao$^{2}$\\
Jason Saragih$^{2}$~~~~
Michael Zollhöfer$^{2}$~~~~
Jessica Hodgins$^{1}$
\vspace{0.1cm} \\ 
$^{1}$Carnegie Mellon University~~~
$^{2}$Reality Labs Research~~~
}

\begin{document}

% choose your color
 \newcommand{\ZW}[1]{{\color{blue}{\bf ZW: #1}}}
 
 \newcommand{\SL}[1]{{\color{olive}{\bf SL: #1}}}
 \newcommand{\GN}[1]{{\color{orange}{\bf GN: #1}}}
 
 \definecolor{tscolor}{RGB}{153,51,255}
 \newcommand{\TS}[1]{{\color{tscolor}{\bf TS: #1}}}

 \definecolor{mzcolor}{RGB}{255,50,00}
 \newcommand\MZ[1] {\textbf{\textcolor{mzcolor}{MZ: #1}}}

 \newcommand{\CR}[1]{#1}
 
 \newcommand{\CL}[1]{{\color{red}{\bf{CL: #1}}}}
 
 \newcommand{\JKH}[1]{{\color{blue}{\bf JKH: #1}}}
 
 \definecolor{cccolor}{RGB}{0,50,255}
 \newcommand\CC[1] {\textbf{\textcolor{cccolor}{CC: #1}}}

 % various shortcuts
 \newcommand{\bp}{\mathbf{p}}
 \newcommand{\bc}{\mathbf{c}}
 \newcommand{\bft}{\mathbf{f}}
 \newcommand{\bz}{\mathbf{z}}
 \newcommand{\bv}{\mathbf{v}}
 \newcommand{\br}{\mathbf{r}}
 \newcommand{\bV}{\mathbf{V}}
 \newcommand{\bR}{\mathbb{R}}

 \newcommand{\mL}{\mathcal{L}}
 \newcommand{\mR}{\mathcal{R}}
 \newcommand{\plh}{\mkern-1.5mu{\times}\mkern-2mu}
 
 \newcommand{\wt}[1]{\widetilde{#1}}
 
 \newcommand{\ttt}[1]{\small\texttt{#1}}

 \newcommand{\IGNORE}[1]{}

\definecolor{gold}{rgb}{1.0, 0.84, 0.0}
\newcommand{\gB}[1]{\fcolorbox{white}{gold}{#1}}
\definecolor{silver}{rgb}{0.75, 0.75, 0.75}
\newcommand{\sB}[1]{\fcolorbox{white}{silver}{#1}}
\definecolor{bronze}{rgb}{0.8, 0.5, 0.2}
\newcommand{\bB}[1]{\fcolorbox{white}{bronze}{#1}}

\maketitle

%%%%%%%%% ABSTRACT
\begin{abstract}

\iffalse
Hair, as an innegligible part of personal identity and appearance, is crucial for a high-quality, realistic avatar.
%
Prior works either focus on modeling the facial region only or uses personal specific model for avatar modeling, which limits their ability to generalize or capture at scale.
%
In this work, we present a novel method for creating high-fidelity avatars with good hair appearance at scale.
%
Our method takes the advantage of the local similarity between different hairstyles and learns an universal hair appearance prior at that level.
%
The universal hair appearance prior takes 3D aligned features as input and can generate dense radiance fields conditioned on a sparse colored point cloud.
%
We demonstrate in the experiment that our model is capable of capturing a diverse set of hairstyles and generalize to challenging new hairstyles.
%
We empirically show that our method outperforms prior works on capturing and generating photorealistic avatars with complete hair.
\fi

Hair plays a significant role in personal identity and appearance, making it an essential component of high-quality, photorealistic avatars. 
Existing approaches either focus on modeling the facial region only or rely on personalized models, limiting their generalizability and scalability. 
In this paper, we present a novel method for creating high-fidelity avatars with diverse hairstyles. 
Our method leverages the local similarity across different hairstyles and learns a universal hair appearance prior from multi-view captures of hundreds of people.
This prior model takes 3D-aligned features as input and generates dense radiance fields conditioned on a sparse point cloud with color. 
As our model splits different hairstyles into local primitives and builds prior at that level, it is capable of handling various hair topologies.
Through experiments, we demonstrate that our model captures a diverse range of hairstyles and generalizes well to challenging new hairstyles. 
Empirical results show that our method improves the state-of-the-art approaches in capturing and generating photorealistic, personalized avatars with complete hair.

\end{abstract}

%%%%%%%%% BODY TEXT
\section{Introduction}
Creating and capturing high-fidelity 3D human avatars is an essential capability for mixed reality. 
A high-quality, photorealistic 3D avatar can blur the boundary between the real and virtual world and facilitate VR/AR applications such as social telepresence, virtual gaming and virtual shopping.
%
%Such a photorealistic avatar is supposed to faithfully reflect every detail of a human like facial expression, hairstyles and body movements.
%
One critical aspect of achieving a lifelike avatar is accurately capturing and modeling hair, as it plays a vital role in establishing personal identity and achieving personalized avatars.
%
%In this paper, we focus on capturing and modeling hair, an integral component for establishing personal identity as well as for personalized avatars.
%
%As an innegligible part of establishing personal identity, decent quality hair is essential for both the authenticity and aesthetic of a personalized avatar.
%
However, hair is potentially challenging to capture due to its complex geometry and high-frequency texture.
Furthermore, different hairstyles exhibit large intra-class variance in terms of appearance and shape, which adds to the complexity of efficiently creating personalized avatars for a large group of individuals.%learning a generalizable model for capturing diverse hairstyles at once is even more challenging as there is a large intra-class variance in hair appearance and shape, which requires careful model design as well as a large amount of training data.

In this work, we look at the problem of how to capture diverse hair appearance for efficient and accurate creation of photorealistic, personalized 3D avatars.
%
%Towards that goal of efficient and accurate creation of personalized avatars, we present a novel generalizable volumetric model to generate high-quality hairstyles conditioned on a sparse point cloud.
%
Given the complexity of hair geometry, several 3D representations have been explored with the goal of improving modeling accuracy.
%
%There are many works on capturing a mesh-based 3D avatar~\cite{steve_meshvae, tewari2017mofa, tran2018nonlinear3dmm, li2017flame, xiang2020monoclothcap, saito2021scanimate, bagautdinov2021driving}. 
%
%Those mesh-based representations mostly capture human parts like skin and cloth well with details restored while using a low memory footprint.
%
%Although quite efficient to store and render, they are mostly limited to modeling the non-volumetric part of a human, thus not suitable for representing many hairstyles.
%
Mesh-based representations work well for capturing the surface details and are most efficient to store.
But they are not well-suited for modeling hairstyles that exhibit volumetric properties.
Strand-based representations can capture hair with the highest accuracy and are easy to manipulate.
However, modeling and rendering complex hairstyles with strand-based representations can be computationally expensive.
%
%More recently, many work~\cite{park2021nerfies, Gafni_2021_CVPR, Zheng_2022_CVPR, Hong_2022_CVPR, Grassal_2022_CVPR, Kornilova_2022_CVPR, kania2022conerf, park2021hypernerf, zheng2023pointavatar, wang2021hvh} have achieve impressive results on capturing photorealistic 3D avatars with a volumetric like representation.
%
%Those volumetric-based methods showcased their flexibility in modeling the diversity of human avatars with hairstyles in various geometry and appearance, and also generate photorealistic rendering results.
%
%In comparison, volumetric-based methods are flexible in modeling the diversity of human avatars with hairstyles in various geometry and appearance.
%
%They use tailored geometry and appearance modeling for hair which is biased to a person-specific configuration.
%, which makes it non-trivial to extend them for capturing haired avatars with varied topology and texture.
%
In comparison, many recent volumetric-based methods~\cite{park2021nerfies, Gafni_2021_CVPR, Zheng_2022_CVPR, Hong_2022_CVPR, Grassal_2022_CVPR, Kornilova_2022_CVPR, kania2022conerf, park2021hypernerf, zheng2023pointavatar, wang2021hvh} modeled human avatars with photorealistic hairstyles with diverse geometry and appearance.
The avatars can be optimized from images or videos via differentiable raymarching in an end-to-end learning framework.
%, and also generate photorealistic appearance of hair.
%
However, the model accuracy relies on a person-specific model which does not generalize and requires extensive training time.
Those properties prohibit these approaches from creating personalized avatars for a large group of individuals.
%
%Some recent approaches~\cite{Raj_2021_CVPR, keypoint_nerf} create a photorealistic volumetric avatar from a few snapshots efficiently.
%
%They build generalizable volumetric avatars by densely conditioning their model on pixel-aligned information, inspired by recent literature on generalizable NeRF~\cite{yu2021pixelnerf, mvsnerf}.
%
%However, they are usually limited to small camera baseline settings or might not generalize well to complex geometry and detailed parts of a human avatar.
%
%Cao \textit{et al.}~\cite{ica_chen} achieves better generalization on personalized geometry and appearance by stacking pixel-aligned features on a tracked head mesh in various scales, resulting in better 3D awareness of the generated avatar.
%
%However, due to the constrained topology and limited resolution of a head mesh, long hair or very complex hairstyles might not be properly generalized to.

%To circumvent those challenges for the efficient and accurate creation of personalized avatars
To achieve both efficiency and accuracy in creating personalized avatars for individuals, we present a universal 3D hair appearance model that captures diverse hairstyles with high fidelity and helps to generate high-quality hair appearance for personalized avatar creation.
%
%Our universal 3D hair appearance model is conditioned on 3D point-level hair features that are anchored by a group of volumetric primitives and regress the dense radiance field stored in each volumetric primitive in parallel with a UNet.
%
Our universal 3D hair appearance model is conditioned on a group of hair feature volumes that are diffused from a 3D hair point cloud with color.
Those hair feature volumes are anchored by volumetric primitives that tightly bound the hair point cloud, and reflect the local structures and appearance at a primitive level.
To amplify the sparse point cloud for dense appearance modeling, we learn a UNet to transfer those hair feature volumes into dense radiance fields.
By spatially compositing those volumetric primitives, we get a set of volumetric radiance fields that fully cover different parts of the hair.
Those volumetric radiance fields are not restricted to certain topologies and can model different hairstyles.
The design choice to learn an appearance prior at a primitive level rather than the hairstyle level is based on the observation that different hairstyles share a more similar pattern within a local region than at a global scale.
%
%As hairstyles vary a lot both in geometry and appearance, it is hard to manually align different hairstyles into one template.
%
%As hairstyles vary a lot both in geometry and appearance, l
Learning a compact embedding space at a global scale to express such variety is nearly implausible, with a limited amount of training data.
Learning a local prior model helps us to achieve better generalization as it splits each hairstyle into multiple local volumetric primitives.
%However, different hairstyles can share similar geometry or appearance within a local region.
%
%For example, long straight hair and short straight hair might differ in their global shape.
%
%But they might exhibit similar geometric patterns within certain local regions.
%
%Based on this observation, we design a model that is aware of local hair structures and we hope such a model can learn a meaningful prior of local hair patterns for in-the-wild hair capture with a RGB-D scanner.
%However, capturing a human head with hair still remains challenging, due to its complex geometry and appearance.
%
%And capturing various hairstyles at scale becomes even harder as there is large intra-class variance in hair texture and shape. 
%

We conduct extensive experiments on multiview captures of multiple identities as well as in-the-wild captures of new identities with a limited number of viewpoints.
We empirically show that our method outperforms previous state-of-the-art methods~\cite{keypoint_nerf,ica_chen} in capturing diverse hairstyles, demonstrating improved quality and generalizing to new hairstyles for personalized avatar creation.
Given sparse views as input, we also find that our local prior model serves as a good initialization for more efficient acquisition of personalized avatars with less finetuning.
In summary, our contributions are
\begin{itemize}%[leftmargin=*, noitemsep]
    \item We present a novel volumetric feature representation based on a point cloud with color and a local appearance model that is generalizable to various complex hairstyles.
    \item We empirically show that our method outperforms previous state-of-the-art approaches in capturing high-fidelity avatars with diverse hairstyles and generating photorealistic appearances for novel identities with challenging hairstyles.
    Our method also enables the efficient capture of personalized avatars from an iPhone captures.
\end{itemize}
\section{Related Works}

In this section, we will discuss related literature on hair modeling, neural radiance fields and volumetric avatars.

\subsection{Hair Strand Reconstruction}

Multiview stereo has been heavily explored in reconstructing hair strand representation from visual capture systems.
Early works~\cite{paris2008hair_photobooth, paris2004capture, wei2005modeling} utilize 2D orientation maps as additional information for doing triangulation on 3D strand structures.
Luo \textit{et al.}~\cite{luo2012multi, luo2013wide} and Hu \textit{et al.}~\cite{hu2014robust} fit structural hair representations like strands and ribbons into point cloud from MVS for hair reconstruction.
Nam \textit{et al.}~\cite{nam2019lmvs} relax the plane assumption in PatchMatch MVS for thin strand reconstruction and Sun \textit{et al.}~\cite{sun2021hairinverse} leverages patterned illuminations for high-fidelity strand geometry as well as reflectance material recovery.
NeuralStrands~\cite{rosu2022neural} learns a neural hair generator for hair structure fitting from the reconstruction.
NeuralHDHair~\cite{wu2022neuralhdhair} and DeepMVSHair~\cite{kuang2022deepmvshair} learn to regress a hair growing field for hair strand generation based on sparse multiview images of hair.
However, those methods require a dense camera array and long optimization time which is prohibitive for creating avatars to individuals.
Compared to the hair capture methods using a multiview capture system, many works~\cite{chai2012single, chai2016autohair, hu2015single, liang2018video, saito20183d, yang2019dynamic, zhang2018hair, zhou2018hairnet, zheng2023hairstep} explore hair reconstruction from a single view which is easier to scale up.
They take advantage of the existing database of synthetic 3D hair or user interactions to serve as a prior for inferring the underlying hair structure given partial observation like a monocular video or image.
However, those methods usually generate overly smooth results and are limited in resolution.
Furthermore, they are also limited by the variety of the existing synthetic hair datasets, which hampers their models' ability to generate complex hairstyles like hair bundles or extremely curly hair.

\subsection{Neural Radiance Fields}
Neural radiance fields (NeRF)~\cite{mildenhall2020nerf} learn a radiance field from multiple calibrated RGB images via differentiable volumetric raymarching, where they parameterize the radiance field implicitly with MLPs.
As a result of its simplicity and impressive results, many recent works have improved NeRF on dimensions such as accurate geometry~\cite{wang2021neus, yariv2021volume, oechsle2021unisurf} and anti-aliasing~\cite{barron2021mip, barron2022mip, barron2023zip}.
However, most of these approaches suffer from long training times and are not very efficient to render.

Many works~\cite {liu2020nsvf, yu2021plenoctrees, kilonerf, lindell2021autoint, mueller2022instant} have focused on improving the rendering efficiency of NeRF. 
AutoInt~\cite{lindell2021autoint} speeds up the volumetric raymarching process by learning closed-form solutions to the integral in volumetric raymarching. 
NSVF~\cite{liu2020nsvf} and PlenOctrees~\cite{yu2021plenoctrees} both optimize the run-time efficiency of NeRF by stacking it in a sparse structure like an octree. 
KiloNeRF~\cite{kilonerf} optimizes the run-time efficiency by shortening the model inference time by substituting a single MLP with multiple tiny and shallow MLPs.
TensRF~\cite{chen2022tensorf} factorizes the 4D tensor of radiance field into compact components like vector and matrix which leads to not only smaller model size but also more efficient reconstruction of the scene.
Instant-NGP~\cite{mueller2022instant} combines both tiny MLP and sparse structures.
Furthermore, to achieve efficient storage of the radiance field without losing the reconstruction fidelity, a spatial hash function is presented to store the multi-resolution spatial features which will be used as input to the tiny MLP to retrieve the radiance value.
Our method also leverages sparse structures for efficient rendering of radiance fields.
%However, the octree structure is optimized specifically for a given static scene, making animation or control of the scene not trivial. 

\subsection{Volumetric Avatar}
Most recently, volumetric representation is favored in avatar creation due to its completeness in modeling various hair geometry and its simplicity in optimization via a differentiable volumetric raymarching.
Neural Volumes~\cite{steve_nvs} first presents a compact formulation for differentiable volumetric raymarching for creating a volumetric head avatar with photorealistic hair from multiview RGB videos. 
%
%One of the strength of this work is that it could directly optimize a volume grid from multiview images and could still have good quality on semi-transparent objects like hair. 
One of the follow-up works~\cite {wang2021learning} combines the volumetric representation and coordinate-based representation into a hybrid form for better rendering quality and drivability. 
With the recent success of NeRF~\cite{mildenhall2020nerf} in modeling 3D scenes with good appearance from multiple images, there are many works for building avatar with NeRF~\cite{Raj_2021_CVPR, park2021nerfies, Gafni_2021_CVPR, Zheng_2022_CVPR, Hong_2022_CVPR, Grassal_2022_CVPR, Kornilova_2022_CVPR, kania2022conerf, keypoint_nerf, park2021hypernerf, zheng2023pointavatar}. 
Those NeRF-based avatars are typically composed of a spatial 3D warp field and a canonical appearance field. 
Nerfies~\cite{park2021nerfies} learns a volumetric deformation field and canonical space NeRF for modeling dynamics-related changes. 
HyperNeRF~\cite{park2021hypernerf} uplifts the deformation field from 3D Euclidean space to a high dimensional hyper semantic space to better model large variations in expressions. 
NeRFace~\cite{Gafni_2021_CVPR} improves the controllability of NeRF-based avatars by using a 3D face morphable model to control the radiance field defined on faces. 
IM Avatar~\cite{Zheng_2022_CVPR} improves NeRFace for more complete expressions based on implicit skinning fields following FLAME~\cite{li2017flame}. 
HeadNeRF~\cite{Hong_2022_CVPR} learns a parametric head model with illumination using NeRF. 
PointAvatar~\cite{zheng2023pointavatar} learns a point cloud-based avatar with a temporally conditioned volumetric deformation field for capturing a 3D avatar from video.
INSTA~\cite{zielonka2023instant} builds an efficient pipeline for learning a NeRF-based avatar based on instant-ngp~\cite{mueller2022instant}.
However, those methods mostly assume hair to be rigidly attached to the head without motion. And most of them are still limited by the prohibiting rendering time which is also the limitation of NeRF. 

In contrast to the line of NeRF-based avatars, a mixture of volumetric primitives (MVP)~\cite{steve_mvp} builds a volumetric representation that can generate extremely high-quality and real-time renderings that look realistic even on challenging materials, like hair and clothing. 
The key idea is to model a dynamic head by stacking multiple volumetric primitives only on a tracked head mesh, without wasting memory on empty spaces. 
Following MVP, HVH~\cite{wang2021hvh} models hair and head in separate layers and present a hybrid model of guide hair stands and volumetric primitives. 
A structure-aware strand tracking algorithm and a 3D scene flow optimization method are presented for dynamic capture of hair.
%based on per-frame 3D flow optimization is presented to track hair at strand level. 
%
%To refine the tracking results and enforce temporal smoothness, a fine-level volumetric raymarching algorithm on dense 3D volumetric scene flow is presented. 
%
NeuWigs~\cite{wang2023neuwigs} further improves the tracking robustness and builds a data-driven model for hair animation.
However, similar to the line of NeRF-based avatars, those methods build a person-specific model which does not generalize to novel identities and it is non-trivial to reuse those methods for large-scale capture of diverse avatars due to the large variance in hair topology and appearance.

Some recent works~\cite{Raj_2021_CVPR, keypoint_nerf, ica_chen} looked at the problem of building a generalizable model for scalable avatar creation.
Pixel-Aligned Avatars~\cite{Raj_2021_CVPR} utilizes pixel-aligned information as additional inputs for NeRF to extend its drivability and generalization over sequence data. 
Inspired by PixelNeRF~\cite{yu2021pixelnerf} and MVSNeRF~\cite{mvsnerf}, KeypointNeRF~\cite{keypoint_nerf} improves the generalization of avatars and NeRF's robustness to sparse views by using a new spatial encoding technique with sparse 3D key points. 
However, they are usually limited to modeling very complex geometry and are restricted to small camera baseline setups.
Cao \textit{et al.}~\cite{ica_chen} extends MVP to in-the-wild scenarios and unseen faces. 
By learning a cross-identity hyper network that controls the expression and identity change on volumetric avatars from a large corpus of data, the model can be easily adapted to newly captured identities even from an iPhone scan. 
However, it relies on an accurately tracked head mesh for recovering the volumetric texture on top of it and is limited to model hairstyles that can not be approximated by a spherical mesh surface.

\begin{figure*}[tb!]
    \centering
    \includegraphics[width=\textwidth]{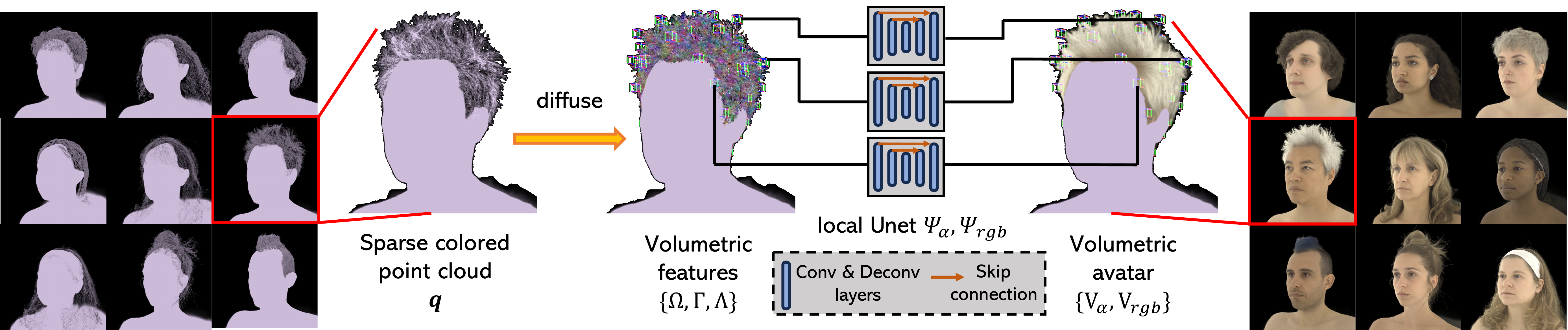}
    \caption{\label{prop_eval:fig_nvs_iph}\textbf{Pipeline of Our Method.} We present a pipeline to achieve large scale capture of diverse hairstyles for avatar creation. The core of our pipeline is a local UNet that can generate local appearance field conditioned on colored point cloud $\bm{q}$. Our method is robust to various challenging hairstyles and can generate photorealistic appearance of those hairstyles.}
\end{figure*}
\vspace{-0.03cm}
\section{Method}
Our goal is to achieve efficient and accurate hair appearance capture of a large number of individuals with a single model.
We use a compositional volumetric representation for hair modeling.
There are several benefits of using such a representation for hair modeling.
First, volumetric representation is flexible enough to model complex hair configuration while also yielding a decent rendering quality with high fidelity on detailed geometry.
Second, rendering a compositional volumetric representation is more efficient than a volumetric representation, as the raymarching process is guided by a sparse structure to skip empty spaces.
To accommodate the diverse topology as well as capture the generalizable prior of various hairstyles, we learn a local hair appearance prior model $\Psi_\alpha$ and $\Psi_{rgb}$ that can regress the compositional volumetric presentation based on sparse color point clouds $\bm{q}$. 
The model takes input as a group of local hair feature volumes diffused by the colored hair point cloud, which is agnostic to the ordering of the point cloud and outputs a compositional volumetric representation $\mathbf{V}_{rgb} + \mathbf{V}_\alpha$ that can be rendered from different viewpoints.

\subsection{Preliminaries: Volumetric Rendering}~\label{prop_work:sec:volrender}
We render and optimize our compositional volumetric representation with the differentiable volumetric raymarching algorithm in MVP~\cite{steve_mvp}.
%
%Given all volume rotations $\bm{R}_t^{all}=[\bm{R}_t^{x,n}, \bm{R}_t^{p,n}]$, translations $\bm{d}_t^{all}=[\bm{d}_t^{x,n}, \bm{d}_t^{p,n}]$, scales $\bm{s}_t^{all}=[\bm{s}_t^{x, n}, \bm{s}_t^{p, n}]$ and local radiance fields $\bm{V}_t^{all}=[\bm{U}_t, \bm{V}_{t}]$, we can render them into image $\mathcal{I}_{cam_i}$ and compare it with $I_{cam_i}$ to optimize all volumes.
%
%Using an optimized BVH implementation similar to MVP~\cite{steve_mvp}, we can efficiently determine how each ray intersects with each volume.
%
Given the camera center as $\bm{c}$ and a ray direction $\bm{v}(p)$ associated with pixel $p$, we can define a ray function $\bm{r}$ as follow:
\begin{align*}
    \bm{r}(p)=\bm{c}+\tau\bm{v}(p),
\end{align*}
\noindent where $\tau$ is the traversal depth along the ray and $\bm{v}(p)$ is the direction starting from camera center $\bm{c}$ and pointing to pixel $p$.

To render the compositional volumetric representation, we aggregate all the alpha, the RGB values and the semantic labels from the volumetric field.
The image formation process can be formulated as:
\begin{align*}
    \mathcal{I}_p &= \int_{\tau_{min}}^{\tau_{max}} \mathbf{V}_{rgb}(\mathbf{r}_{p}(\tau))\frac{dT(\tau)}{d\tau}d\tau, \\
    \mathcal{M}_p &= \int_{\tau_{min}}^{\tau_{max}} \mathbf{V}_{label}(\mathbf{r}_p(\tau))\frac{dT(\tau)}{d\tau}d\tau, \\
    T(l) &= min(\int_{\tau_{min}}^{\tau_{max}} \mathbf{V}_{\alpha}(\mathbf{r}_{p}(\tau))d\tau , 1),
\end{align*}

%as $\bm{r}(\mathfrak{p}, l)=\bm{o}(\mathfrak{p})+l\bm{v}(\mathfrak{p})$ shooting from pixel $\mathfrak{p}$ in direction of $\bm{v}(\mathfrak{p})$ with a depth $l$ in range of $(l_{min}, l_{max})$.
%
%The differentiable formation of an image given the volumes can then be formulated as below:
%
\noindent where we composite the RGB and semantic label from near to far in a weighted sum manner.
$\mathbf{V}_{\clubsuit}(\cdot)$ indicates the volumetric field function that outputs the function value of a specific spatial point, where $\clubsuit$ can be alpha, RGB, or a semantic label.
For efficient rendering of $\mathbf{V}_\clubsuit$, we use a BVH to speed up the process of finding intersections between rays and volumetric primitives following MVP~\cite{steve_mvp}.

To get the full rendering, we composite the rendered image as $\mathcal{\tilde{I}}_p=\mathcal{I}_\mathfrak{p} + (1-\mathcal{A}_p)I_{p,bg}$ where $\mathcal{A}_p=T(l_{max})$ and $I_{p,bg}$ is the background image.
Similar to NeuWigs~\cite{wang2023neuwigs}, we model the hair and head of an avatar in separate layers and render their segmentation map as $\mathcal{M}_p$.
Please refer to the supplemental materials for implementation details.

\subsection{Local Hair Appearance Prior Model}\label{prop_work:sec:method}
In this part, we describe how we achieve the conditional generation of $\mathbf{V}_{alpha}$ and $\mathbf{V}_{rgb}$ based on sparse colored hair point cloud $\bm{q}$ with a local appearance prior model. 
We will first describe how we create the input to the local appearance model with a colored hair point cloud.
%Our design for the local hair appearance prior model that regresses the compositional volumetric representation. 
%
Then we will introduce our design for the architecture of the local appearance prior model as well as objectives for training the model.

\noindent\textbf{Hair feature volumes.}
As different hairstyles might be in different topologies and different sizes, it is not practical to learn a model that has a fixed output size to regress the appearance field of both long hair and short hair in the 3D space directly.
To achieve modeling across different hairstyles and capture the common appearance prior among them, we learn a local appearance prior model to regress the radiance field for each of those hair volumes in separate runs.
Given a colored hair point cloud $\bm{q}$ defined in a head-centered coordinate system, we prepare the input to the local appearance prior model by first partitioning it into a group of hair feature volumes.
Specifically, we first perform furthest point sampling~\cite{qi2017pointnet} on $\bm{q}$ to get $k$ centroids $\{\rho_i|i=1,2,\cdots,k\}$ that roughly span over the point cloud manifold uniformly.
Then for each centroid $\rho_i$, we diffuse the points into a volume grid $\Omega^{\rho_i}$ centered at $\rho_i$.
The volume grid is axis-aligned with the head-centered coordinate system with a length of $\delta$ and a grid resolution of $m$. 
Each vertex in the grid $\Omega^{\rho_i}$ stores both the spatial occupancy and RGB values, where the occupancy is 0 if no point is found within the radius of $\sqrt{3}\delta/2m$ around that vertex otherwise 1.
The RGB value will be the mean of the colors from all found points and $\bm{0}$ if no point is found.
In addition to the occupancy and RGB value, we aggregate the spatial coordinate of each vertex in the head-centered coordinates as well as the per-vertex viewing direction into volume grids $\Gamma^{\rho_i}$ and $\Lambda^{\rho_i}$ respectively.
Given viewing direction camera center $\bm{c}$ under the head centered coordinate, we calculate $\Lambda^{\rho_i} = norm(\Gamma^{\rho_i} - \bm{c})$ where we take the normalized vector between each point in $\Lambda^{\rho_i}$ and $\bm{c}$ as the per-vertex viewing direction.

\noindent\textbf{Local appearance UNet.} We learn two separate UNets~\cite{ronneberger2015unet} with skip connections that takes volume grid $\Omega^\rho_i$ as input and outputs the corresponding dense radiance field $\nu^{\rho_i}_{\alpha}$ and $\nu^{\rho_i}_{rgb}$ respectively as follows:

\begin{align*}
    \nu^{\rho_i}_{\alpha} &= \Psi_{\alpha}(\Omega^\rho_i, \Gamma^{\rho_i}|\bm\theta_{\alpha}) \\
    \nu^{\rho_i}_{rgb} &= \Psi_{rgb}(\Omega^\rho_i, \Gamma^{\rho_i}, \bm{\hat{c}}, \Lambda^{\rho_i}|\bm\theta_{rgb}),
\end{align*}

\noindent where $\bm\theta_{\alpha}$ and $\bm\theta_{rgb}$ are the learnable parameters for each network.
By spatially compositing the volumetric primitives $\{\nu^{\rho_i}_{\alpha}, \nu^{\rho_i}_{rgb}|i=1,2,\cdots,k\}$ with respect to their centroids $\{\rho_i|i=1,2,\cdots,k\}$, we get $\mathbf{V}_{alpha}$ and $\mathbf{V}_{rgb}$.
The UNet $\Psi_{\alpha}(\cdot)$ that regresses the alpha field of hair takes the occupancy and RGB field $\Omega^{\rho_i}$ as well as the grid coordinate $\Gamma^{\rho_i}$ as input.
To model the view conditioned appearance of hair, we learn a separate UNet $\Psi_{rgb}(\cdot)$ to regress the RGB field $\nu^{\rho_i}_{rgb}$ that takes additional input of the per-vertex viewing direction $\Lambda^{\rho_i}$ as well as the normalized camera center $\bm{\hat{c}}$ as viewing direction.
$\Lambda^{\rho_i}$ is served as additional information to the input of the UNet $\Psi_{rgb}(\cdot)$ and $\bm{\hat{c}}$ is injected at the bottleneck level where it is repeated and appended to every hair features at the coarsest resolution map.
We find that the usage of $\Gamma^{\rho_i}$ and $\Lambda^{\rho_i}$ improves the model's convergence by a large margin, which will be discussed in detail in the experiment section.
The encoder part of each UNet consists of convolutional layers with a kernel size of $3\times3$ with stride $1$ and $2$ to extract the features of $\Omega^\rho_i$ at different scales. 
The decoder part of each UNet consists of convolutional layers with a kernel size of $3\times3$ with stride $1$ and deconvolutional layers with a kernel size of $4\times4$ with stride $2$.
At each scale, we use a $1\times1$ convolutional layer as skip connections to add early conditions from the encoder features to the decoder features respectively.
All layers are followed with a LeakyReLU layer as activation.
We find that the usage of skip connections can greatly help the network to capture detailed geometry and more salient textures on the regressed hair radiance field.

\noindent\textbf{Training objectives and details.}
To learn the parameters $\bm\theta_{\alpha}$ and $\bm\theta_{rgb}$ of the local appearance model $\Psi_{\alpha}(\cdot)$ and $\Psi_{rgb}(\cdot)$, we construct image level reconstruction losses.
We formulate the training objective $\mathcal{L}$ as below:
\begin{align*}
    \mathcal{L} = \mathcal{L}_1 + \lambda_{VGG}\mathcal{L}_{VGG} + \lambda_{seg}\mathcal{L}_{seg},
\end{align*}
\noindent where $\lambda_{VGG}$ and $\lambda_{seg}$ are positive values for rebalancing each term in the training objectives.
The first term $\mathcal{L}_1$ measures the difference between the rendered image $\mathcal{\tilde{I}}$ and the ground truth image $I_{gt}$.
The second term is a perceptual loss between the rendered image $\mathcal{\tilde{I}}$ and the ground truth image $I_{gt}$, which aims at enhancing the visual quality and adding high frequency details of the rendered image.
The third term is a segmentation loss for better disentangling hair and non-hair regions.
Please refer to the supplimental materials for more training related details.

\iffalse
%
The first term $\mathcal{L}_1$ measures the difference between the rendered image $\mathcal{\tilde{I}}$ and the ground truth image $I_{gt}$:
%
\begin{align*}
    \mathcal{L}_1 = ||\mathcal{\tilde{I}} - I_{gt}||_1.
\end{align*}
%
To enhance the rendering fidelity and achieve better convergence on $\mathcal{L}_1$, we add a second term of perceptual loss as
%
\begin{align*}
    \mathcal{L}_{VGG} = \sum_{i} ||VGG_i(\mathcal{\tilde{I}}) - VGG_i(I_{gt})||_1,
\end{align*}
%
\noindent where $VGG_i(\cdot)$ indicates extracting the intermediate feature from the $i$th layer of a pretrained VGG network.
%
The last term $\mathcal{L}_{seg}$ is segmentation loss, 
%
\begin{align*}
    \mathcal{L}_{seg} = ||\mathcal{M} - M_{gt}||_1,
\end{align*}
%
\noindent which is the $L_1$ distance between the rendered mask $\mathcal{M}$ and the ground truth segmentation mask $M_{gt}$.

To mitigate overfitting, we perform data augmentation while training.
%
In order to mimic the noise pattern in the input point cloud, we randomly jitter each point in the point cloud $\bm{q}$ with a gaussian noise.
%
We find that data augmentation helps stabilize the training.
\fi
\section{Experiments}

%\subsection{Dataset} 

\noindent\textbf{Dataset.} We collect a multiview RGB image dataset of multiple identities with diverse personalized hairstyles. The dataset contains lightstage capture of around 260 identities and each capture has around 160 views covering most perspectives around the participant with a focus on the head and hair region. We exclude captures of 8 identities from training our model and use them just for test purposes. For all 256 training identities, we also hold out 7 views for testing and the rest of the views will be used for training.

\begin{figure}[tb!]
\setlength\tabcolsep{0pt}%%
\renewcommand{\arraystretch}{0}%%
\centering
\begin{tabular}{cccc}
 \textbf{\scriptsize KeypointNeRF~\cite{keypoint_nerf}} &
 \textbf{\scriptsize Cao ~\textit{et al.}~\cite{ica_chen}} &
 \textbf{\scriptsize Ours} & 
 \textbf{\scriptsize Ground Truth} \\
 \begin{tikzpicture}
    \node[anchor=south west,inner sep=0] (image) at (0,0) {\adjincludegraphics[width=0.25\columnwidth, trim={{0.1\width} {0.2\height} {0.05\width} {0.05\height}}, clip]{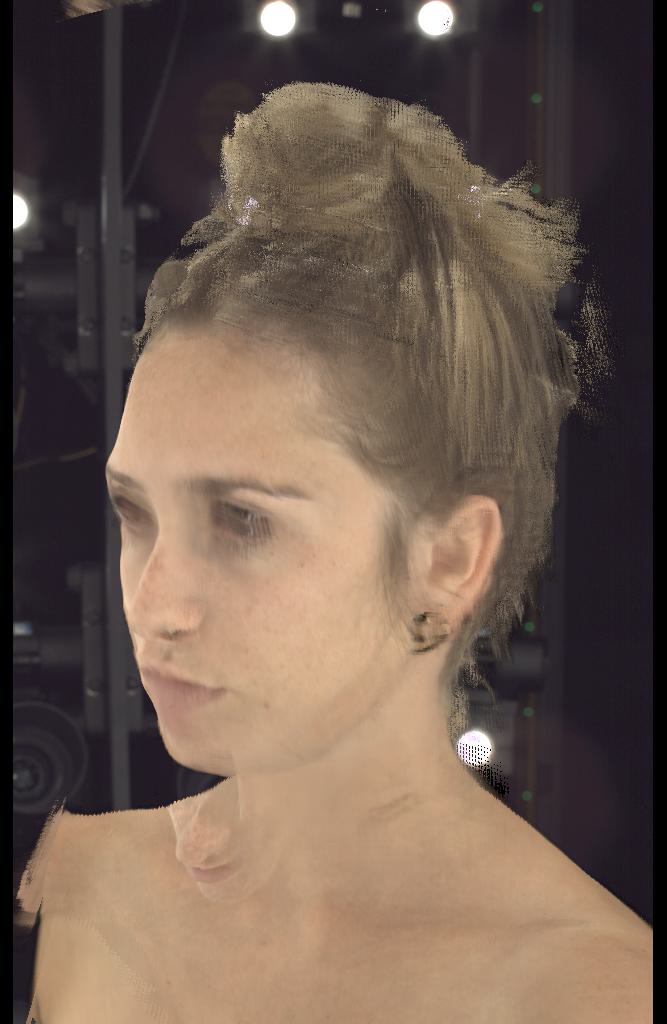}};
 \end{tikzpicture}
&
\begin{tikzpicture}
    \node[anchor=south west,inner sep=0] (image) at (0,0) {\adjincludegraphics[width=0.25\columnwidth, trim={{0.1\width} {0.2\height} {0.05\width} {0.05\height}}, clip]{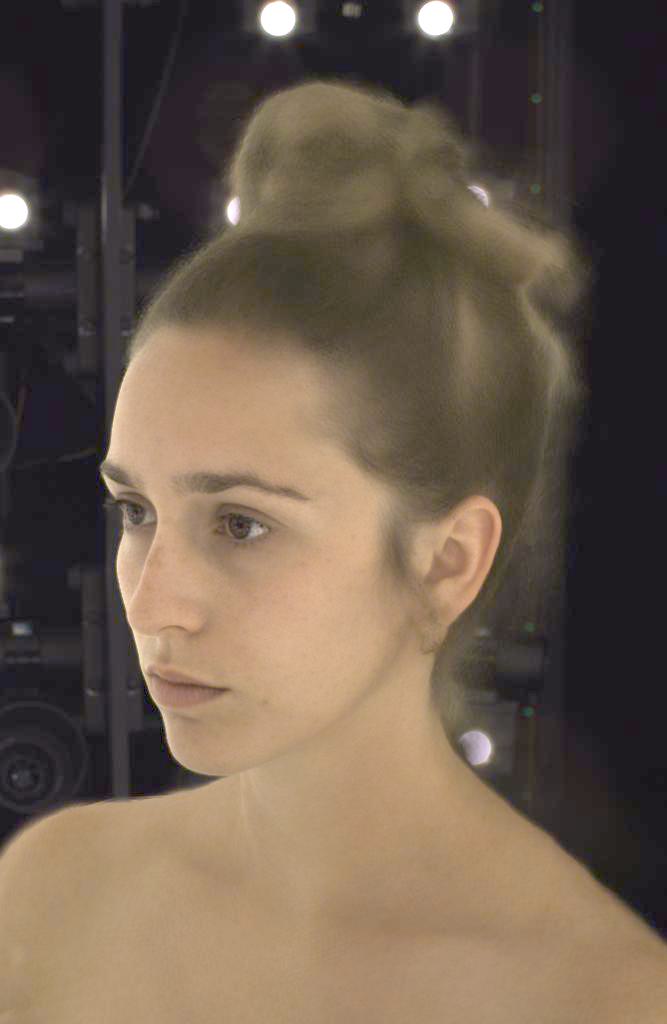}};
 \end{tikzpicture}
&
\begin{tikzpicture}
    \node[anchor=south west,inner sep=0] (image) at (0,0) {\adjincludegraphics[width=0.25\columnwidth, trim={{0.1\width} {0.2\height} {0.05\width} {0.05\height}}, clip]{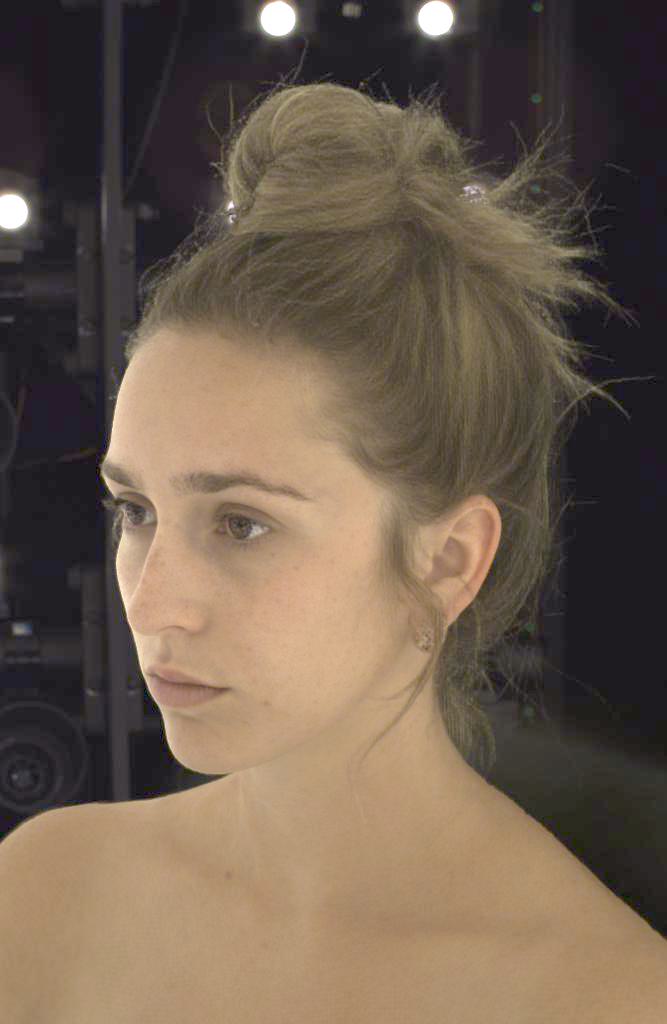}};
 \end{tikzpicture}
&
\begin{tikzpicture}
    \node[anchor=south west,inner sep=0] (image) at (0,0) {\adjincludegraphics[width=0.25\columnwidth, trim={{0.1\width} {0.2\height} {0.05\width} {0.05\height}}, clip]{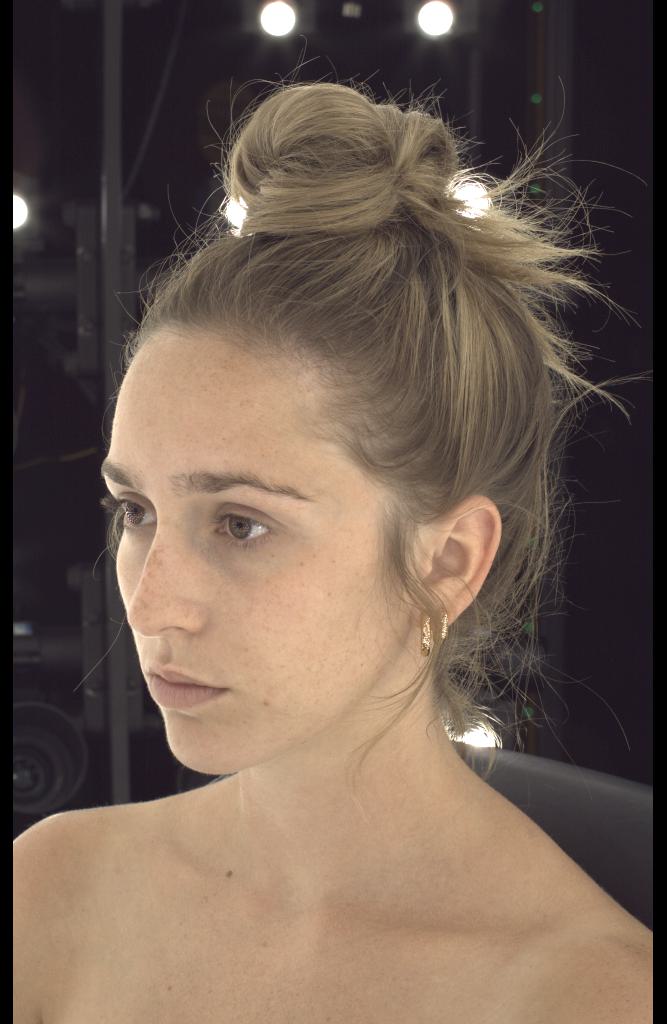}};
 \end{tikzpicture}
\\ 
 \begin{tikzpicture}
    \node[anchor=south west,inner sep=0] (image) at (0,0) {\adjincludegraphics[width=0.25\columnwidth, trim={{0.1\width} {0.2\height} {0.05\width} {0.05\height}}, clip]{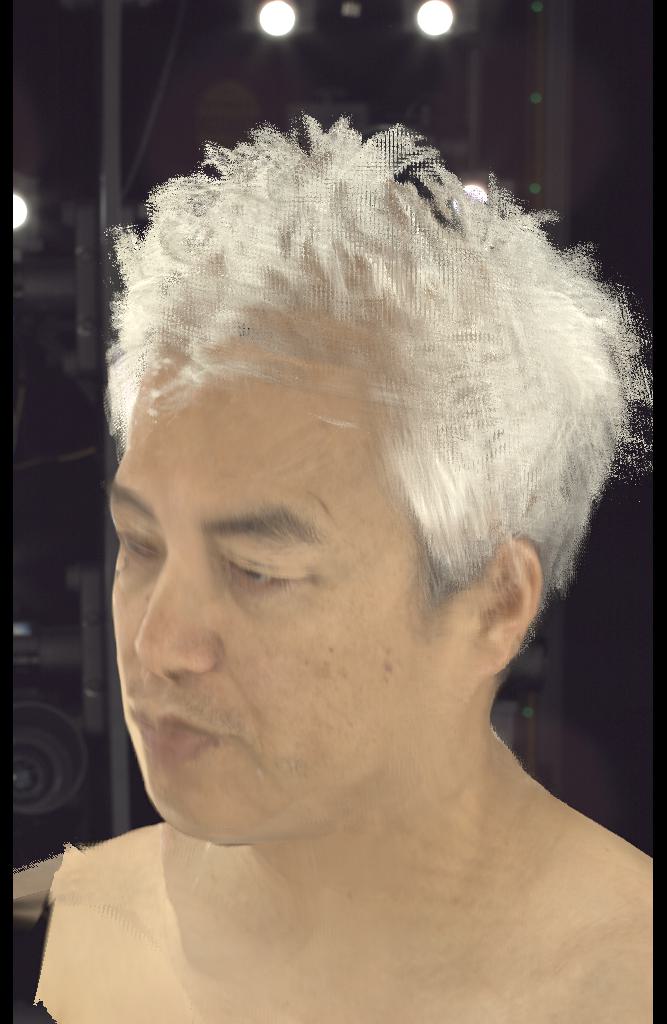}};
 \end{tikzpicture}
&
\begin{tikzpicture}
    \node[anchor=south west,inner sep=0] (image) at (0,0) {\adjincludegraphics[width=0.25\columnwidth, trim={{0.1\width} {0.2\height} {0.05\width} {0.05\height}}, clip]{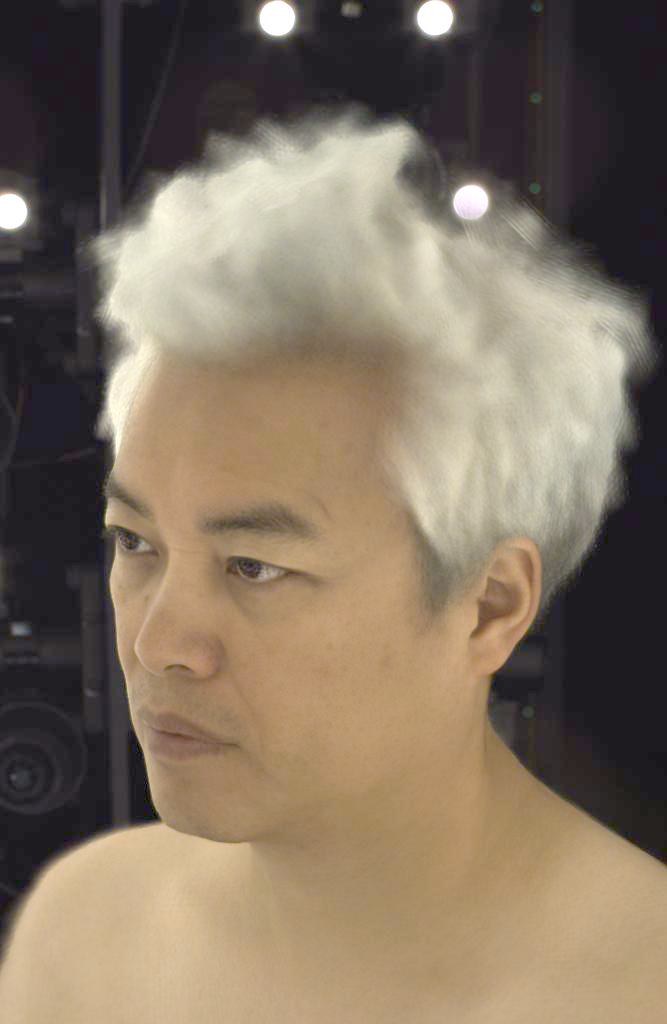}};
 \end{tikzpicture}
&
\begin{tikzpicture}
    \node[anchor=south west,inner sep=0] (image) at (0,0) {\adjincludegraphics[width=0.25\columnwidth, trim={{0.1\width} {0.2\height} {0.05\width} {0.05\height}}, clip]{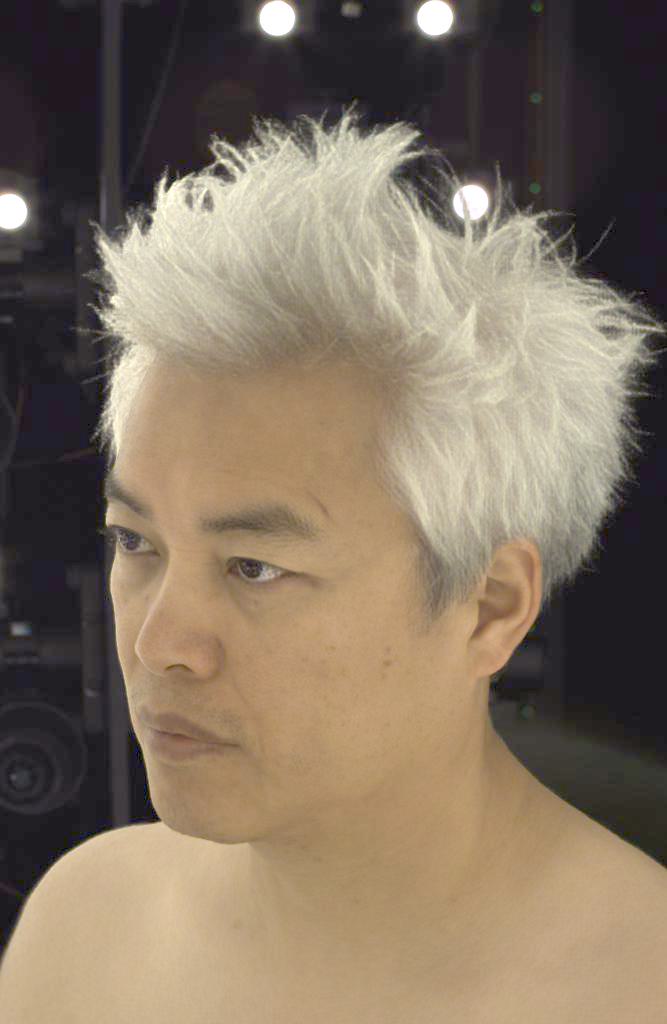}};
 \end{tikzpicture}
&
\begin{tikzpicture}
    \node[anchor=south west,inner sep=0] (image) at (0,0) {\adjincludegraphics[width=0.25\columnwidth, trim={{0.1\width} {0.2\height} {0.05\width} {0.05\height}}, clip]{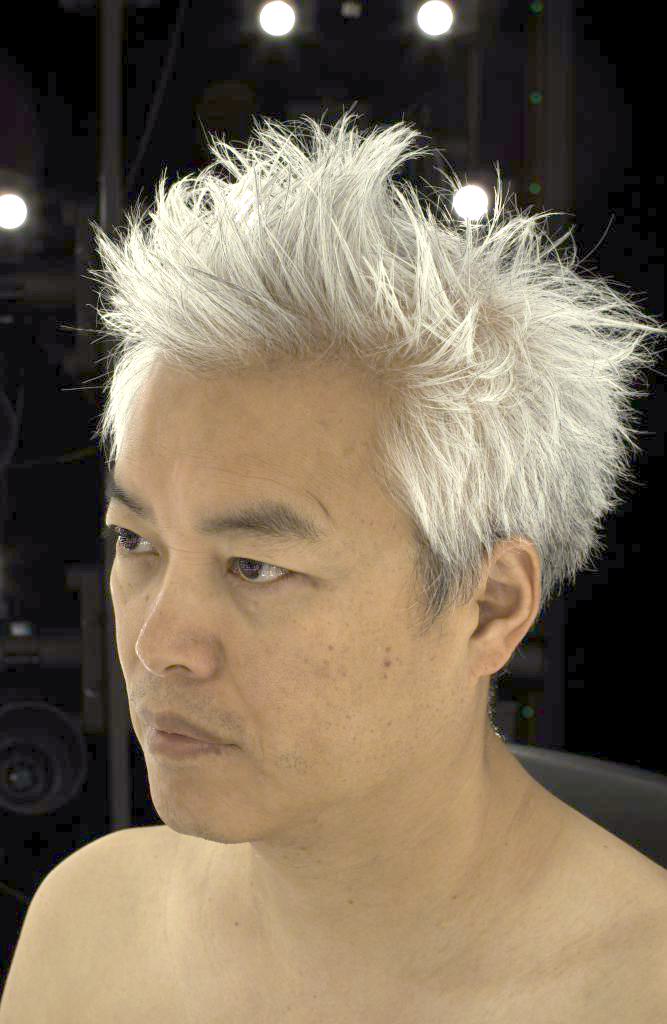}};
 \end{tikzpicture}
\\ 
 \begin{tikzpicture}
    \node[anchor=south west,inner sep=0] (image) at (0,0) {\adjincludegraphics[width=0.25\columnwidth, trim={0 {0.2\height} {0.05\width} {0.05\height}}, clip]{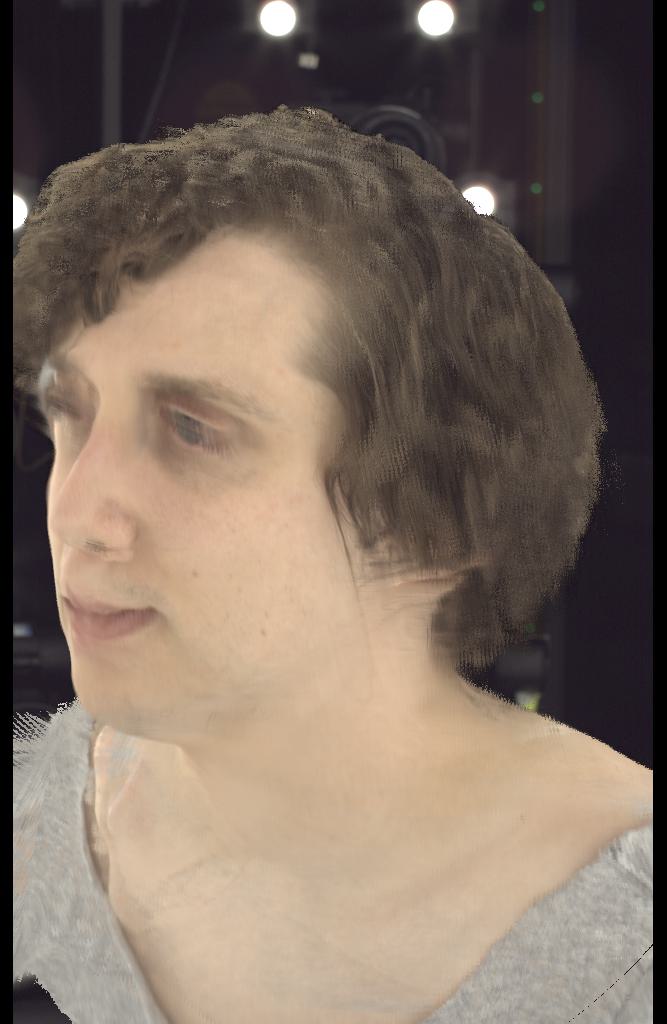}};
 \end{tikzpicture}
&
\begin{tikzpicture}
    \node[anchor=south west,inner sep=0] (image) at (0,0) {\adjincludegraphics[width=0.25\columnwidth, trim={0 {0.2\height} {0.05\width} {0.05\height}}, clip]{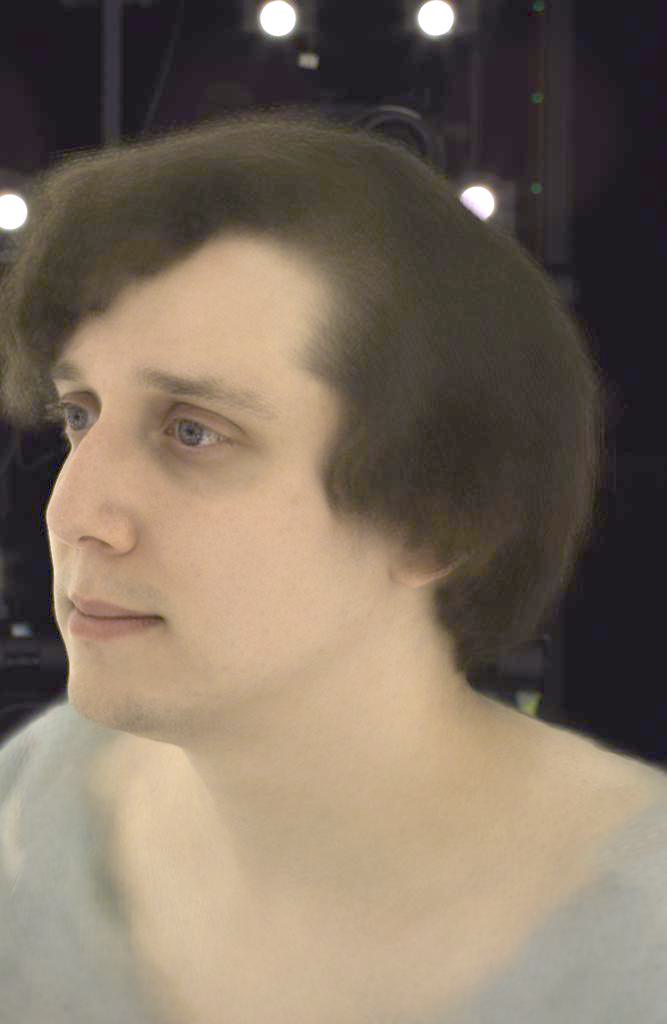}};
 \end{tikzpicture}
&
\begin{tikzpicture}
    \node[anchor=south west,inner sep=0] (image) at (0,0) {\adjincludegraphics[width=0.25\columnwidth, trim={0 {0.2\height} {0.05\width} {0.05\height}}, clip]{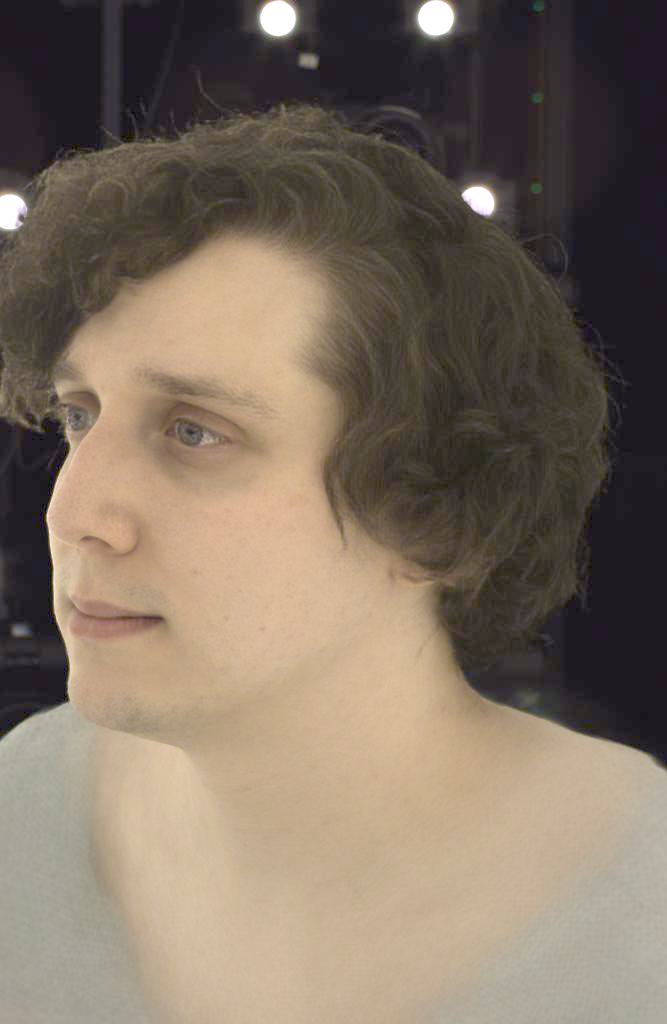}};
 \end{tikzpicture}
&
\begin{tikzpicture}
    \node[anchor=south west,inner sep=0] (image) at (0,0) {\adjincludegraphics[width=0.25\columnwidth, trim={0 {0.2\height} {0.05\width} {0.05\height}}, clip]{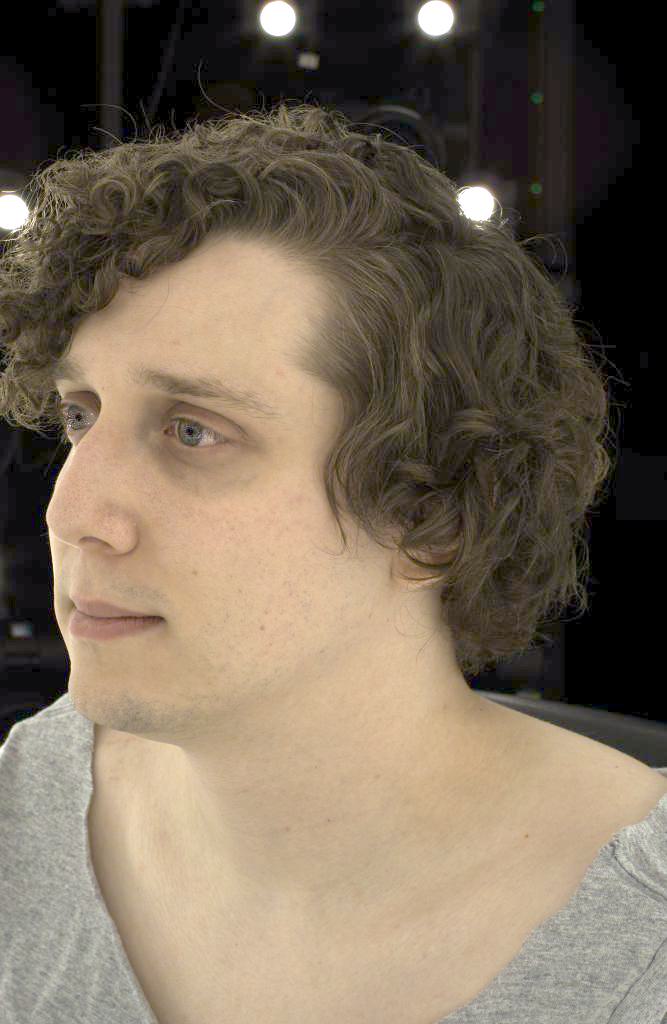}};
 \end{tikzpicture}
\\ 
 \begin{tikzpicture}
    \node[anchor=south west,inner sep=0] (image) at (0,0) {\adjincludegraphics[width=0.25\columnwidth, trim={{0.2\width} {0.2\height} {0.05\width} {0.2\height}}, clip]{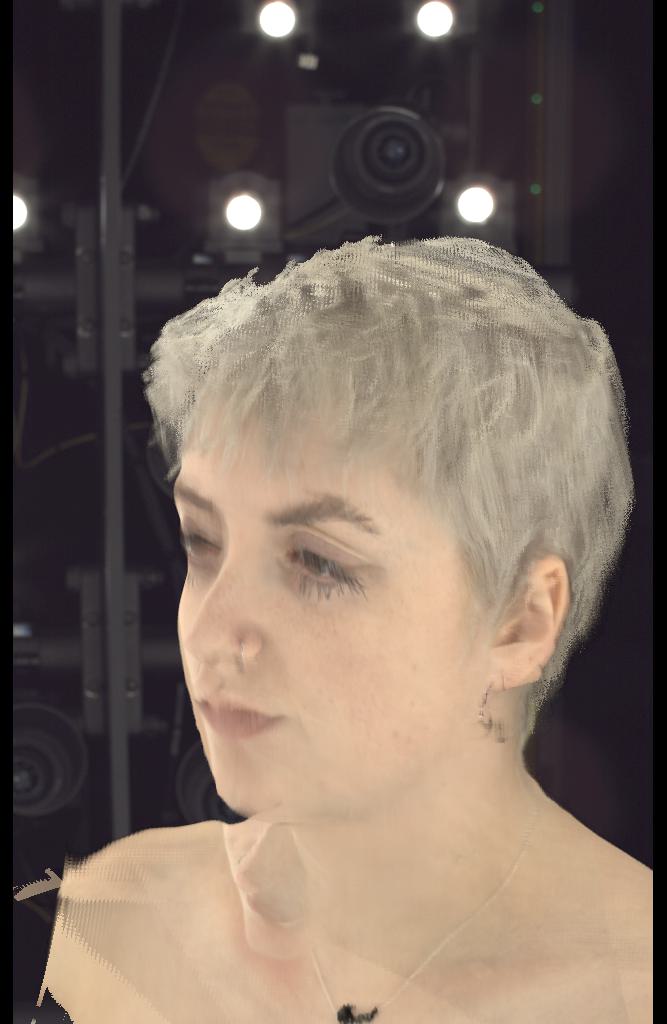}};
 \end{tikzpicture}
&
\begin{tikzpicture}
    \node[anchor=south west,inner sep=0] (image) at (0,0) {\adjincludegraphics[width=0.25\columnwidth, trim={{0.2\width} {0.2\height} {0.05\width} {0.2\height}}, clip]{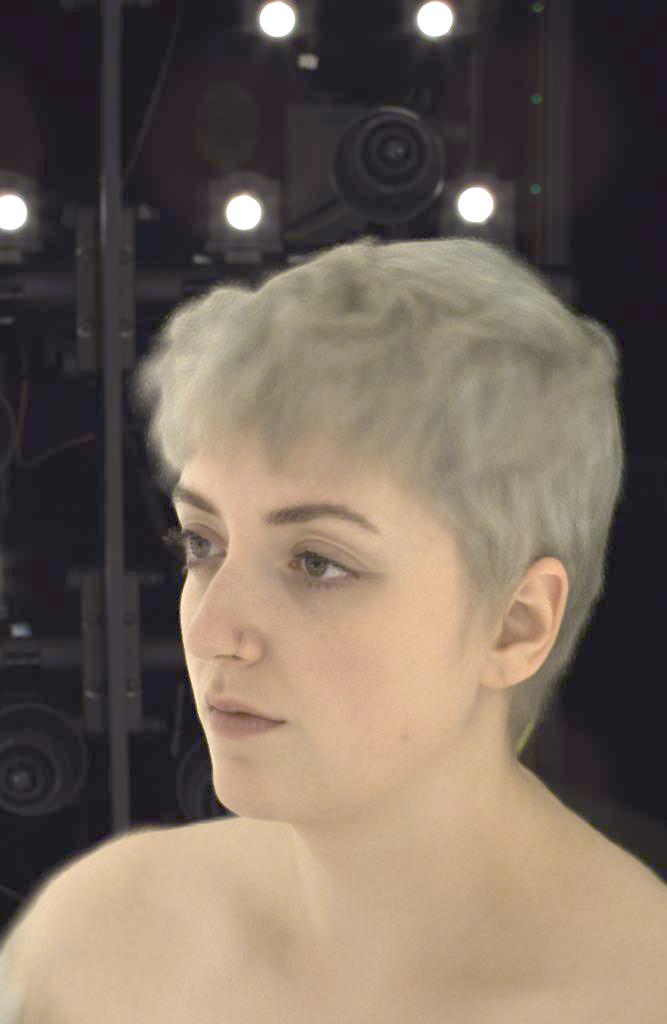}};
 \end{tikzpicture}
&
\begin{tikzpicture}
    \node[anchor=south west,inner sep=0] (image) at (0,0) {\adjincludegraphics[width=0.25\columnwidth, trim={{0.2\width} {0.2\height} {0.05\width} {0.2\height}}, clip]{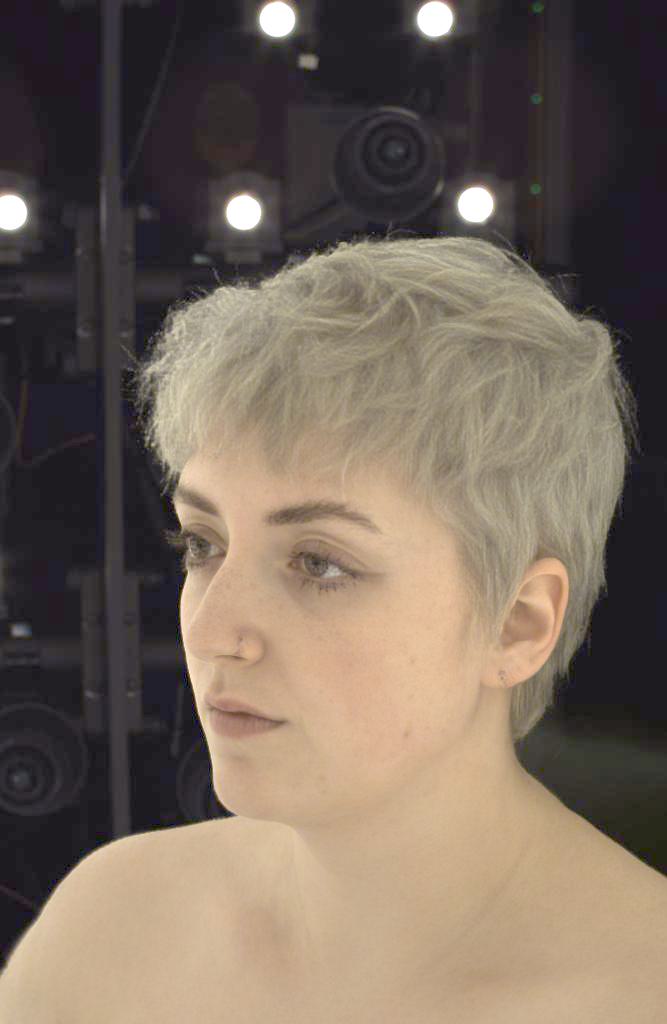}};
 \end{tikzpicture}
&
\begin{tikzpicture}
    \node[anchor=south west,inner sep=0] (image) at (0,0) {\adjincludegraphics[width=0.25\columnwidth, trim={{0.2\width} {0.2\height} {0.05\width} {0.2\height}}, clip]{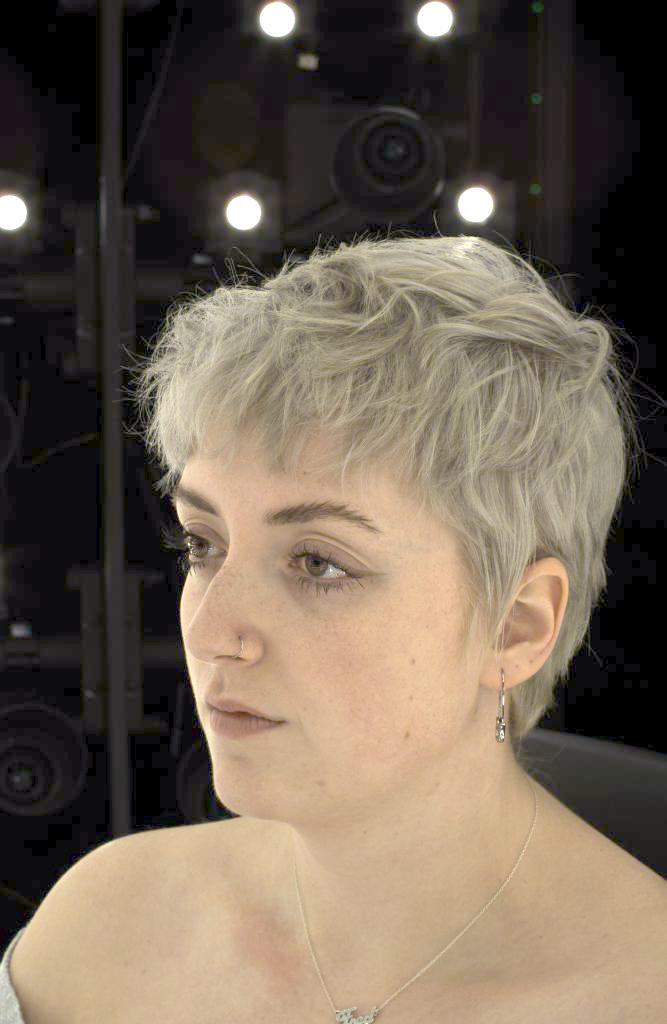}};
 \end{tikzpicture}
\end{tabular}
\caption{\label{prop_eval:fig_nvs_train}\textbf{Novel View Synthesis.} Rendering results on the holdout views of the training identities. We compare our method with KeypointNeRF~\cite{keypoint_nerf} and Cao ~\textit{et al.}~\cite{ica_chen}. Our method is compatible with different hair geometries and captures the detailed volumetric texture of varied hairstyles. Please refer to the supplemental materials for more rendering results and comparisons.\iffalse\textcolor{red}{Would be good to include rendering of point cloud used in our method. Similarly, comparing against classical texturized 3D recon is an important baseline}\fi}
\end{figure}
%
%\noindent\textbf{Novel view synthesis.} 
\subsection{Novel View Synthesis}
\begin{figure}[tb!]
\setlength\tabcolsep{0pt}%%
\renewcommand{\arraystretch}{0}%%
\centering
\begin{tabular}{cccc}
 \textbf{\scriptsize KeypointNeRF~\cite{keypoint_nerf}} &
 \textbf{\scriptsize Cao ~\textit{et al.}~\cite{ica_chen}} &
 \textbf{\scriptsize Ours} & 
 \textbf{\scriptsize Ground Truth} \\
 \begin{tikzpicture}
    \node[anchor=south west,inner sep=0] (image) at (0,0) {\adjincludegraphics[width=0.25\columnwidth, trim={{0.15\width} {0.2\height} {0.\width} {0.08\height}}, clip]{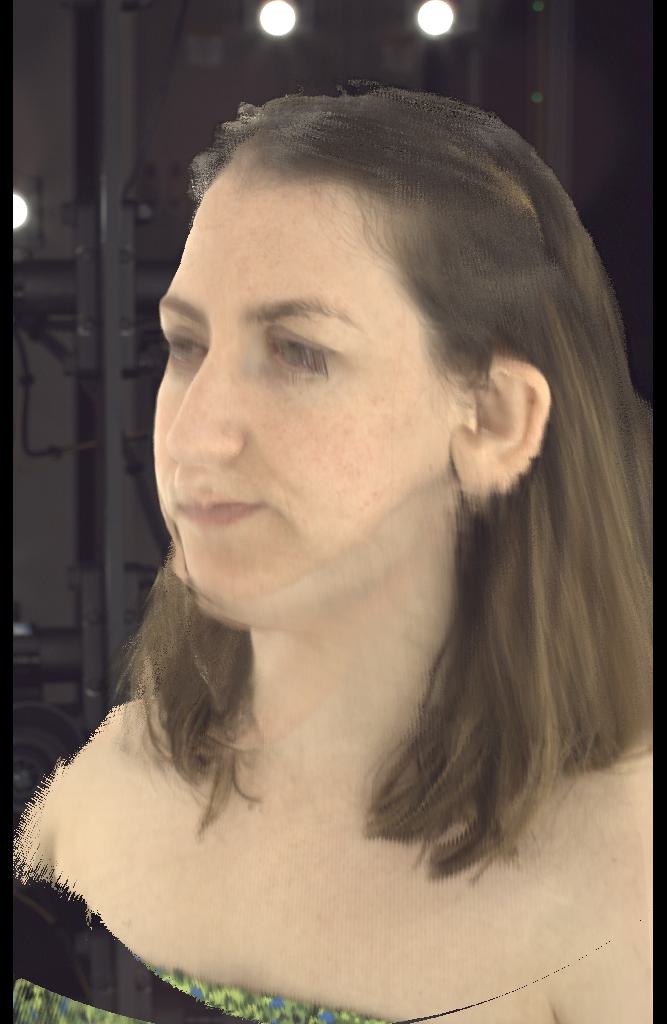}};
 \end{tikzpicture}
&
\begin{tikzpicture}
    \node[anchor=south west,inner sep=0] (image) at (0,0) {\adjincludegraphics[width=0.25\columnwidth, trim={{0.15\width} {0.2\height} {0.\width} {0.08\height}}, clip]{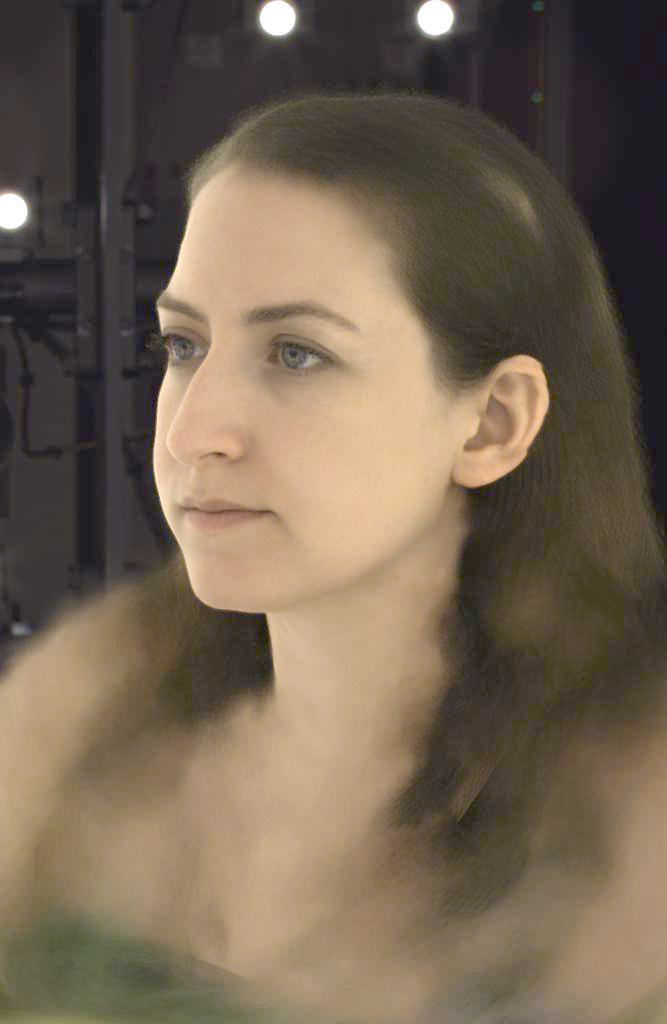}};
 \end{tikzpicture}
&
\begin{tikzpicture}
    \node[anchor=south west,inner sep=0] (image) at (0,0) {\adjincludegraphics[width=0.25\columnwidth, trim={{0.15\width} {0.2\height} {0.\width} {0.08\height}}, clip]{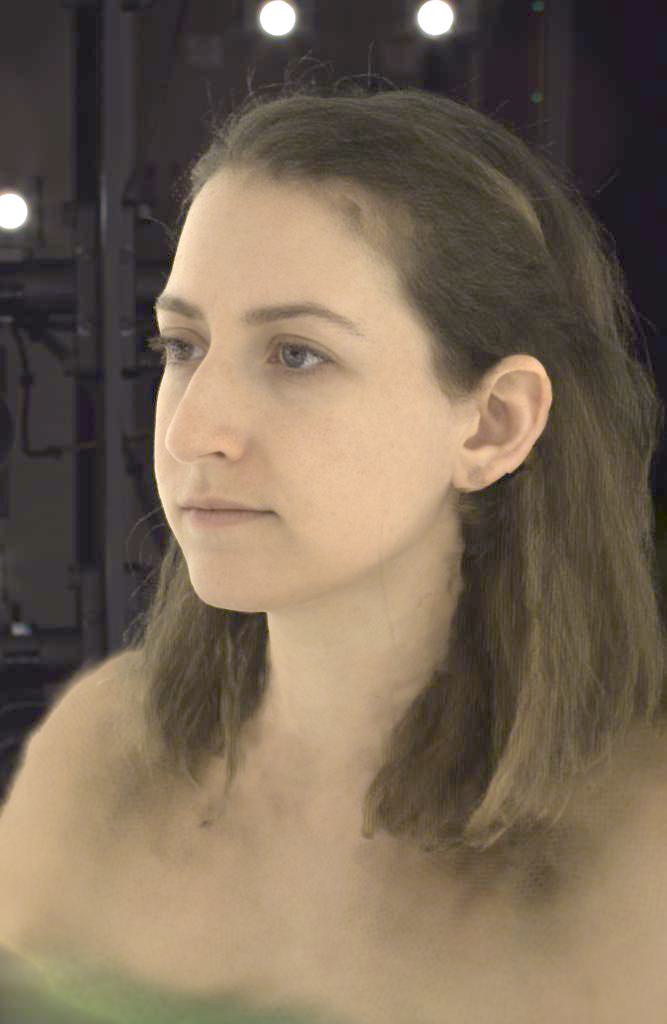}};
 \end{tikzpicture}
&
\begin{tikzpicture}
    \node[anchor=south west,inner sep=0] (image) at (0,0) {\adjincludegraphics[width=0.25\columnwidth, trim={{0.15\width} {0.2\height} {0.\width} {0.08\height}}, clip]{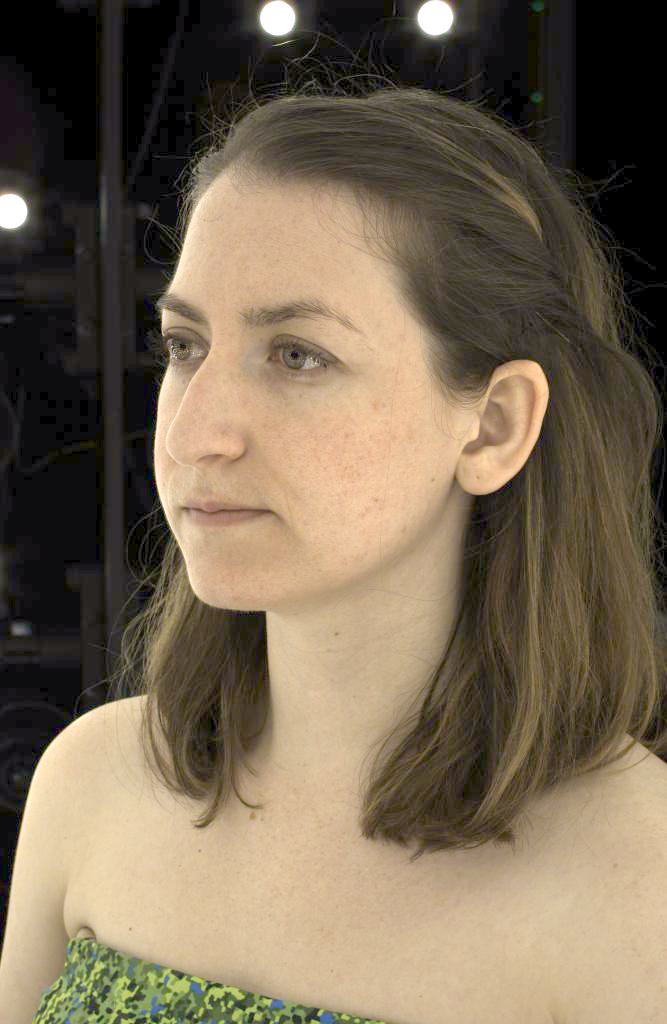}};
 \end{tikzpicture}
\\ 
 \begin{tikzpicture}
    \node[anchor=south west,inner sep=0] (image) at (0,0) {\adjincludegraphics[width=0.25\columnwidth, trim={{0.1\width} {0.1\height} {0.05\width} {0.15\height}}, clip]{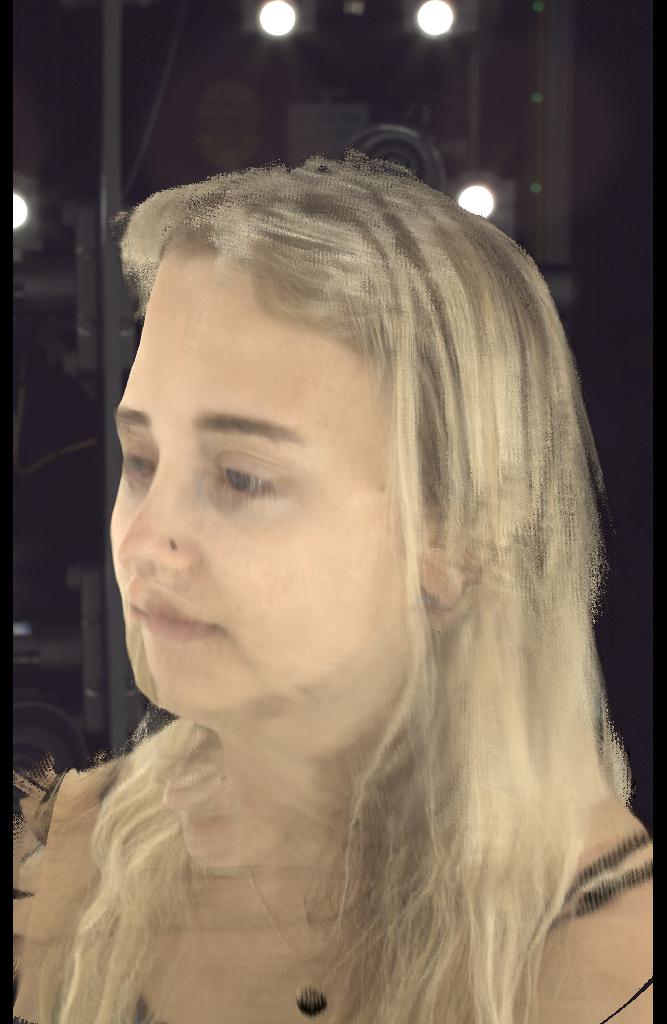}};
 \end{tikzpicture}
&
\begin{tikzpicture}
    \node[anchor=south west,inner sep=0] (image) at (0,0) {\adjincludegraphics[width=0.25\columnwidth, trim={{0.1\width} {0.1\height} {0.05\width} {0.15\height}}, clip]{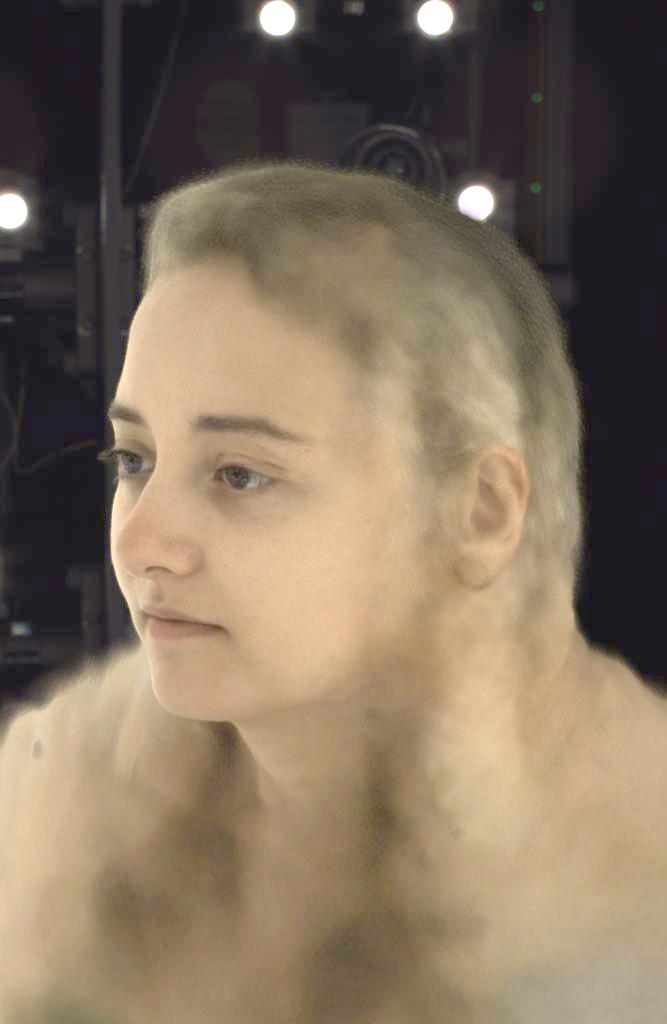}};
 \end{tikzpicture}
&
\begin{tikzpicture}
    \node[anchor=south west,inner sep=0] (image) at (0,0) {\adjincludegraphics[width=0.25\columnwidth, trim={{0.1\width} {0.1\height} {0.05\width} {0.15\height}}, clip]{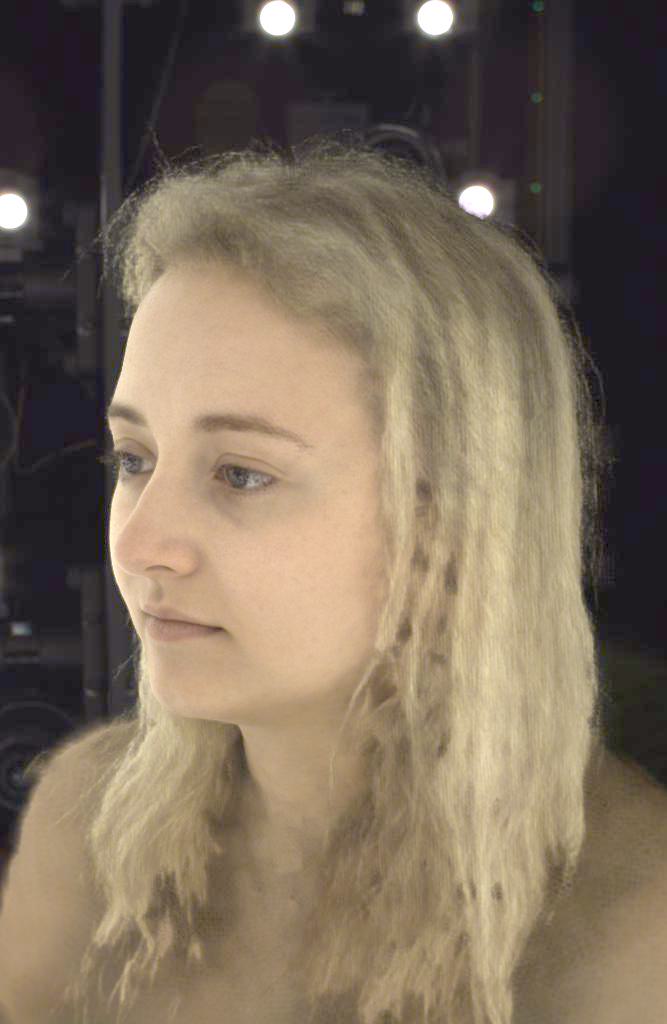}};
 \end{tikzpicture}
&
\begin{tikzpicture}
    \node[anchor=south west,inner sep=0] (image) at (0,0) {\adjincludegraphics[width=0.25\columnwidth, trim={{0.1\width} {0.1\height} {0.05\width} {0.15\height}}, clip]{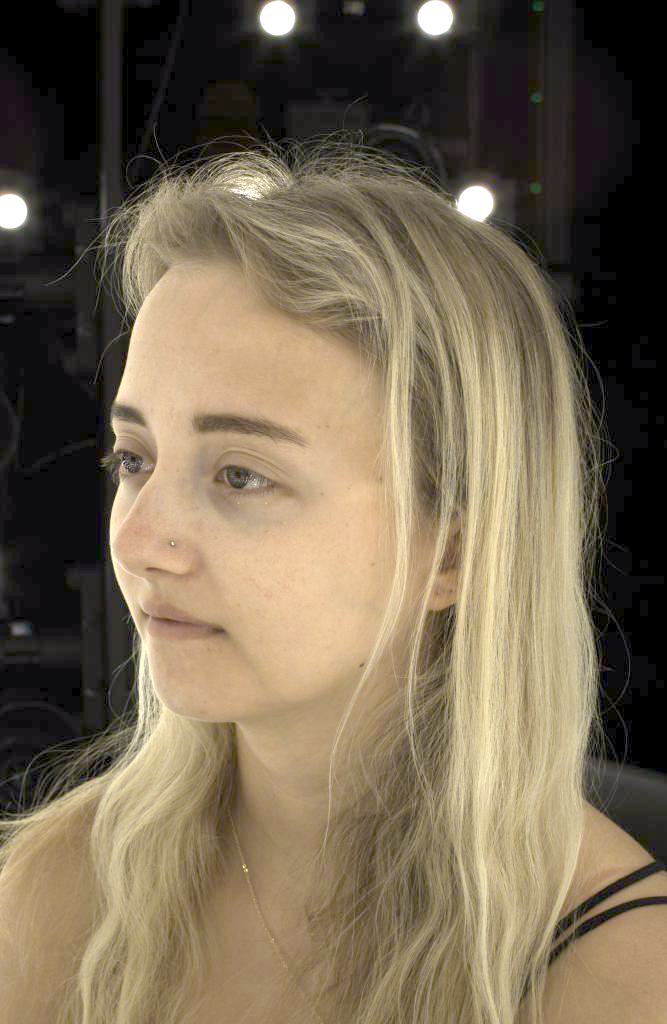}};
 \end{tikzpicture}
\\ 
 \begin{tikzpicture}
    \node[anchor=south west,inner sep=0] (image) at (0,0) {\adjincludegraphics[width=0.25\columnwidth, trim={{0.1\width} {0.2\height} {0.05\width} {0.05\height}}, clip]{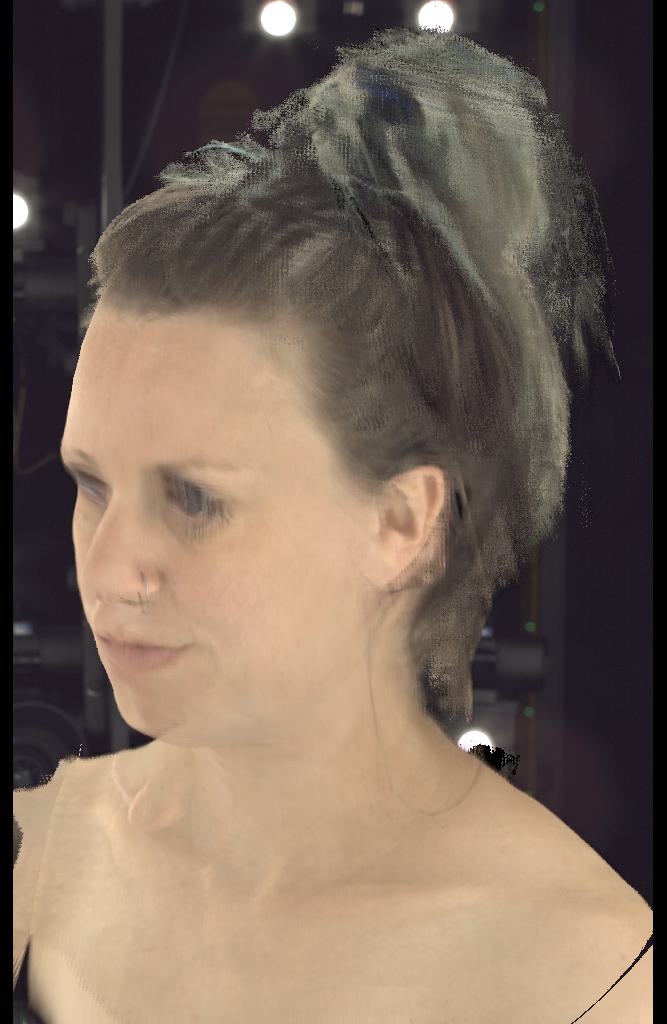}};
 \end{tikzpicture}
&
\begin{tikzpicture}
    \node[anchor=south west,inner sep=0] (image) at (0,0) {\adjincludegraphics[width=0.25\columnwidth, trim={{0.1\width} {0.2\height} {0.05\width} {0.05\height}}, clip]{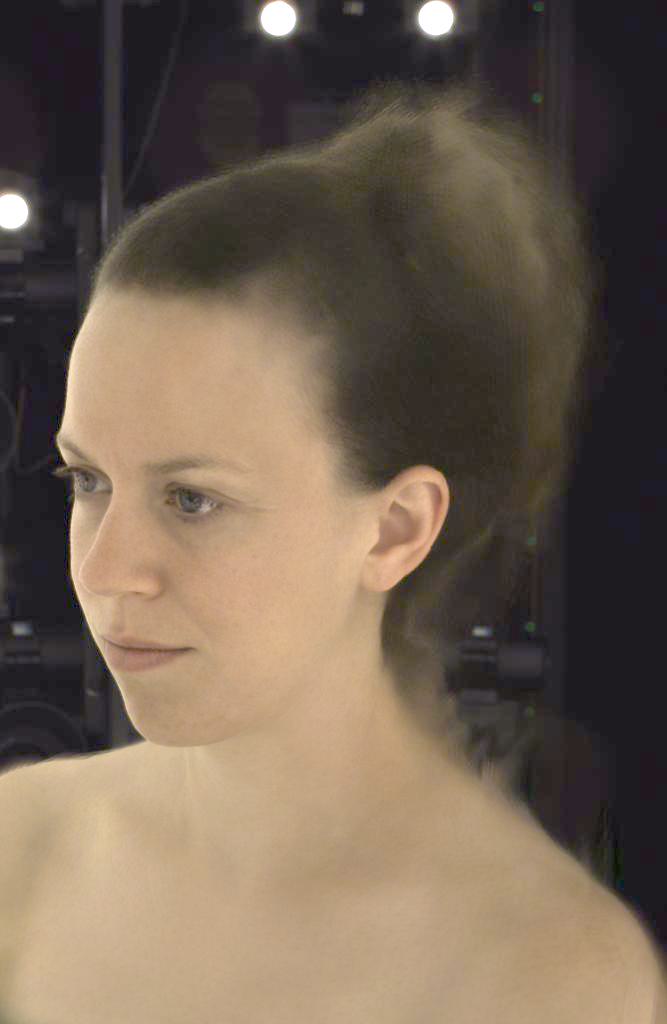}};
 \end{tikzpicture}
&
\begin{tikzpicture}
    \node[anchor=south west,inner sep=0] (image) at (0,0) {\adjincludegraphics[width=0.25\columnwidth, trim={{0.1\width} {0.2\height} {0.05\width} {0.05\height}}, clip]{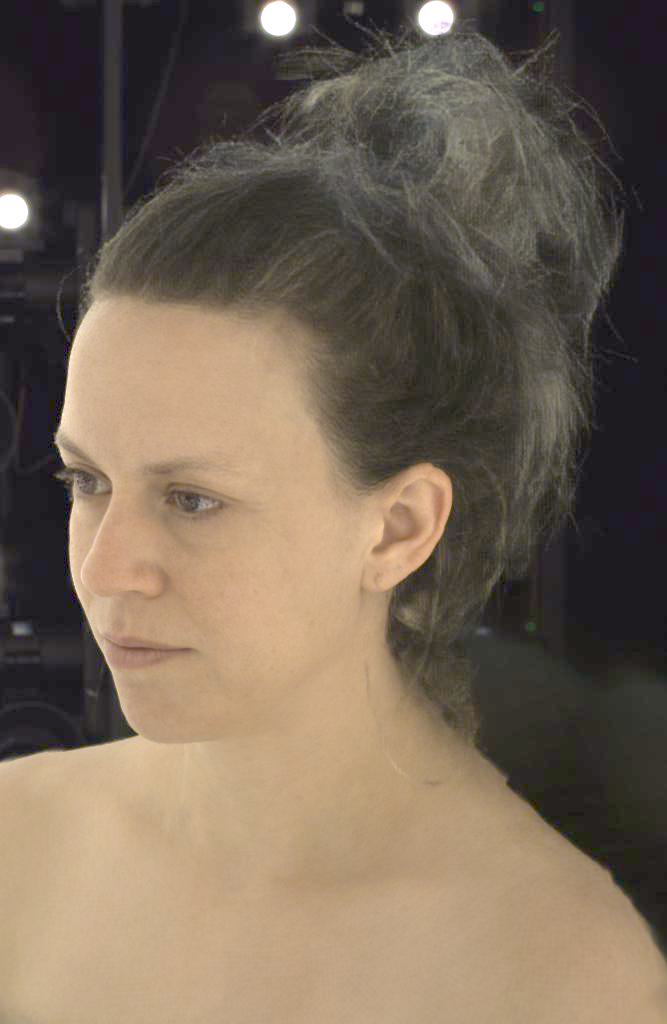}};
 \end{tikzpicture}
&
\begin{tikzpicture}
    \node[anchor=south west,inner sep=0] (image) at (0,0) {\adjincludegraphics[width=0.25\columnwidth, trim={{0.1\width} {0.2\height} {0.05\width} {0.05\height}}, clip]{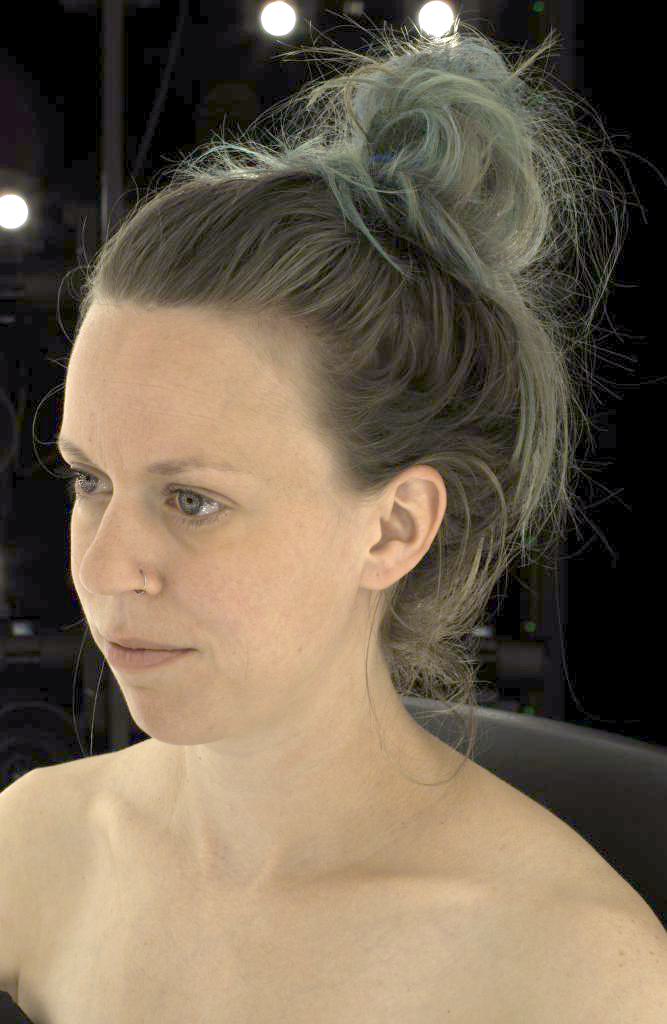}};
 \end{tikzpicture}
\\ 
 \begin{tikzpicture}
    \node[anchor=south west,inner sep=0] (image) at (0,0) {\adjincludegraphics[width=0.25\columnwidth, trim={{0.2\width} {0.3\height} {0.05\width} {0.1\height}}, clip]{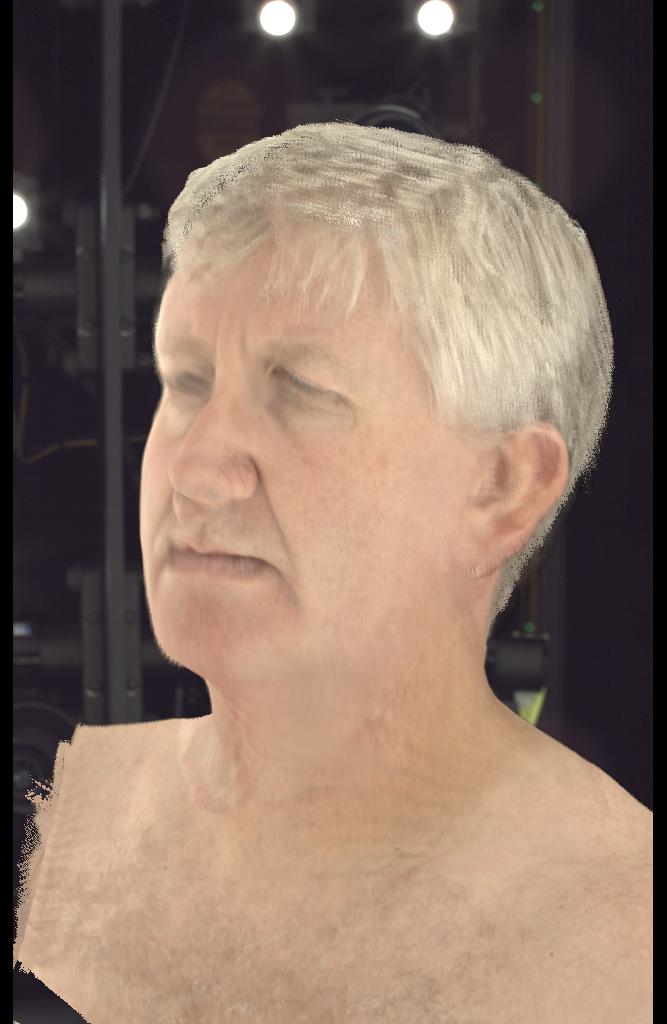}};
 \end{tikzpicture}
&
\begin{tikzpicture}
    \node[anchor=south west,inner sep=0] (image) at (0,0) {\adjincludegraphics[width=0.25\columnwidth, trim={{0.2\width} {0.3\height} {0.05\width} {0.1\height}}, clip]{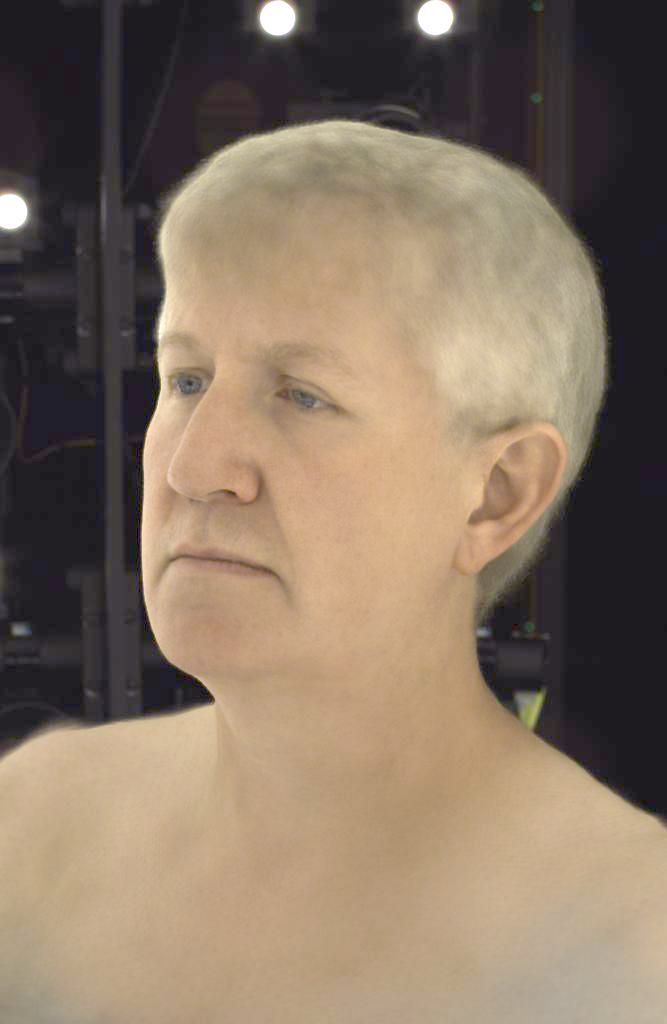}};
 \end{tikzpicture}
&
\begin{tikzpicture}
    \node[anchor=south west,inner sep=0] (image) at (0,0) {\adjincludegraphics[width=0.25\columnwidth, trim={{0.2\width} {0.3\height} {0.05\width} {0.1\height}}, clip]{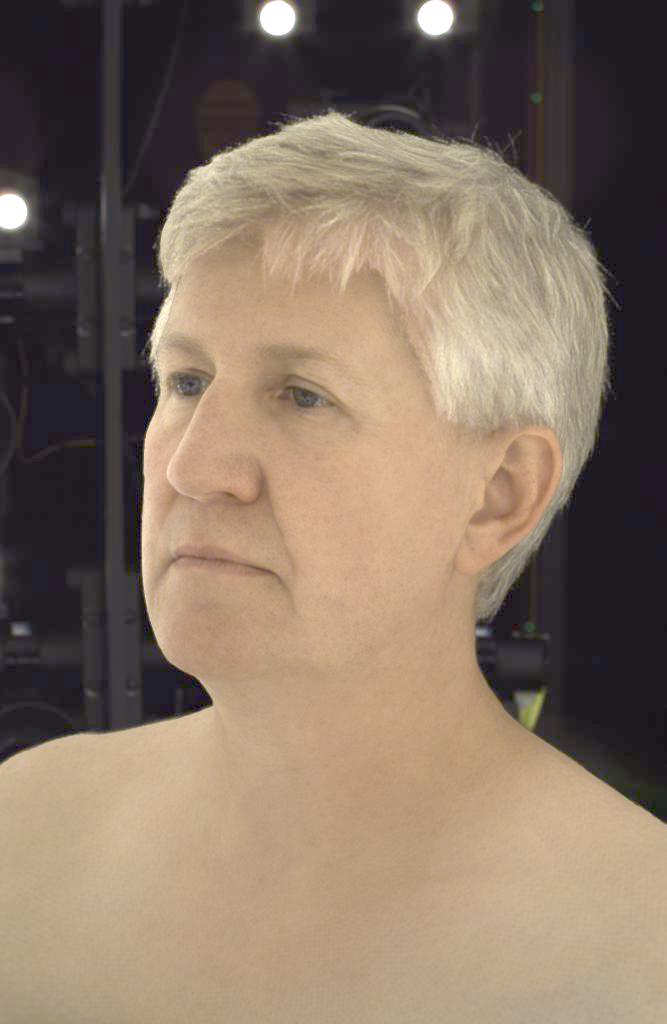}};
 \end{tikzpicture}
&
\begin{tikzpicture}
    \node[anchor=south west,inner sep=0] (image) at (0,0) {\adjincludegraphics[width=0.25\columnwidth, trim={{0.2\width} {0.3\height} {0.05\width} {0.1\height}}, clip]{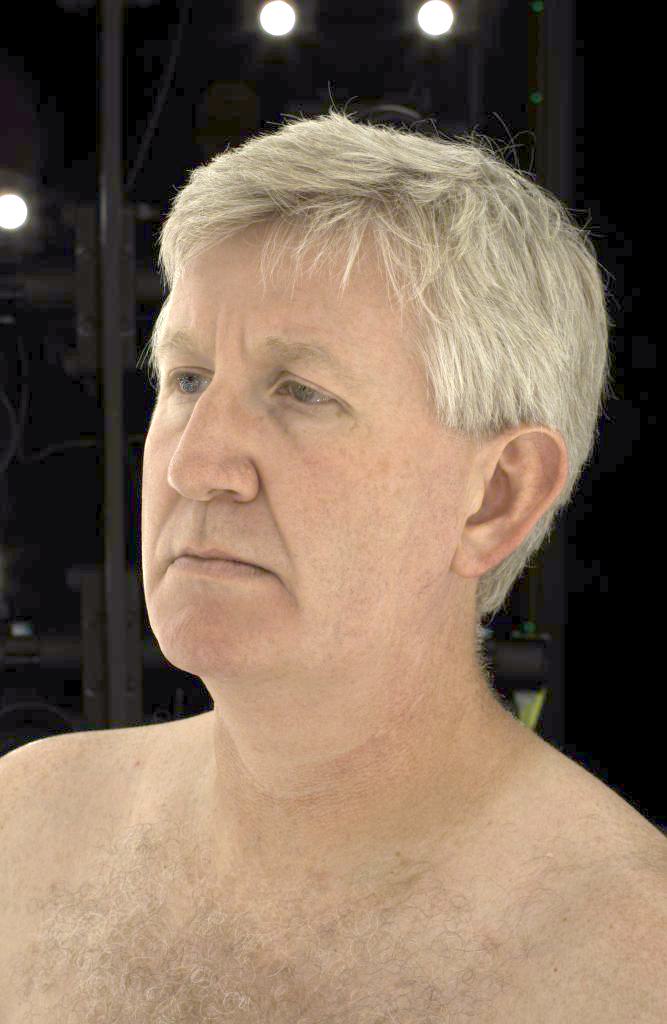}};
 \end{tikzpicture}
\end{tabular}
\caption{\label{prop_eval:fig_nvs_test}\textbf{Novel View Synthesis.} Rendering results on the test identities. We compare our method with KeypointNeRF~\cite{keypoint_nerf} and Cao \textit{et al.}~\cite{ica_chen}. Our method generalizes reasonably to new identities and is capable of generating a photorealistic appearance without any finetuning. Please refer to the supplemental materials for more rendering results and comparisons.}
\end{figure}

We test our model on the task of novel view synthesis and compare it with previous state-of-the-art approaches on generating personalized avatars like the universal prior model(UPM) in Cao \textit{et al.}~\cite{ica_chen} and a generalizable NeRF model KeypointNeRF~\cite{keypoint_nerf}. 
\begin{table}[tb!]
\centering
\resizebox{\columnwidth}{!}
{
\begin{tabular}{rcccc}
\toprule 
\multicolumn{1}{c}{TRAIN} & MSE($\downarrow$)    & PSNR($\uparrow$)  & SSIM($\uparrow$)   & LPIPS($\downarrow$)  \\
\midrule
KeypointNeRF~\cite{keypoint_nerf}              & 257.42 & 24.57 & 0.86   & 0.3140 \\
Cao ~\textit{et al.}~\cite{ica_chen}                        & 159.37 & 27.12 & 0.7961 & 0.3117 \\
Ours                      & \textbf{130.07} & \textbf{27.73} & \textbf{0.8922} & \textbf{0.1993} \\
\bottomrule
\toprule
\multicolumn{1}{c}{TEST}  & MSE($\downarrow$)    & PSNR($\uparrow$)  & SSIM($\uparrow$)   & LPIPS($\downarrow$)  \\
\midrule
KeypointNeRF~\cite{keypoint_nerf}              & 303.60 & 23.77 & 0.8596 & 0.3389 \\
Cao ~\textit{et al.}~\cite{ica_chen}                        & 334.68 & 23.89 & 0.7883 & 0.3511 \\
Ours                      & \textbf{236.46} & \textbf{25.08} & \textbf{0.8741} & \textbf{0.2610} \\
\bottomrule
\end{tabular}
}
\caption{\label{prop_eval:tab1}\textbf{Novel View Synthesis.} We show qualitative results on novel view synthesis. The upper part and lower part of the table report the MSE, PSNR, SSIM and LPIPS computed on the holdout views of training identities and the test identities respectively.
Our method achieves a better result on MSE, PSNR and SSIM compared to previous methods~\cite{ica_chen, keypoint_nerf}.
Our method is capable of generating sharp appearance on detailed geometries which leads to improvement on LPIPS by a large margin.
}
\end{table}
We evaluate different methods using image reconstruction and similarity metrics like MSE, PSRN, SSIM and LPIPS between the ground true image and the reconstructed ones, which are reported in Tab.~\ref{prop_eval:tab1}. 
In the upper part of the table, we report the metrics computed on the holdout views of the training identities. 
The results in Tab.~\ref{prop_eval:tab1} are supposed to reflect how well each model reconstructs the appearance and shape of the head and hair from those training identities. 
Compared to KeypointNeRF~\cite{keypoint_nerf}, we achieve a lower distortion in terms of reconstructing different hairstyles. 
When compared with the UPM model~\cite{ica_chen}, we find that our model enjoys a much larger improvement on LPIPS compared to the other reconstruction metrics like MSE, PSNR and SSIM. 
One of the reasons behind this result is that the perceptual metric is more sensitive to high-frequency information as well as the fine-level details in one's appearance.
The UPM model is capable of reconstructing a coarse-level geometry and appearance but fails to capture the fine-level details which our model does a better job on. 
Furthermore, the UPM model is inherently limited by its use of a mesh to represent complex hair topology, even at a coarse level.
In Fig.~\ref{prop_eval:fig_nvs_train}, the improvements of our methods can be better justified visually, where we show the rendering results of each method on the holdout views of several training identities.
On the lower part of Tab.~\ref{prop_eval:tab1}, we report the metrics computed on the same set of holdout views but of test identities.
Fig.~\ref{prop_eval:fig_nvs_test} shows the rendering results of some of those views.
We can see that the UPM model~\cite{ica_chen} and ours both achieve better generalization on the face than KeypointNeRF~\cite{keypoint_nerf} as a result of having awareness of face geometry.
While pixel-aligned information from multi-view images is used as input for KeypointNeRF, learning a generalizable network to triangulate these 2D observations can be challenging, particularly for diverse geometries.
Our method can achieve a more detailed hair appearance given sparse inputs like point clouds on never-seen-before identities.

\noindent\textbf{Ablation on input features $\Omega^{\rho_i}$, $\Gamma^{\rho_i}$ and $\Lambda^{\rho_i}$.} 
We ablate on the usage of different input features to the local appearance networks $\Psi_\alpha(\cdot)$ and $\Psi_{rgb}(\cdot)$.
Tab.~\ref{prop_eval:tab2} shows the performance of our model under different input configurations.
The base model is $\Psi_\alpha(\Omega^{\rho_i})+\Psi_{rgb}(\Omega^{\rho_i},\bm{\hat{c}})$, which only takes $\Omega^{\rho_i}$ and $\bm{\hat{c}}$ as input and do not have untied bias for each layer.
$+\Gamma^{\rho}$ represents the model that use $\Gamma^{\rho}$ as additional input to both $\Psi_\alpha(\cdot)$ and $\Psi_{rgb}(\cdot)$. 
$+ub$ stands for adding untied bias to each learnable layer in $\Psi_\alpha(\cdot)$ and $\Psi_{rgb}(\cdot)$.

According to Tab.~\ref{prop_eval:tab2}, the inclusion of $\Gamma^{\rho}$ helps the base model to converge better with improved image reconstruction and perceptual metrics on both training and testing sets.
However, adding untied bias solely gives worse results on both data splits.
We argue that using $\Omega^{\rho}$ as the only input makes the network to be aware of only local region and to be agnostic of positional information, which further exposes the network to the noise in $\Omega^{\rho}$.
Including additional network parameters like the untied bias in such a setting will make the network even harder to learn.
If we combine both $\Gamma^{\rho}$ and the untied bias, we get much better performance on the training set.
This improvement suggests that using $\Gamma^{\rho}$ is more effective than just adding more network parameters, which helps the network to capture the underlying correlation between human hair's appearance and spatial position.
Finally, we find that having per-vertex view conditioning $\Lambda^{\rho}$ as additional input help the model to achieve the best performance on both the training and testing set.
As different hair geometry leads to different shadow and reflectance patterns, the per-vertex viewing condition $\Lambda^{\rho}$ serves as a more informative term than the viewing direction $\bm{\hat{c}}$ to infer the view-conditioned appearance.
\begin{table*}[tb!]
\centering
\resizebox{0.95\textwidth}{!}
{
\begin{tabular}{lcccccccc}
\toprule
            & \multicolumn{4}{c}{TRAIN}      & \multicolumn{4}{c}{TEST}         \\
            \cmidrule(lr){2-5}  \cmidrule(lr){6-9}
            & MSE($\downarrow$)    & PSNR($\uparrow$)  & SSIM($\uparrow$)  & LPIPS($\downarrow$) & MSE($\downarrow$)   & PSNR($\uparrow$) & SSIM($\uparrow$)  & LPIPS($\downarrow$) \\
            \midrule
base        & 150.69 & 27.11 & 0.8090 & \bB{0.2039} & 242.56 & 25.01 & 0.8023 & \bB{0.2556} \\
base+$\Gamma^{\rho}$    & \bB{140.38} & \bB{27.42} & \sB{0.8123} & \sB{0.1973} & \sB{228.41} & \sB{25.19} & \bB{0.8059} & \sB{0.2479} \\
base+$ub$     & 164.04 & 26.70 & 0.8022 & 0.2234 & 253.61 & 24.69 & 0.7969 & 0.2705 \\
base+$\Gamma^{\rho}$+$ub$ & \sB{134.60} & \sB{27.54} & \bB{0.8108} & 0.2042 & \bB{236.46} & \bB{25.08} & \sB{0.8081} & 0.2610 \\
base+$\Gamma^{\rho}$+$ub$+$\Lambda^{\rho}$ & \gB{130.07} & \gB{27.73} & \gB{0.8922} & \gB{0.1993} &  \gB{220.00} & \gB{25.45} & \gB{0.8741} & \gB{0.2472} \\
\bottomrule
\end{tabular}
}
\caption{\label{prop_eval:tab2}\textbf{Ablation on different inputs $\Omega^{\rho_i}$, $\Gamma^{\rho_i}$ and $\Lambda^{\rho_i}$.} We evaluate models with different input configurations and report their MSE, PSNR, SSIM and LPIPS on the holdout views of both training and testing data. As we can see, both $\Gamma^{\rho_i}$ and $\Lambda^{\rho_i}$ serve as a more effective way to improve the model performance compared to just increasing the model's capacity like with untied bias $ub$. We also find that the inclusion of per-vertex viewing direction $\Lambda^{\rho_i}$ improves the model's performance on novel view synthesis by a large margin. 
We use \gB{gold}, \sB{silver} and \bB{bronze} to indicate first, second and third places.
}
\end{table*}

\noindent\textbf{Ablation on finetuning efficiency.} 
Even though our model generalizes reasonably to unseen identities and creates photorealistic avatars for them, finetuning or online optimization is still needed for getting metrically correct personalized avatars.
Thus, we evaluate our model in terms of how well it can help get personalized avatars for novel identities efficiently.
Given RGB-D scans of new subjects with novel hairstyles, we finetune our model based on them to capture the personalized 3D avatar.
We ablate on both the number of views needed to do finetuning as well as the number of iterations we update our model.

\begin{figure}[tb!]
    \centering
    \includegraphics[width=0.95\columnwidth]{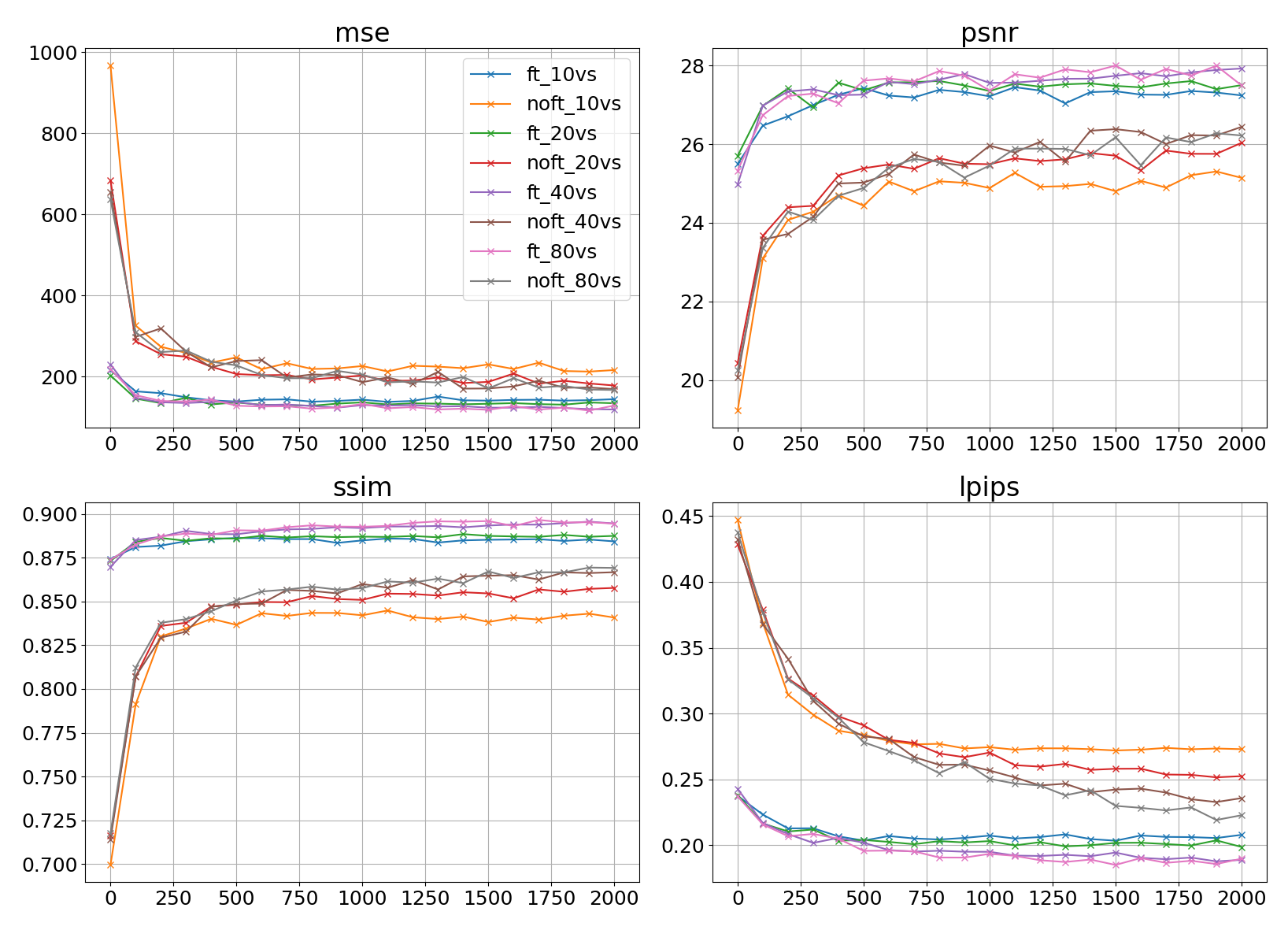}
    \caption{\label{prop_eval:fig_abl_id7}\textbf{Ablation on different finetuning configurations.}
    We show the learning curve of models under different finetuning configurations.
    %
    %ft stands for finetuning a model while noft means training from scratch.
    %
    We finetune(ft) our model as well as train from scratch(noft) with a varied number of training views in $\{10, 20, 40, 80\}$ that are approximately uniformly sampled from all training views.
    Our pre-trained model creates a warm start for avatar personalization and is also robust to the number of views used for finetuning. 
    }
\end{figure}

In Fig~\ref{prop_eval:fig_abl_id7}, we show how MSE, PSNR, SSIM and LPIPS on test views change across optimization steps.
Each curve in Fig~\ref{prop_eval:fig_abl_id7} represents a model with a different finetuning configuration. 
ft stands for finetuning a pre-trained model on the new identity while noft stands for training the same model on the new identity from scratch.
$x$vs indicates that we use a total number of $x$ views to perform the finetuning. 
For example, 10vs means that we use 10 views for finetuning.
To make sure that the finetuning views are not biased to certain viewpoints, we sample the finetuning views from all training views using the furthest point sampling.
As we can see, the pre-trained model could give the finetuning a warm start and leads to better convergence under a short amount of training time, which is also robust to the density of finetuning views.

\begin{figure}[tb!]
\setlength\tabcolsep{0pt}%%
\renewcommand{\arraystretch}{0}%%
\centering
\begin{tabular}{c}
 \begin{tikzpicture}
    \node[anchor=south west,inner sep=0] (image) at (0,0) {\adjincludegraphics[width=0.98\columnwidth, trim={0 0 0 0}, clip]{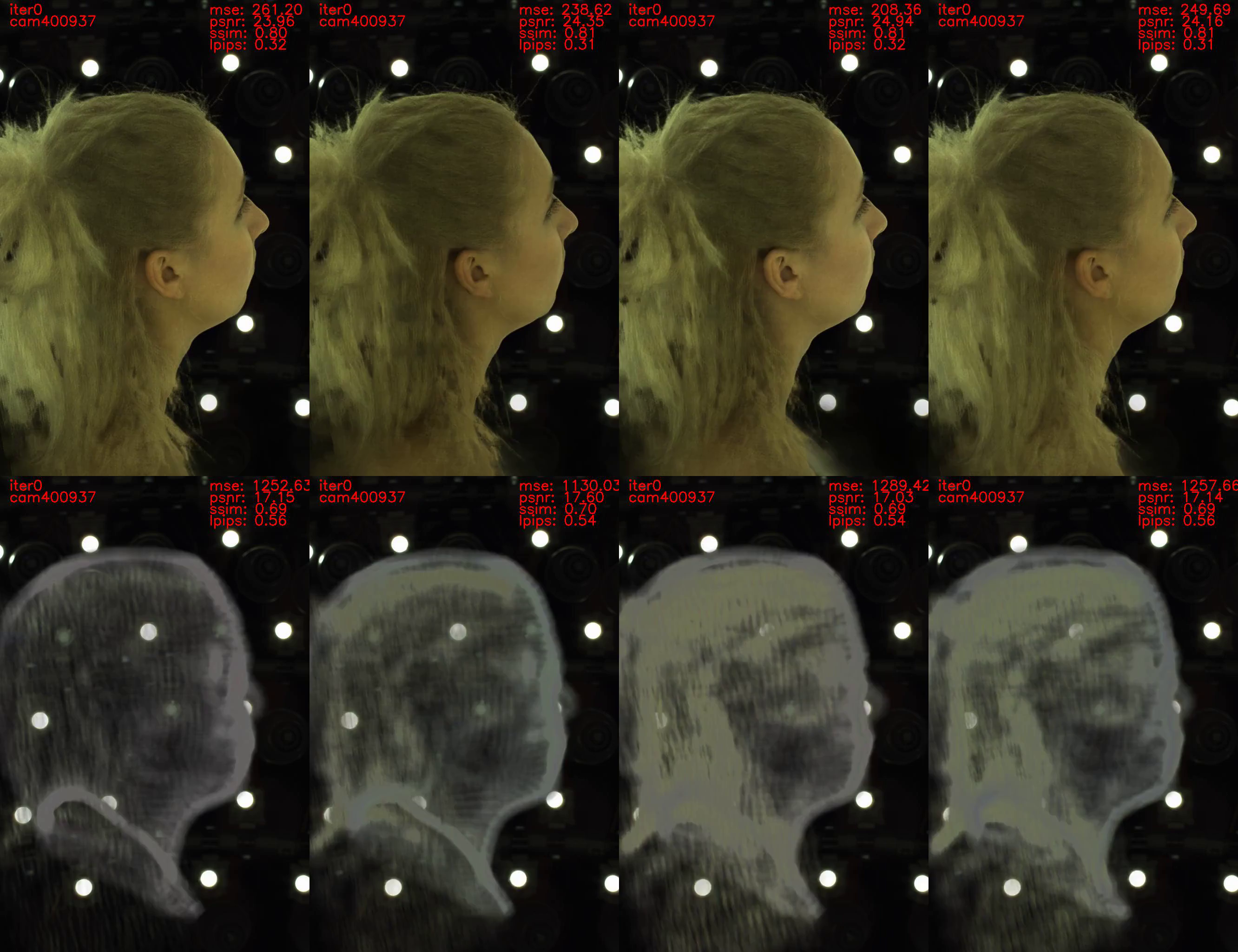}};
 \end{tikzpicture} \\
 %\begin{tikzpicture}
  %  \node[anchor=south west,inner sep=0] (image) at (0,0) {\adjincludegraphics[width=\columnwidth, trim={0 0 0 {0.18\height}}, clip]{figs_props/abl_vs_id7/0001_c-0.png}};
 %\end{tikzpicture} \\
 %\begin{tikzpicture}
  %  \node[anchor=south west,inner sep=0] (image) at (0,0) {\adjincludegraphics[width=\columnwidth, trim={0 {0.03\height} 0 {0.15\height}}, clip]{figs_props/abl_vs_id7/0001_c-1.png}};
 %\end{tikzpicture} \\
 %\begin{tikzpicture}
  %  \node[anchor=south west,inner sep=0] (image) at (0,0) {\adjincludegraphics[width=\columnwidth, trim={0 0 0 {0.18\height}}, clip]{figs_props/abl_vs_id7/0006_c-0.png}};
 %\end{tikzpicture} \\
 %\begin{tikzpicture}
    %\node[anchor=south west,inner sep=0] (image) at (0,0) {\adjincludegraphics[width=\columnwidth, trim={0 0 0 {0.18\height}}, clip]{figs_props/abl_vs_id7/0006_c-1.png}};
    %\node[anchor=south east,inner sep=0] at (current bounding box.south east) {\setlength{\fboxsep}{0pt}\setlength{\fboxrule}{0.2pt}\fcolorbox{red}{white}{\adjincludegraphics[width=0.1\columnwidth, trim={0 0 0 {0.18\height}}, clip]{figs_props/abl_vs_id7/image0007.png}}};
 %\end{tikzpicture} %
 \\
 \begin{tikzpicture}
    \node[anchor=south west,inner sep=0] (image) at (0,0) {\adjincludegraphics[width=0.98\columnwidth, trim={0 0 0 0}, clip]{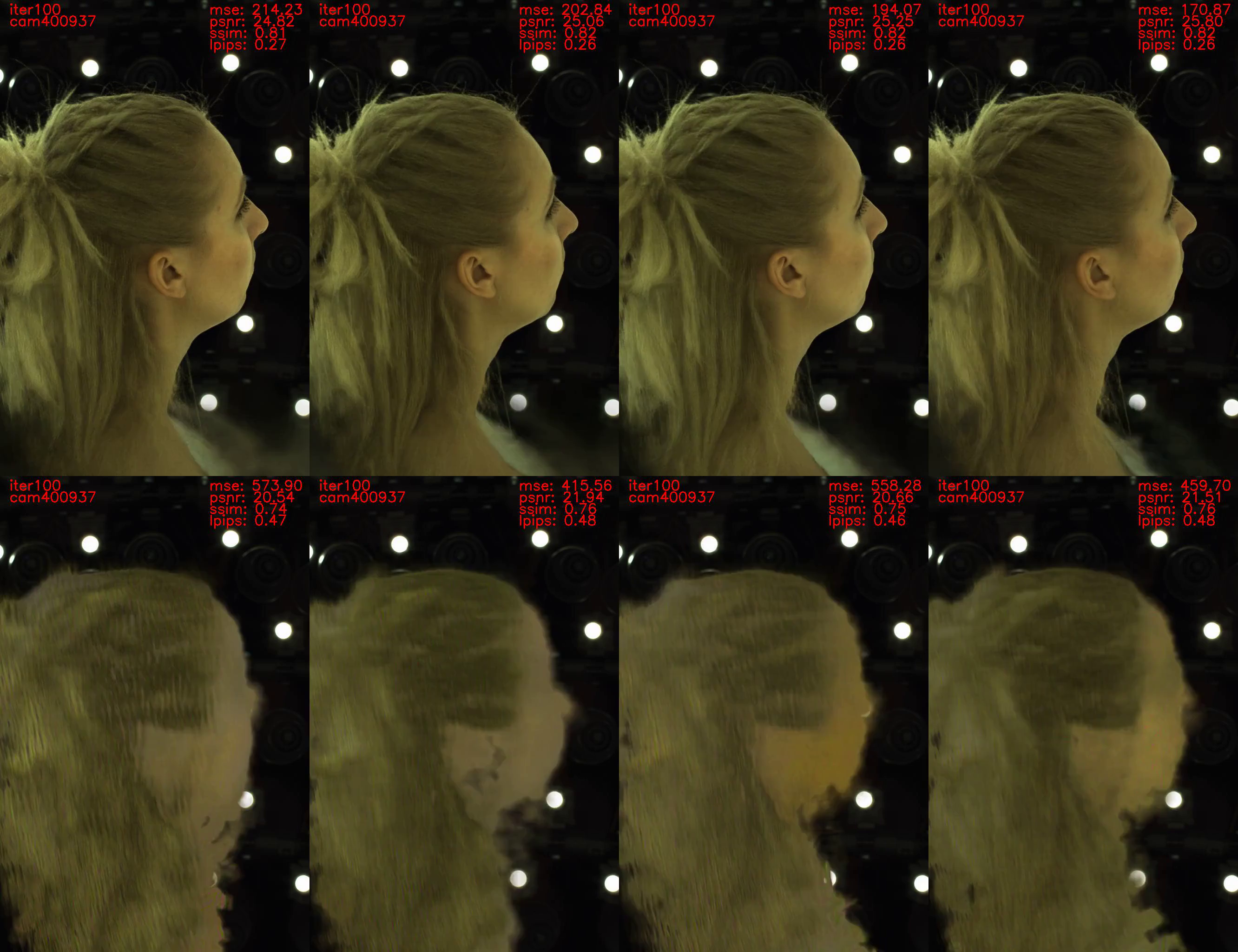}};
    %\node[anchor=south east,inner sep=0] at (current bounding box.south east) {\setlength{\fboxsep}{0pt}\setlength{\fboxrule}{0.2pt}\fcolorbox{red}{white}{\adjincludegraphics[width=0.1\columnwidth, trim={0 0 0 0}, clip]{figs_props/abl_vs_id7/image0007.png}}};
 \end{tikzpicture}
\end{tabular}
\caption{\label{props_eval:abl_iter_id7}\textbf{Rendering results under different finetuning steps and views.}
We show finetuned results under iteration 0 and 100 on the first and third columns respectively and train from scratch results on the second and fourth.
From left to right, the results are from models trained using 10, 20, 40 and 80 views.
In the lower right corner, we show the ground truth image under the rendering view for reference.
}
\end{figure}

In Fig~\ref{props_eval:abl_iter_id7}, we show the rendering results of a test view under different finetuning/training configurations as in Fig.~\ref{prop_eval:fig_abl_id7}.
As we can see, at iteration 0, our model can already reconstruct most details about the new hairstyle and after 100 iterations models trained with different numbers of views gets sharp results, while models trained from scratch have not yet converged.

\begin{figure}[h!]
\setlength\tabcolsep{0pt}%%
\renewcommand{\arraystretch}{0}%%
\centering
\begin{tabular}{ccccc}
 \textbf{\tiny instant-ngp~\cite{mueller2022instant}} & 
 \textbf{\tiny test view} &
 \textbf{\tiny Ours} &
 \textbf{\tiny test view} &
 \textbf{\tiny Ground Truth} \\
 \begin{tikzpicture}
    \node[anchor=south west,inner sep=0] (image) at (0,0) {\adjincludegraphics[width=0.2\columnwidth, trim={{0.\width} {0.294\height} {0.1\width} {0.2\height}}, clip]{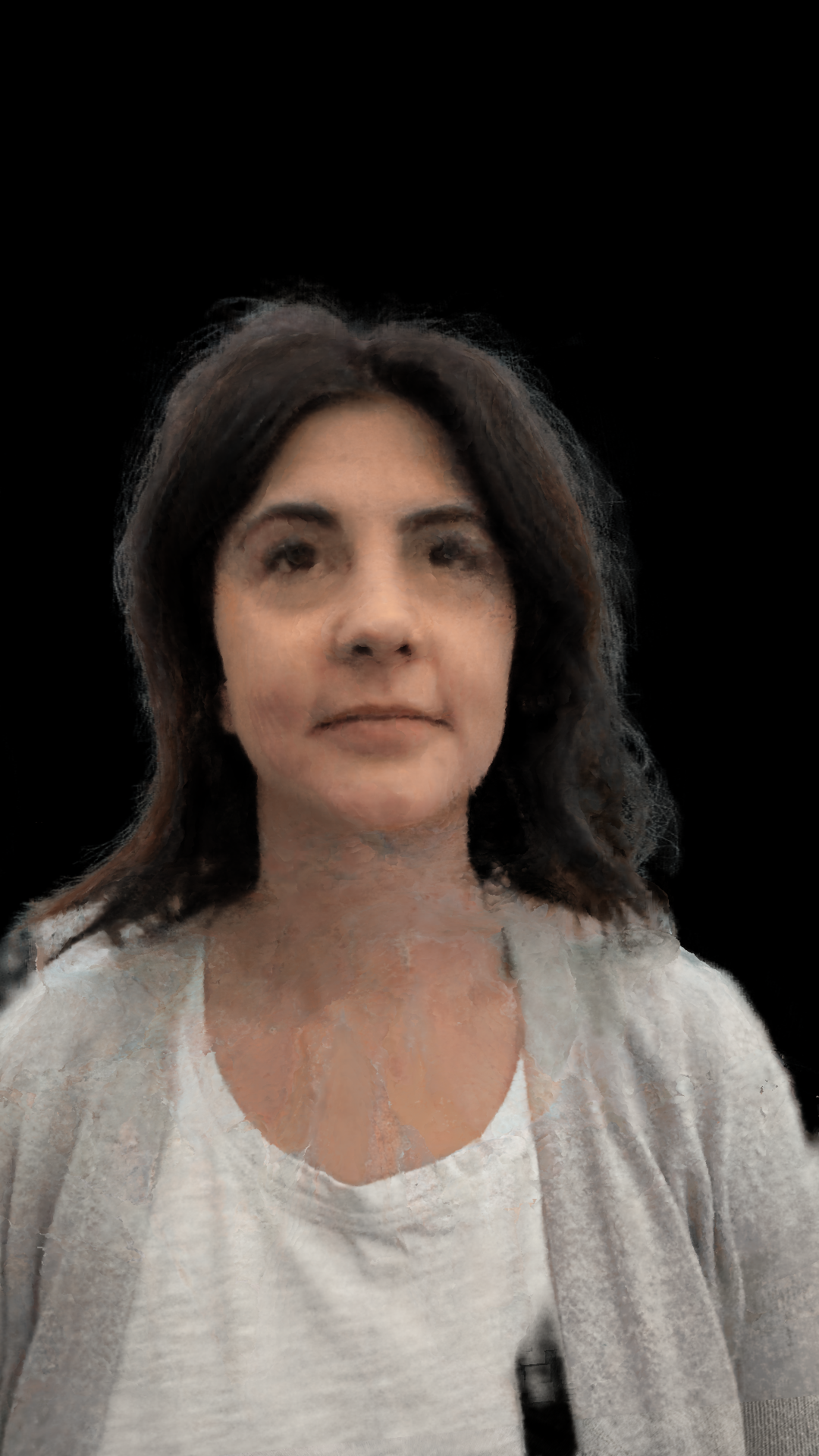}};
 \end{tikzpicture}
&
\begin{tikzpicture}
    \node[anchor=south west,inner sep=0] (image) at (0,0) {\adjincludegraphics[width=0.2\columnwidth, trim={{0.\width} {0.\height} {0.\width} {0.\height}}, clip]{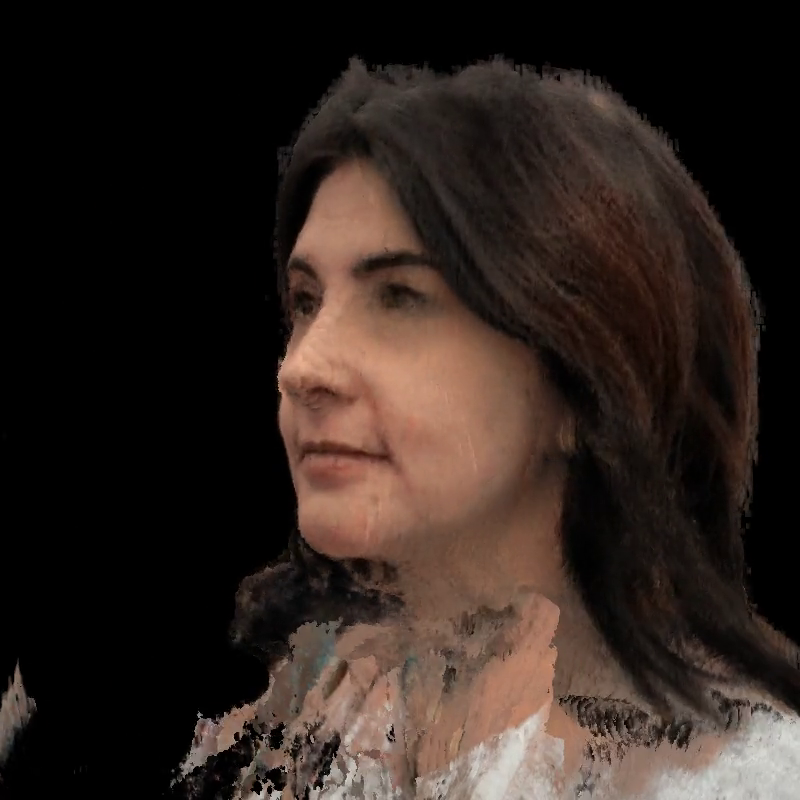}};
 \end{tikzpicture}
 &
\begin{tikzpicture}
    \node[anchor=south west,inner sep=0] (image) at (0,0) {\adjincludegraphics[width=0.2\columnwidth, trim={{0.\width} {0.294\height} {0.1\width} {0.2\height}}, clip]{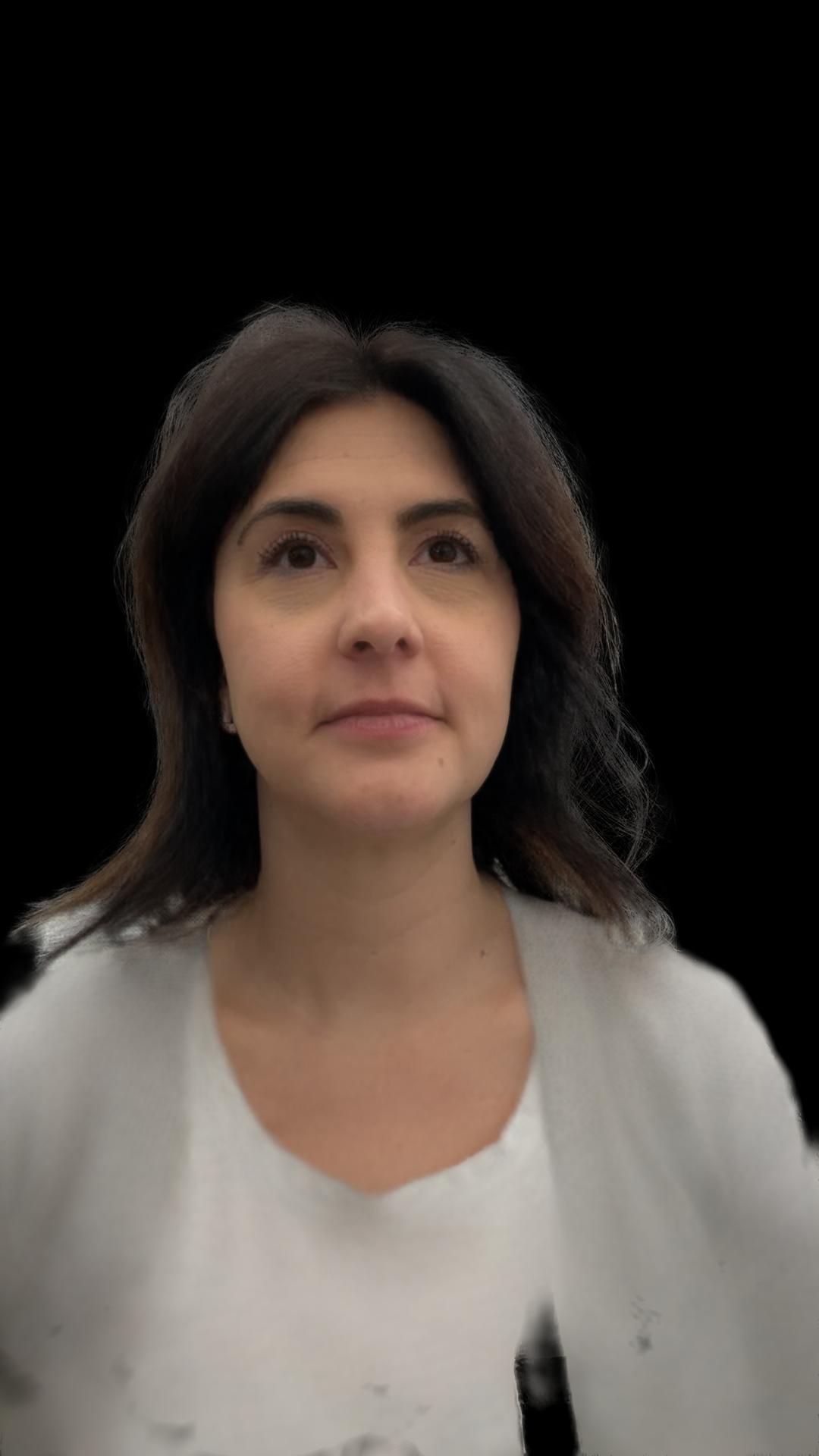}};
 \end{tikzpicture}
 &
\begin{tikzpicture}
    \node[anchor=south west,inner sep=0] (image) at (0,0) {\adjincludegraphics[width=0.2\columnwidth, trim={{0.\width} {0.\height} {0.\width} {0.\height}}, clip]{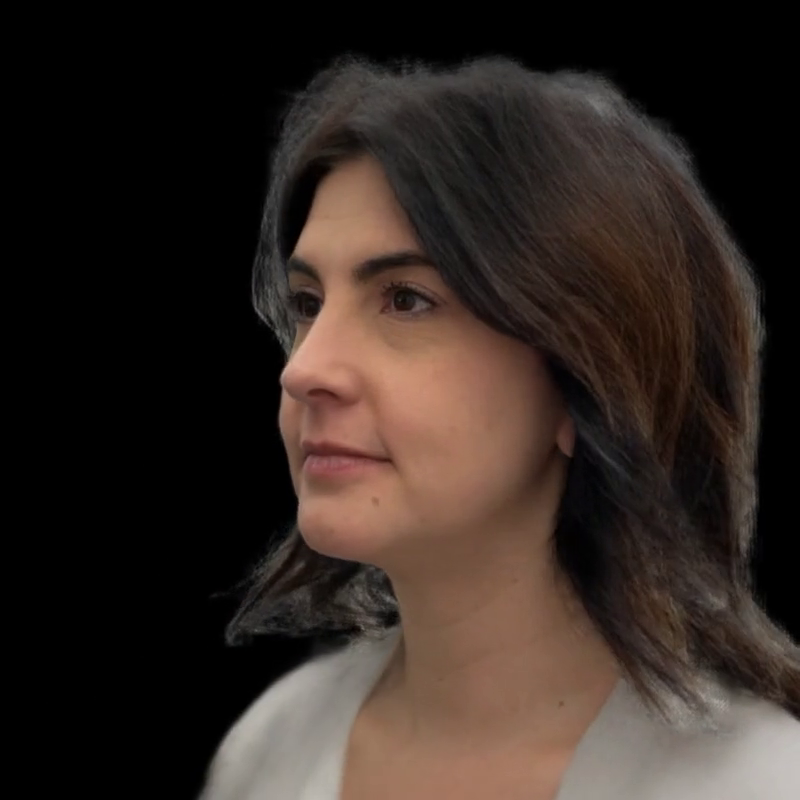}};
 \end{tikzpicture}
&
\begin{tikzpicture}
    \node[anchor=south west,inner sep=0] (image) at (0,0) {\adjincludegraphics[width=0.2\columnwidth, trim={{0.\width} {0.294\height} {0.1\width} {0.2\height}}, clip]{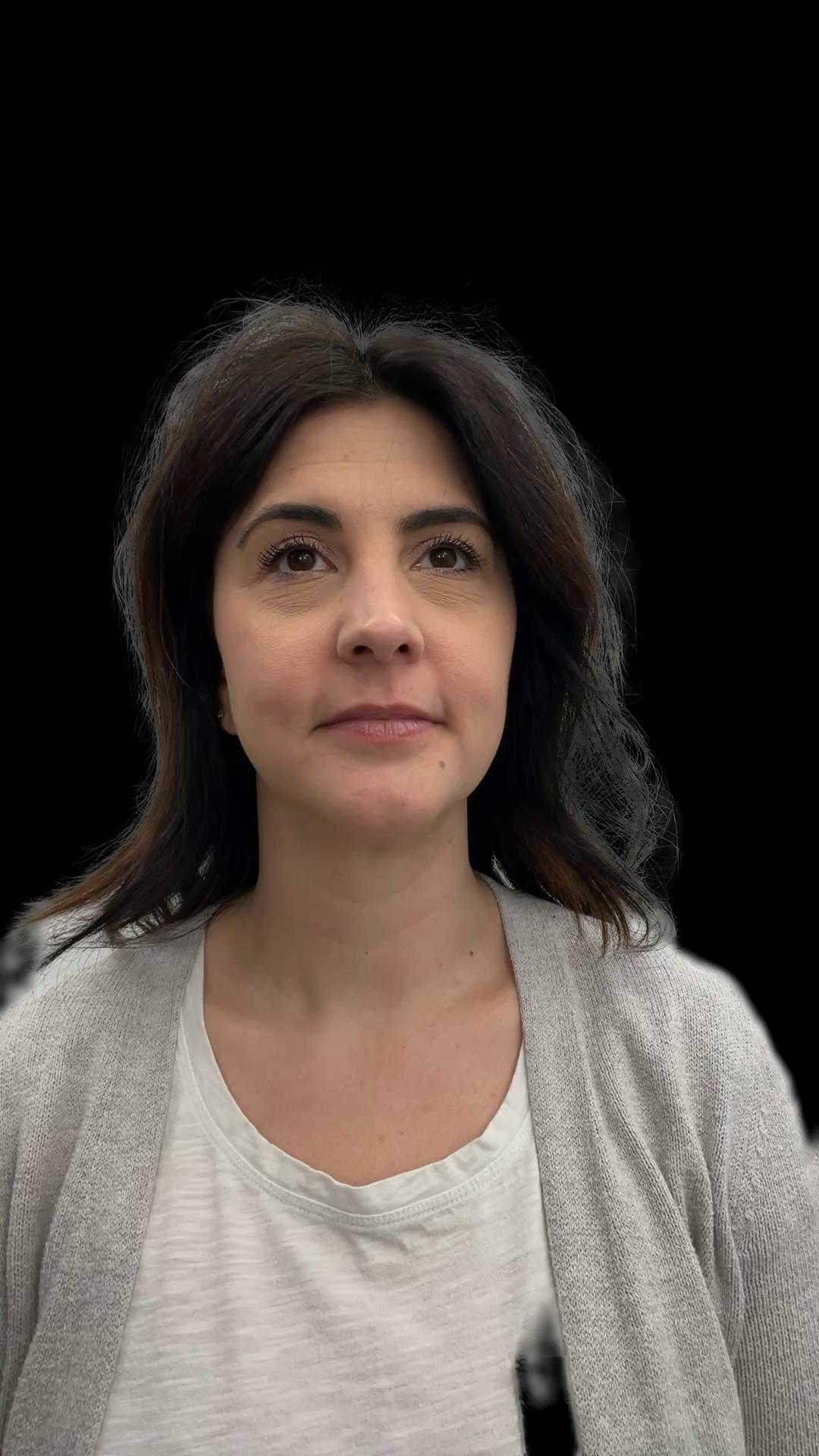}};
 \end{tikzpicture} 
 \\
 \begin{tikzpicture}
    \node[anchor=south west,inner sep=0] (image) at (0,0) {\adjincludegraphics[width=0.2\columnwidth, trim={{0.\width} {0.25\height} {0.1\width} {0.244\height}}, clip]{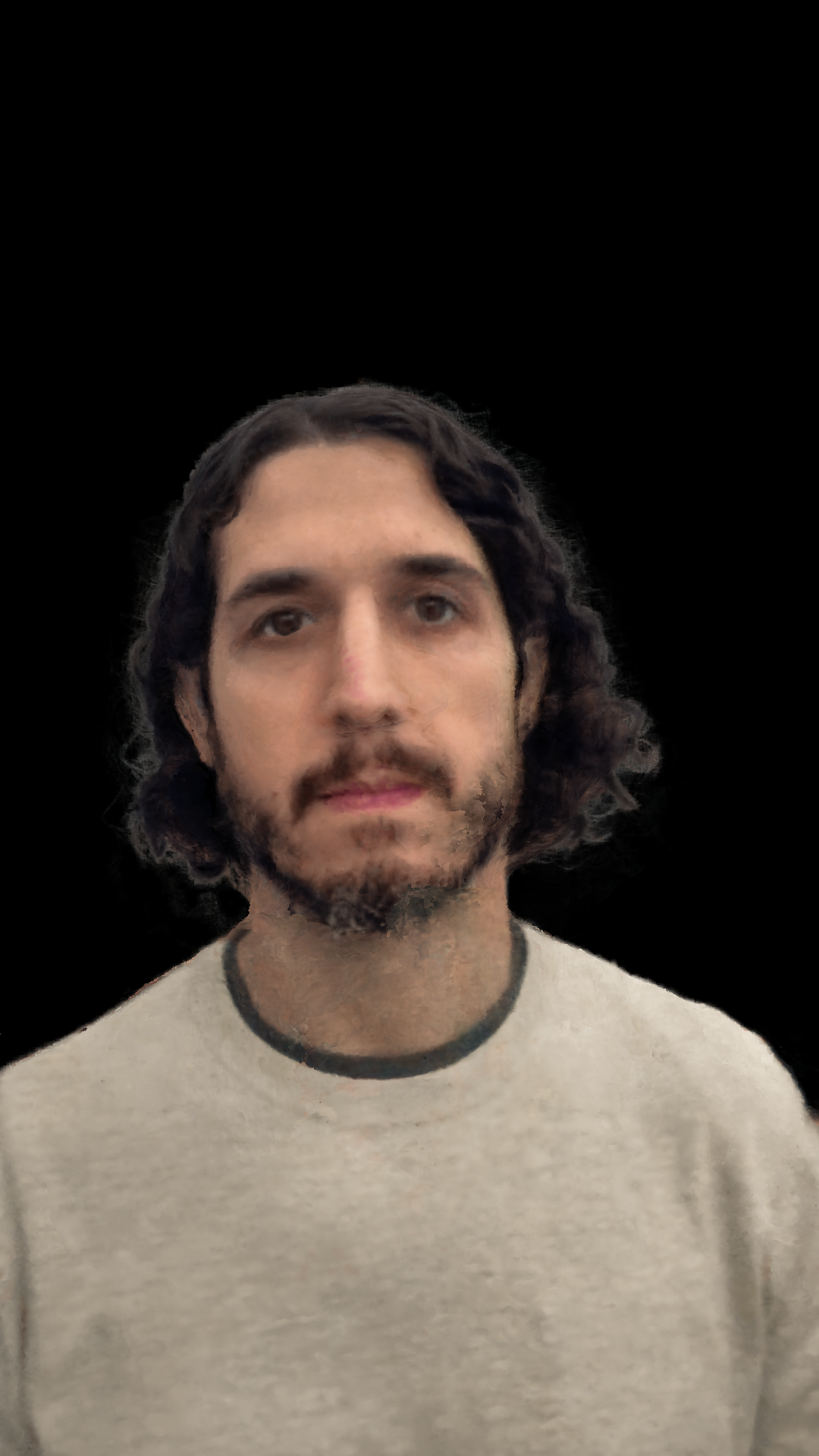}};
 \end{tikzpicture}
&
\begin{tikzpicture}
    \node[anchor=south west,inner sep=0] (image) at (0,0) {\adjincludegraphics[width=0.2\columnwidth, trim={{0.\width} {0.\height} {0.\width} {0.\height}}, clip]{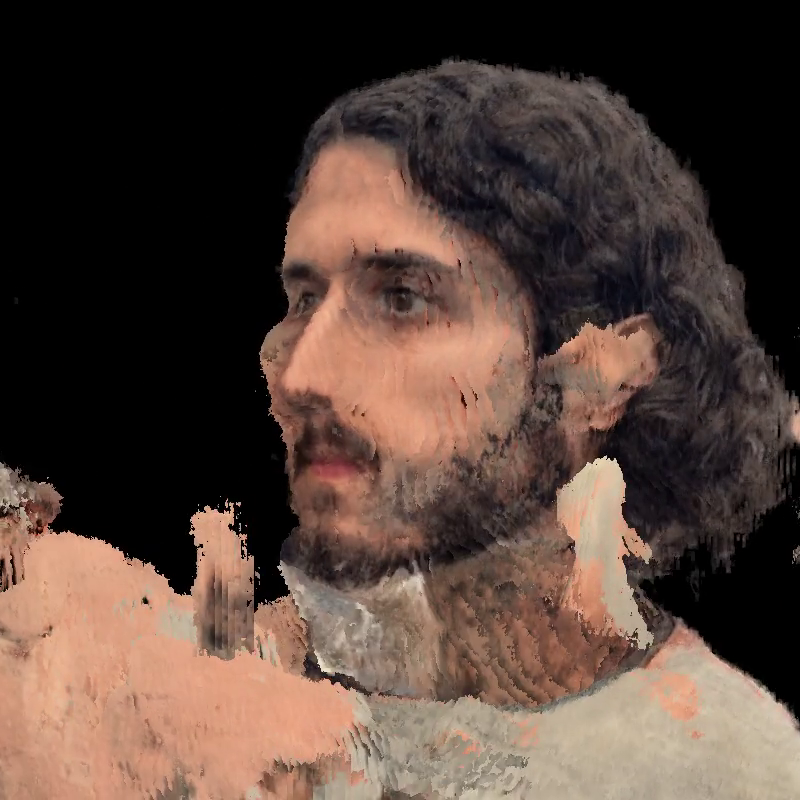}};
 \end{tikzpicture}
&
\begin{tikzpicture}
    \node[anchor=south west,inner sep=0] (image) at (0,0) {\adjincludegraphics[width=0.2\columnwidth, trim={{0.\width} {0.25\height} {0.1\width} {0.244\height}}, clip]{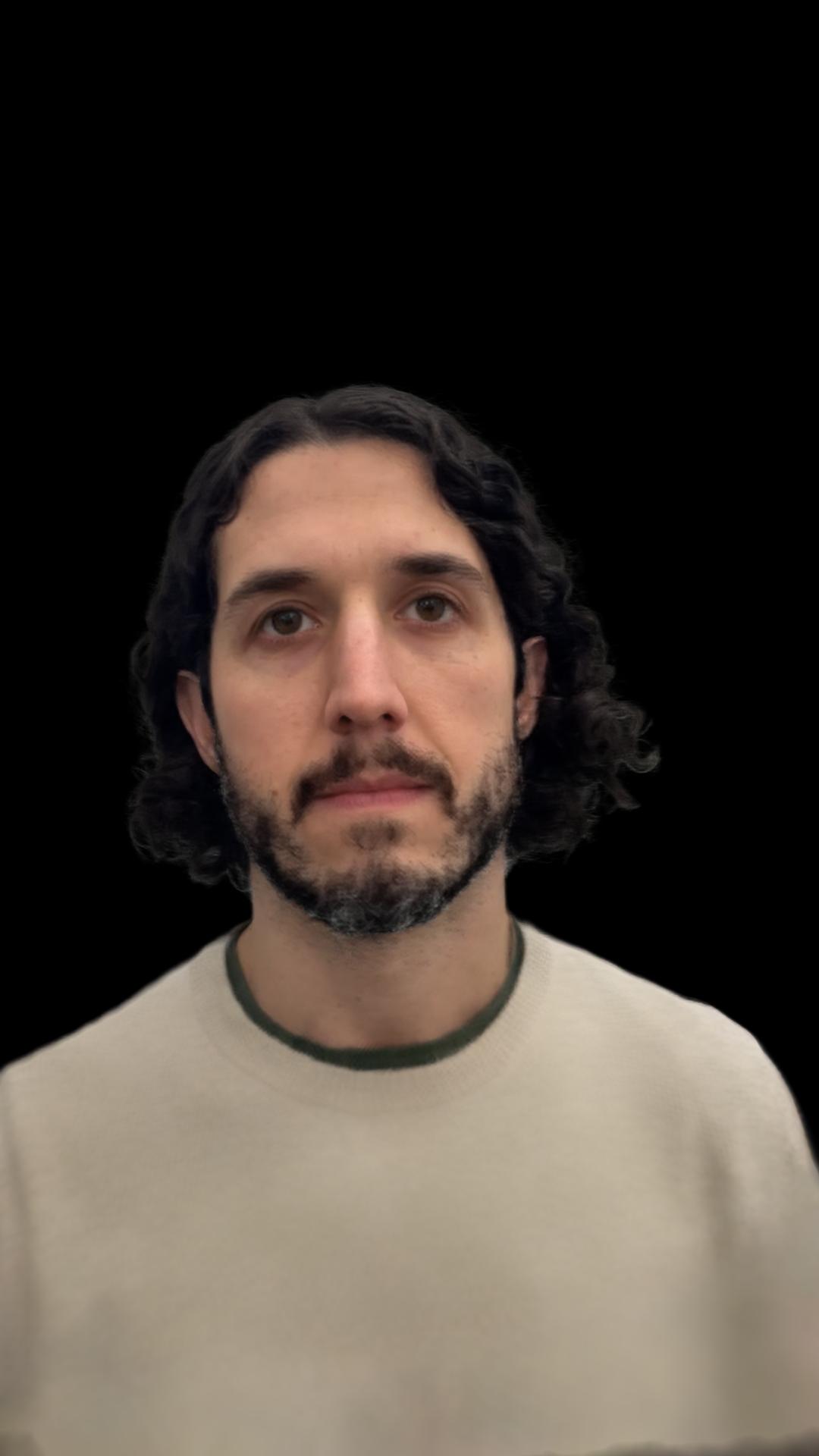}};
 \end{tikzpicture}
 &
\begin{tikzpicture}
    \node[anchor=south west,inner sep=0] (image) at (0,0) {\adjincludegraphics[width=0.2\columnwidth, trim={{0.\width} {0.\height} {0.\width} {0.\height}}, clip]{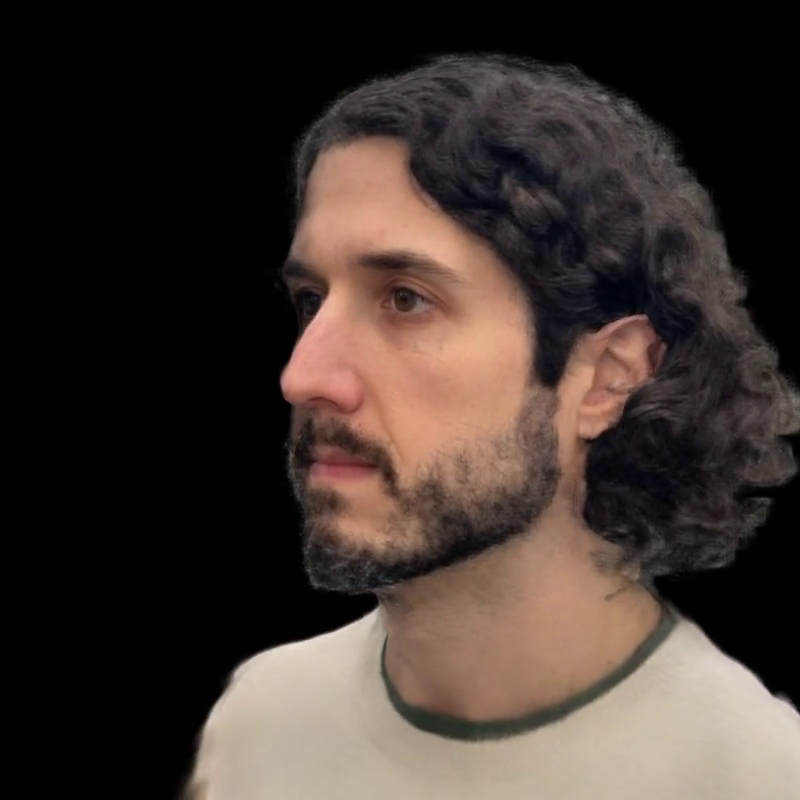}};
 \end{tikzpicture} &
\begin{tikzpicture}
    \node[anchor=south west,inner sep=0] (image) at (0,0) {\adjincludegraphics[width=0.2\columnwidth, trim={{0.\width} {0.25\height} {0.1\width} {0.244\height}}, clip]{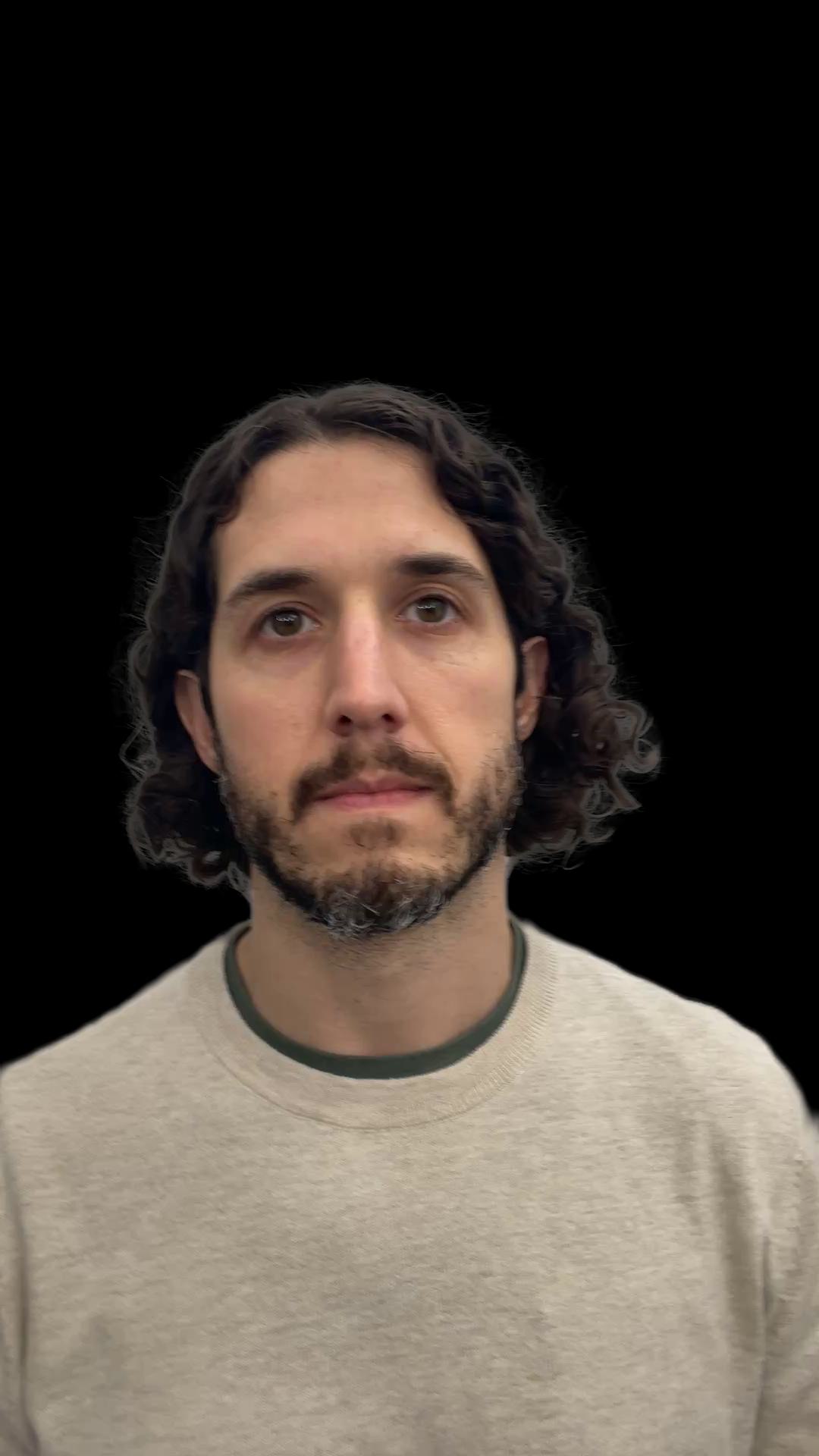}};
 \end{tikzpicture}
 \\\begin{tikzpicture}
    \node[anchor=south west,inner sep=0] (image) at (0,0) {\adjincludegraphics[width=0.2\columnwidth, trim={{0.\width} {0.294\height} {0.1\width} {0.2\height}}, clip]{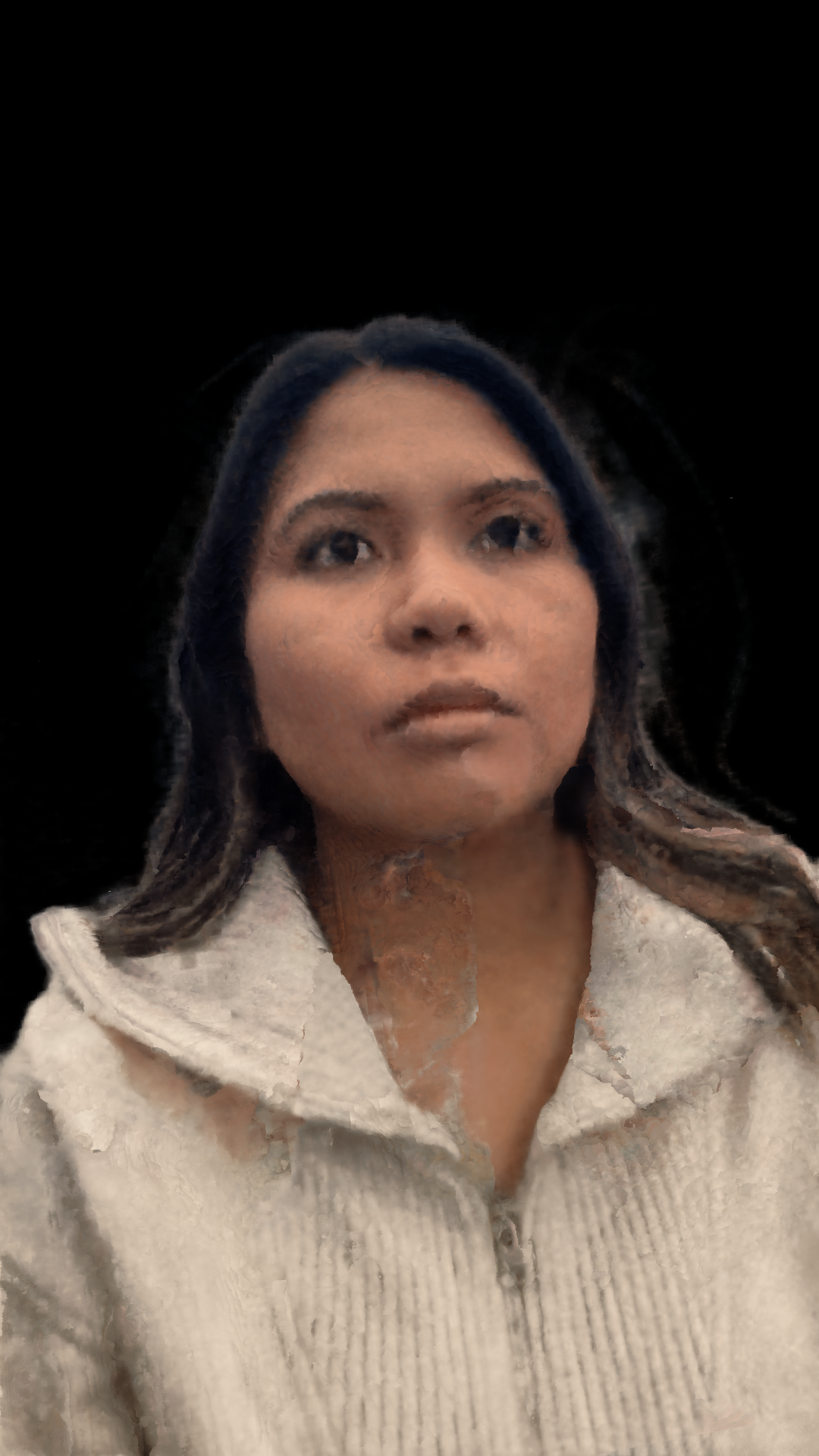}};
 \end{tikzpicture}
&
 \begin{tikzpicture}
    \node[anchor=south west,inner sep=0] (image) at (0,0) {\adjincludegraphics[width=0.2\columnwidth, trim={{0.\width} {0.\height} {0.\width} {0.\height}}, clip]{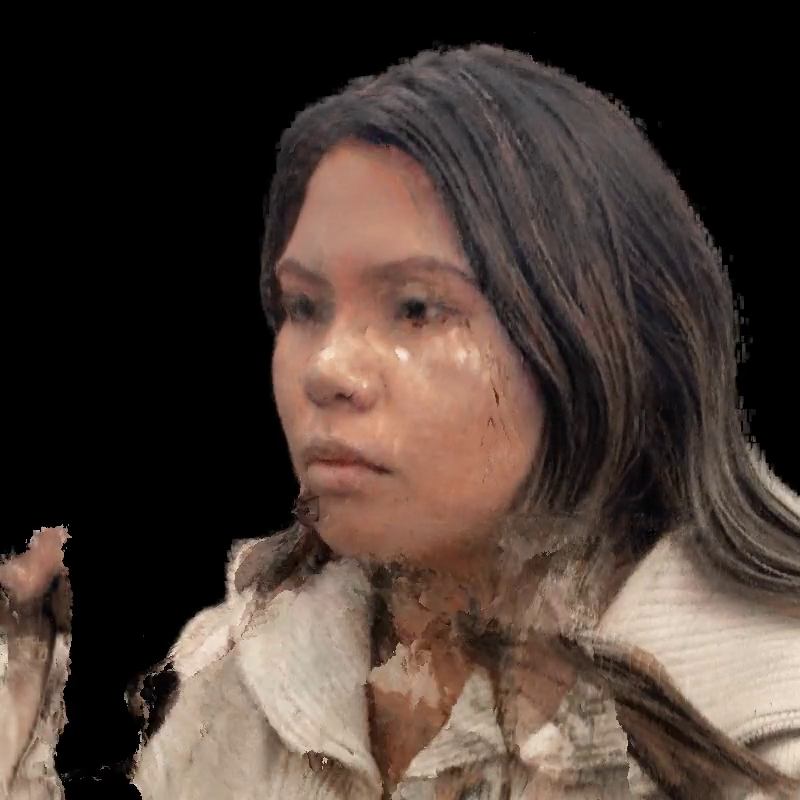}};
 \end{tikzpicture}
&
\begin{tikzpicture}
    \node[anchor=south west,inner sep=0] (image) at (0,0) {\adjincludegraphics[width=0.2\columnwidth, trim={{0.\width} {0.294\height} {0.1\width} {0.2\height}}, clip]{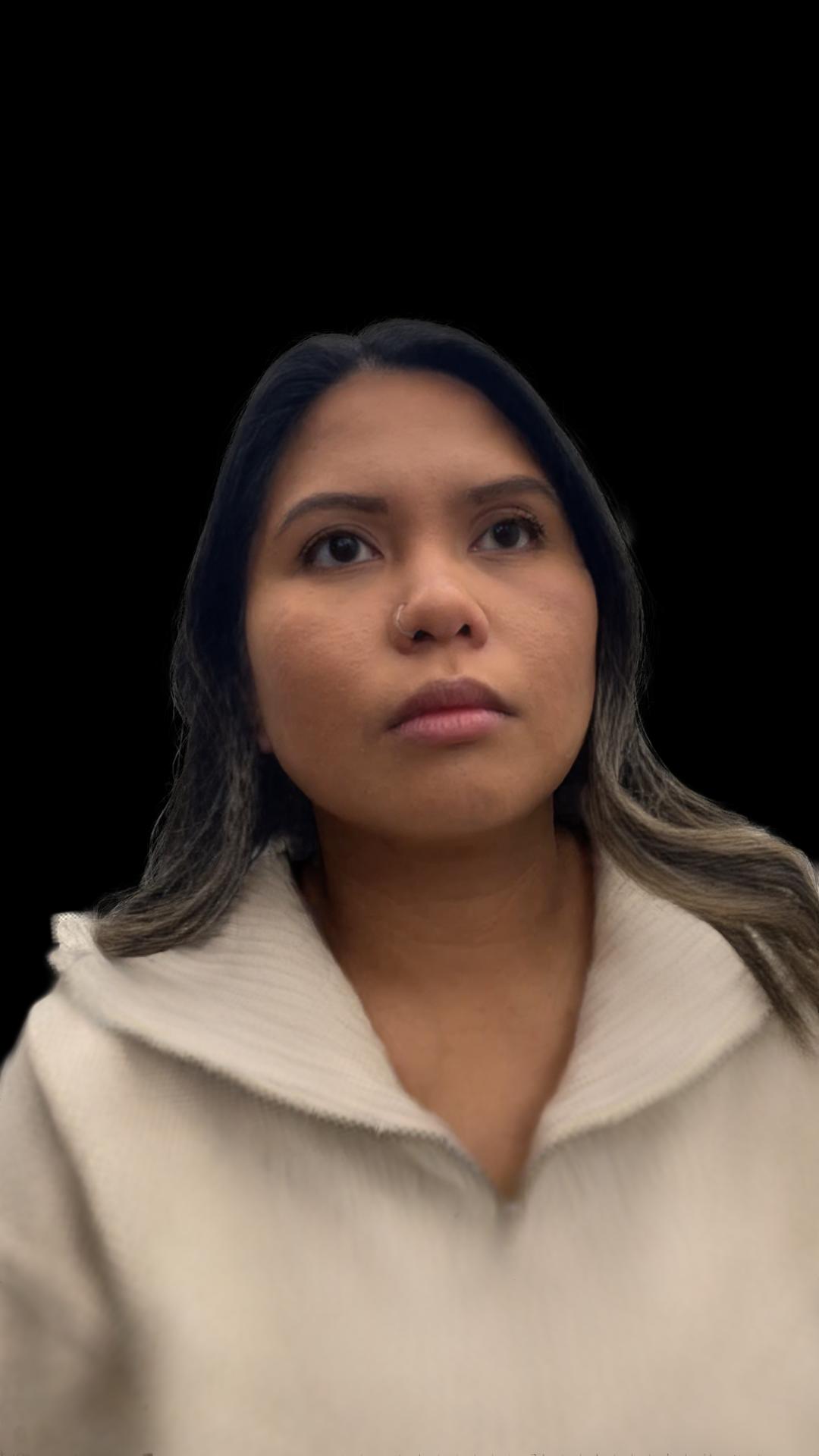}};
 \end{tikzpicture}
 &
\begin{tikzpicture}
    \node[anchor=south west,inner sep=0] (image) at (0,0) {\adjincludegraphics[width=0.2\columnwidth, trim={{0.\width} {0.\height} {0.\width} {0.\height}}, clip]{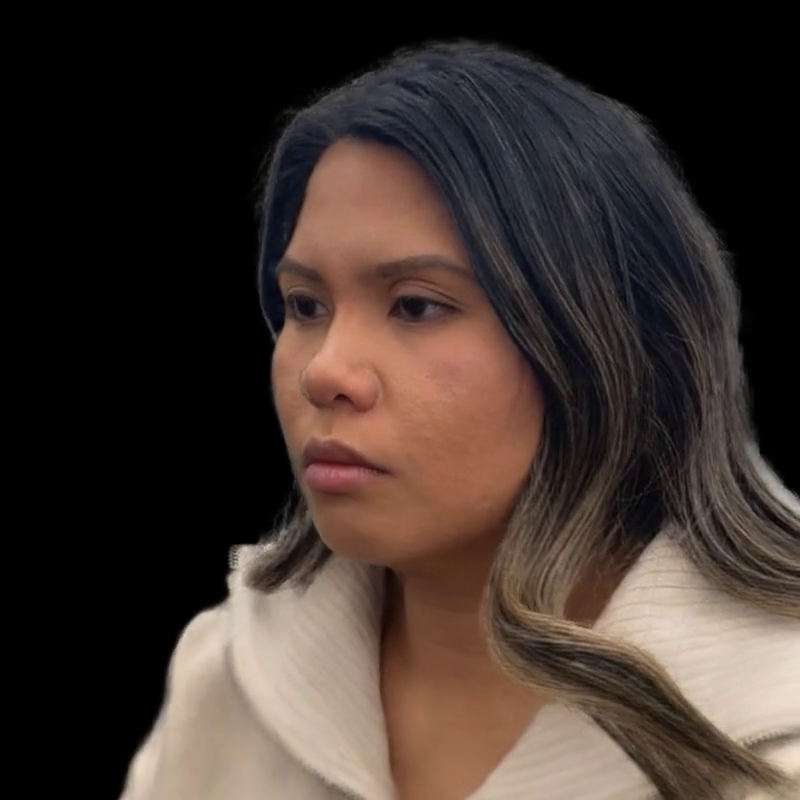}};
 \end{tikzpicture} 
 &
\begin{tikzpicture}
    \node[anchor=south west,inner sep=0] (image) at (0,0) {\adjincludegraphics[width=0.2\columnwidth, trim={{0.\width} {0.294\height} {0.1\width} {0.2\height}}, clip]{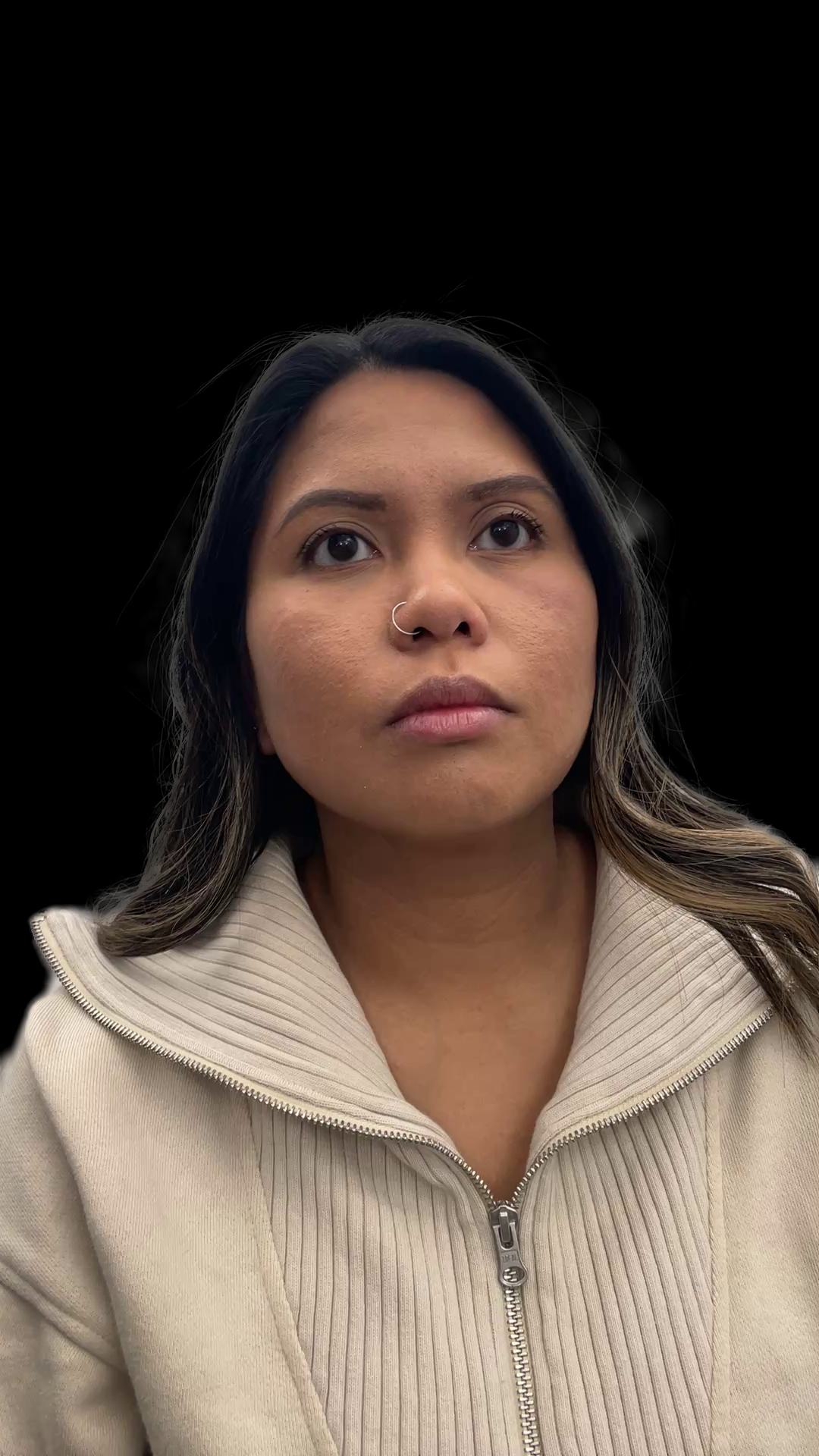}};
 \end{tikzpicture}
 \\
 \begin{tikzpicture}
    \node[anchor=south west,inner sep=0] (image) at (0,0) {\adjincludegraphics[width=0.2\columnwidth, trim={{0.\width} {0.25\height} {0.1\width} {0.244\height}}, clip]{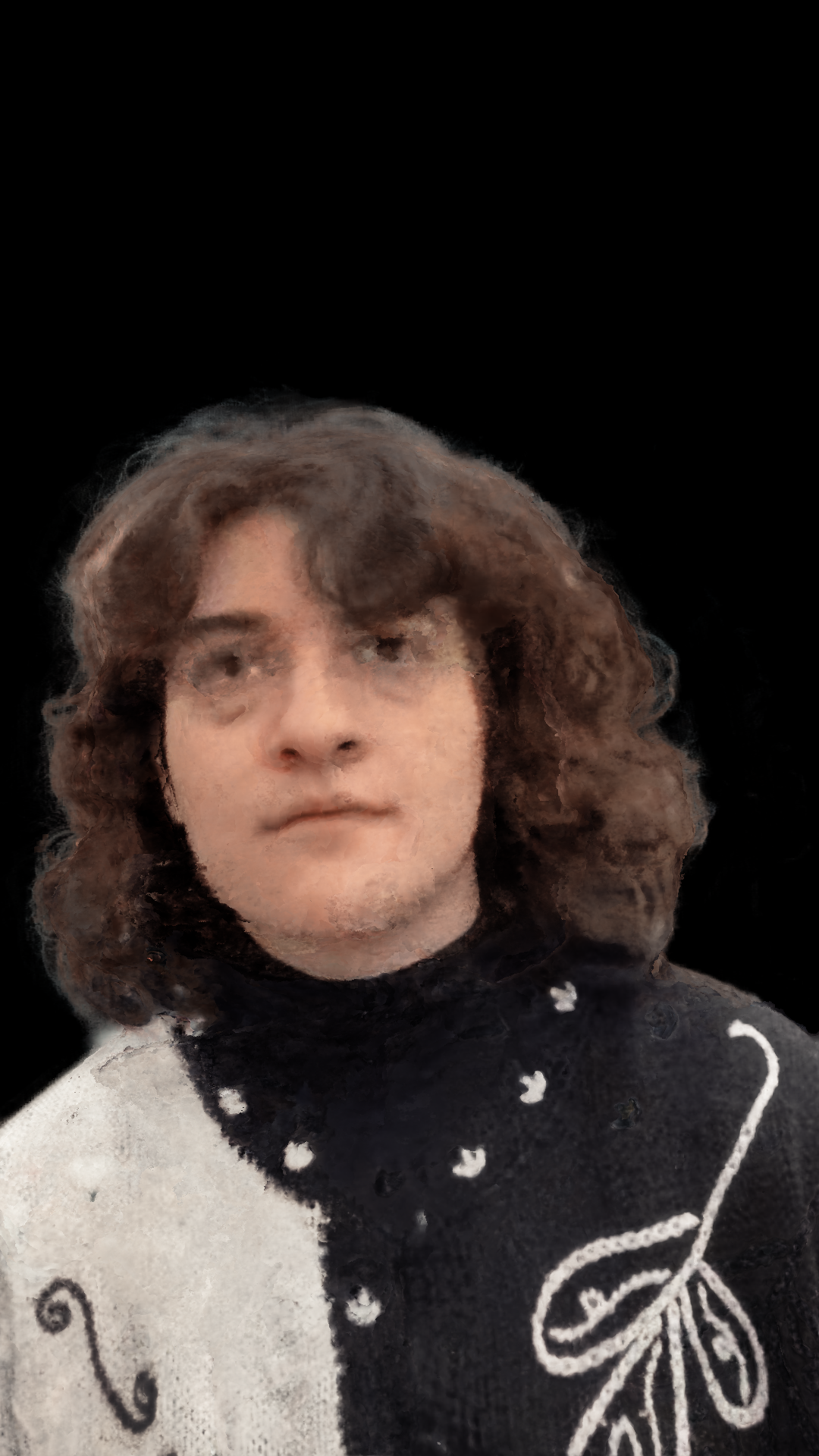}};
 \end{tikzpicture}
 &
 \begin{tikzpicture}
    \node[anchor=south west,inner sep=0] (image) at (0,0) {\adjincludegraphics[width=0.2\columnwidth, trim={{0.\width} {0.\height} {0.\width} {0.\height}}, clip]{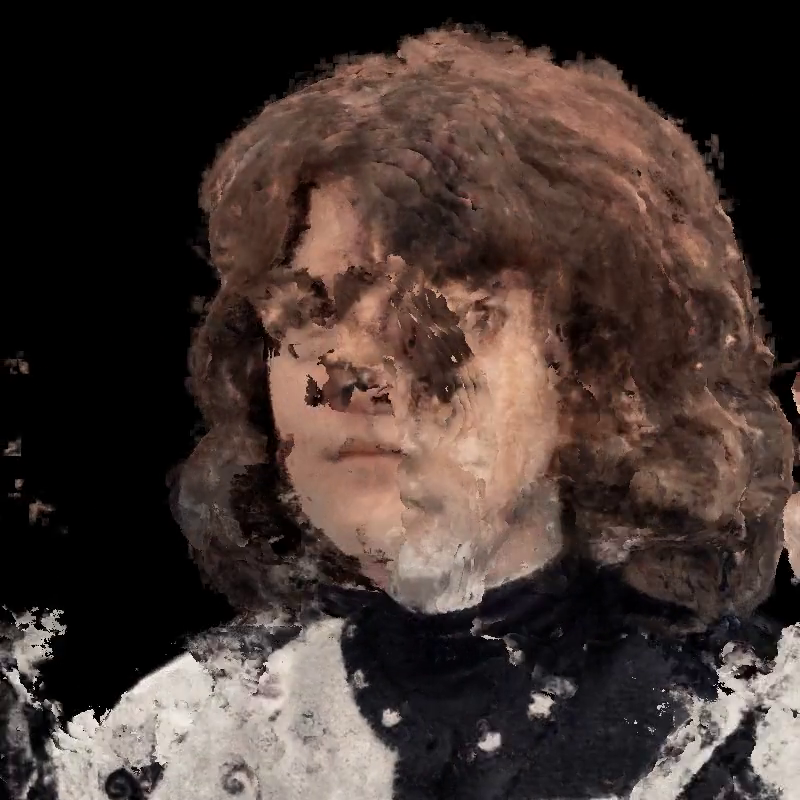}};
 \end{tikzpicture}
&
\begin{tikzpicture}
    \node[anchor=south west,inner sep=0] (image) at (0,0) {\adjincludegraphics[width=0.2\columnwidth, trim={{0.\width} {0.25\height} {0.1\width} {0.244\height}}, clip]{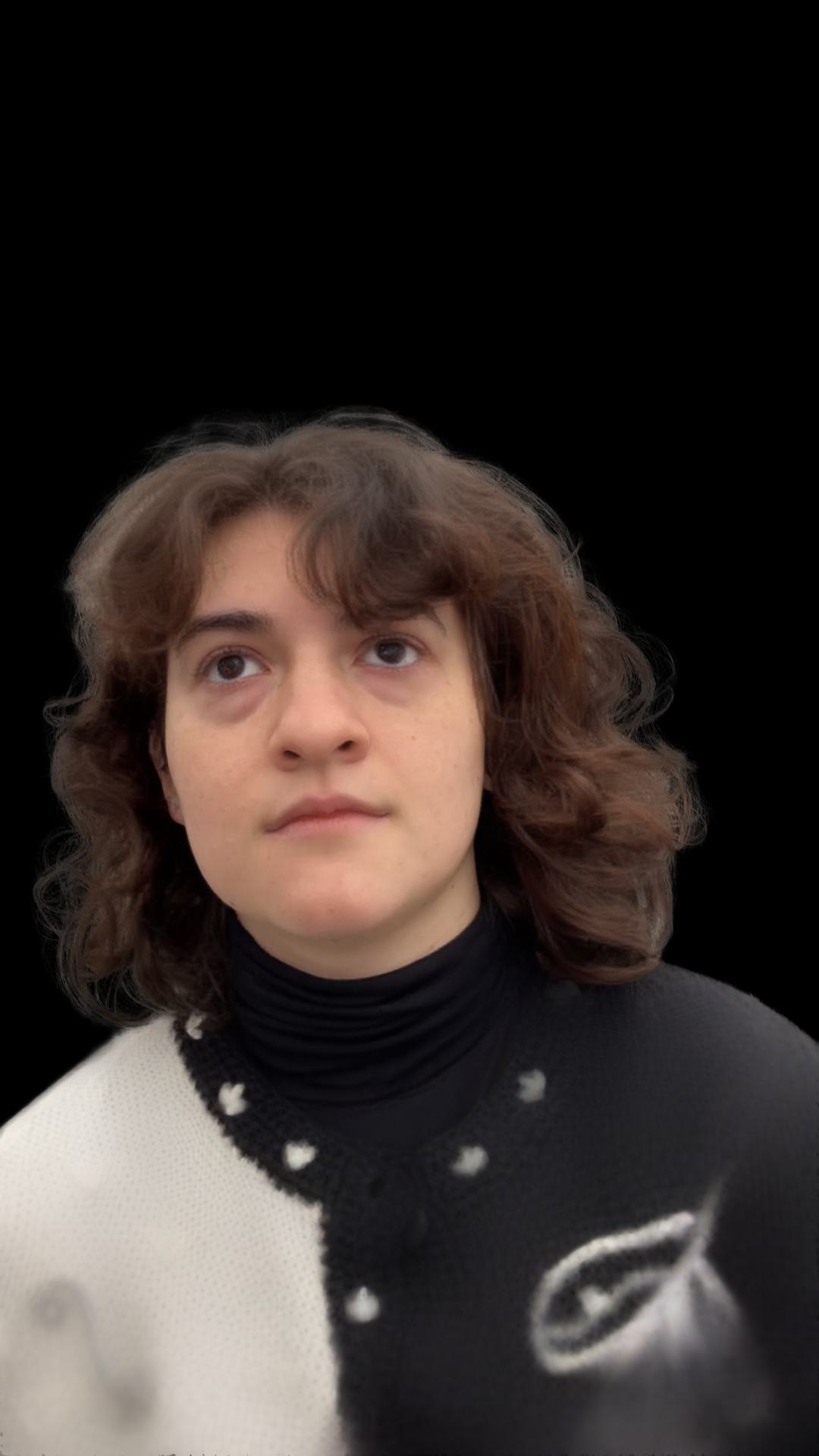}};
 \end{tikzpicture}
 &
\begin{tikzpicture}
    \node[anchor=south west,inner sep=0] (image) at (0,0) {\adjincludegraphics[width=0.2\columnwidth, trim={{0.\width} {0.\height} {0.\width} {0.\height}}, clip]{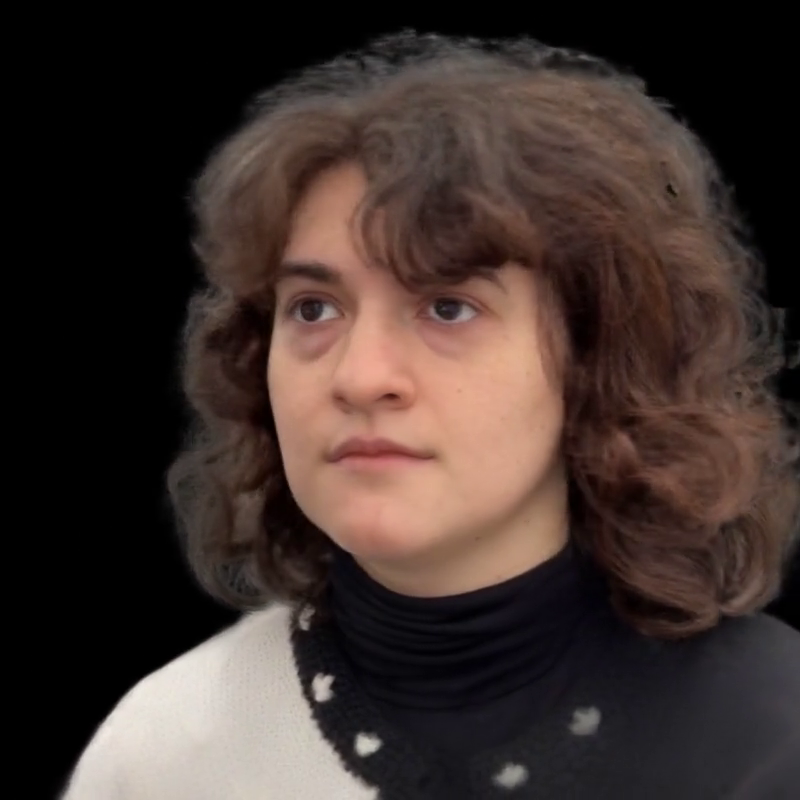}};
 \end{tikzpicture}
&
\begin{tikzpicture}
    \node[anchor=south west,inner sep=0] (image) at (0,0) {\adjincludegraphics[width=0.2\columnwidth, trim={{0.\width} {0.25\height} {0.1\width} {0.244\height}}, clip]{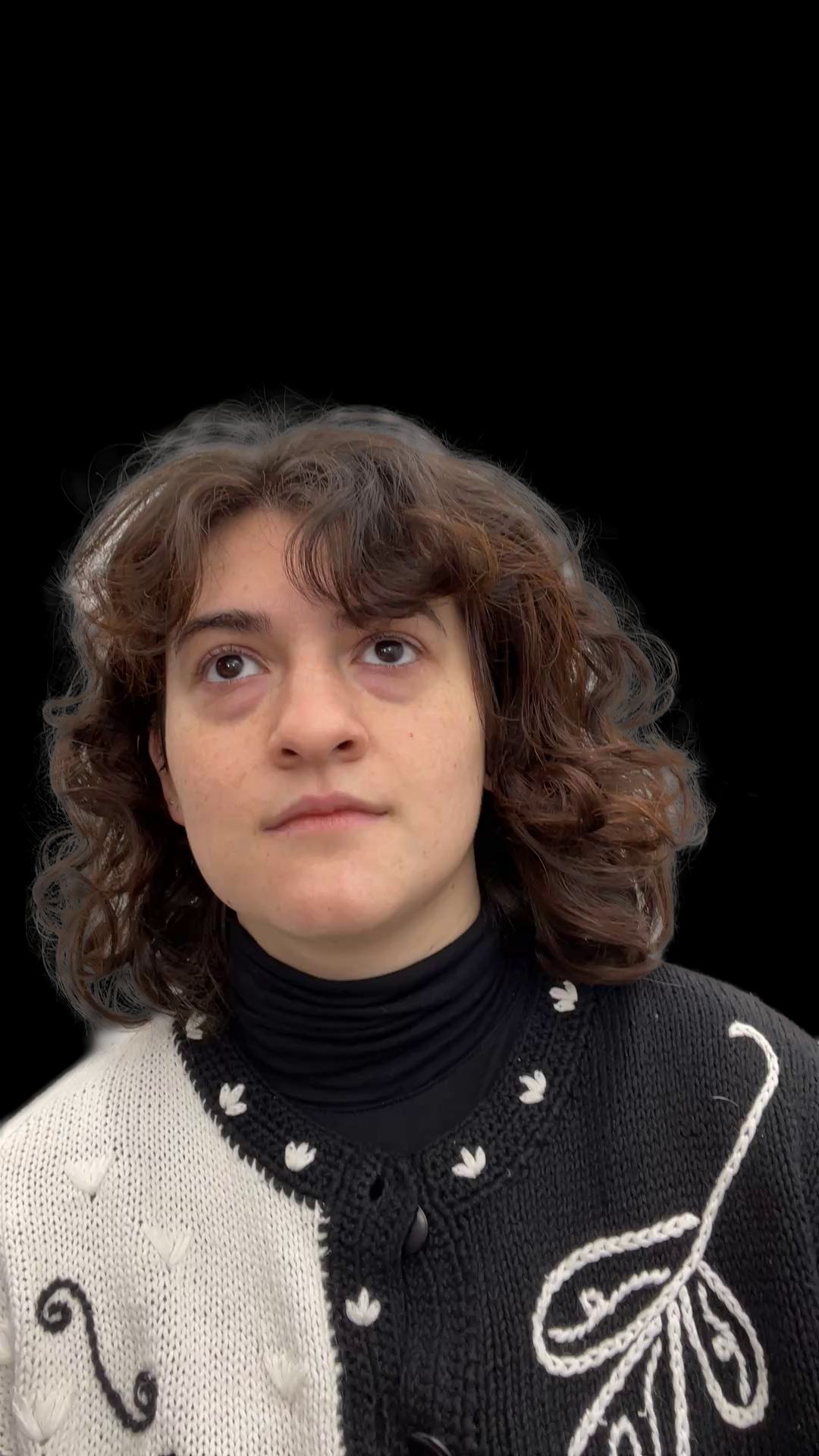}};
 \end{tikzpicture}
\end{tabular}
\caption{\label{prop_eval:fig_nvs_iph_train_eval}\textbf{Rendering results on iPhone captured data.} We show the results of our method and instant-ngp on the iPhone-captured data. Both our method and instant-ngp work well on the training views while our method works better on testing views}
\end{figure}

\subsection{Personalized Avatar from an iPhone Scan}
%\noindent\textbf{Personalized Avatar from an in-the-wild Capture System.}
%
We demonstrate an application enabled by our method, which is the efficient generation of personalized avatars with an iPhone scan.
We used a multi-view camera system to collect the training and validation data for learning the local appearance prior model $\Psi_\alpha$ and $\Psi_{rgb}$.
However, a multi-view capture system is expensive to set up and not readily available to create personalized avatars for individuals.
To accommodate the need for simplicity and scalability, we seek to use a single RGB-D camera (like an iPhone) to substitute for the multi-view capture system. %, which might yield a decayed quality of data.
We demonstrate that our learned model $\Psi_\alpha$ and $\Psi_{rgb}$ can infill the gaps between in-the-wild capture and a lab-level capture accommodate a sparse view and noisy camera tracking setup.

\noindent\textbf{Reconstructing from iPhone Scan.}
We perform further experiments on reconstructing a personalized avatar from an iPhone scan in sparse views.
The data we used is RGB-D data from the iPhone scan. 
We asked the participant to sit still and start our iPhone scan by facing the iPhone towards the participant at a distance of 30-50cm and capturing different perspectives of the participant's face, which results in the participant occupying $70-80\%$ of the whole image. 
The resolution of the captured image is $1024\times667$ and the resolution of the captured depth image is $640\times480$.
%
%The iPhone will first move horizontally to the left side then to the right side and eventually back to the frontal side of the participant's face.
%
%After this, we get a video of the face scan with per-frame depth. 
%
To get the iPhone tracking information, we perform ICP tracking using the per-frame depth as well as color information from the RGB-D scan under different perspectives. %where the first frame depth will be served as the reference frame and loop closure is enforced between the first frame and last frame.
Given the nature of the in-the-wild capture setup, there will be noise in both the depth scan as well as the camera tracking.
We demonstrate that our method is not severely affected by those sources of noise.

We compare our method with instant-ngp~\cite{mueller2022instant} based on the implementation~\cite{torch-ngp} and perform personalized avatar acquisition from the RGB-D scan.
We optimize both instant-ngp and our method using five views with training objectives described in Sec~\ref{prop_work:sec:method}.
To prepare the input to our method, we fuse the RGB-D scans into a point cloud given the ICP tracking results.
We generate the foreground mask using RVM~\cite{lin2022robust} and masked the background in the iPhone captures to be black with the mask.
Both models are optimized for 5min and converge.
%To achieve robust capture from in-the-wild captures, we need a method that is not severely affected by those noises.
%
As shown in Fig.~\ref{prop_eval:fig_nvs_iph_train_eval}, our method and instant-ngp create good-quality rendering on the training views.
However, we also find that instant-ngp yields floating artifacts under the sparse view training setting.
%
%As shown in Fig.~\ref{prop_eval:fig_nvs_iph}, 
When we render from a camera view not in the training set, the quality of instant-ngp decays dramatically.
This result suggests that our method is more robust to sparse views training and camera tracking noise with fewer floating artifacts and suffers from less overfitting on training views.
\section{Conclusion}

In this paper, we develop a method based on compositional volumetric representation for efficient and accurate capturing of human avatars with diverse hairstyles.
Towards that goal, we build a universal hair appearance prior model for modeling the appearance of diverse hairstyles.
To accommodate the large intra-class variance in hair appearance, we split hairstyles into small volumetric primitives and learn a local appearance model that captures the universal appearance prior at that scale.
We empirically show that our model is capable of generating a dense radiance field for a large spectrum of hairstyles with photorealistic appearance and outperforms previous state-of-the-art approaches on both capturing fidelity and generalization.
As a result, our method supports applications like generating personalized avatars from in-the-wild scans using sparse views.
However, our model is limited to static hair and does not support relighting.
How to extend the current model for those purposes are interesting future problems.

{\small
\bibliographystyle{ieeenat_fullname}
\bibliography{egbib}
}

\clearpage
\setcounter{page}{1}
\maketitlesupplementary

\section{Implementation Details on Volumetric Rendering}

\begin{figure}[h!]
    \centering
    \includegraphics[width=\columnwidth]{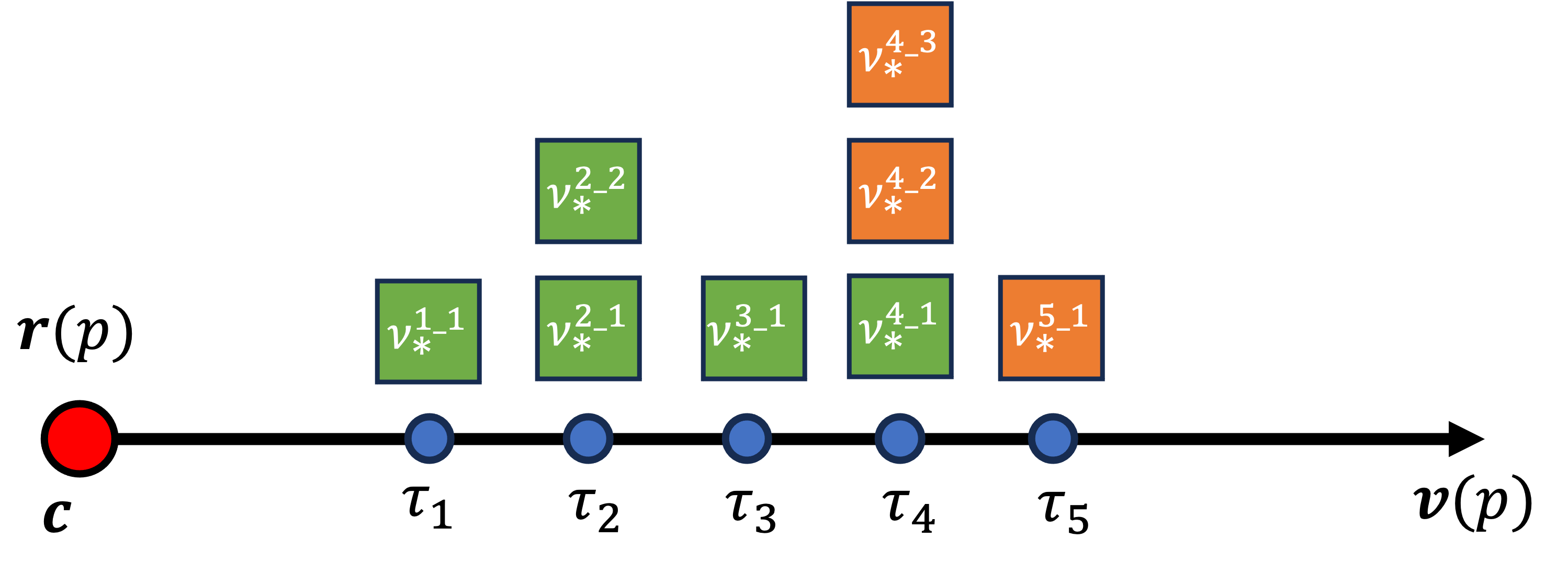}
    \caption{\label{prop_work:supp:rm}\textbf{Raymarching example.}}
\end{figure}

\noindent\textbf{Differentiable Volumetric Raymarching.} Following the formulation in Sec~3.1\iffalse~\ref{prop_work:sec:volrender}\fi, we first explain the implementation details regarding how we aggregate the spatial radiance functions $\mathbf{V}_{rgb}$, $\mathbf{V}_{\alpha}$ and $\mathbf{V}_{label}$ into the renderings as $\mathcal{I}_p$, $\mathcal{M}_p$ and $T(l)$.
We use Riemann sum to approximate the integral in Sec~3.1\iffalse~\ref{prop_work:sec:volrender}\fi.
For simplicity, we use a toy example shown in Figure~\ref{prop_work:supp:rm} for easier illustration.
We denote the red dot $\bm{c}$ as the camera center and the arrow $\bm{v}(p)$ as the raymarching direction of pixel $p$.
We march the ray with uniform step size which results in sample points (blue dots) along the ray with depth $\tau_i$.
And we denote the volumetric primitives that intersect with those sample points as $\nu_{*}^{j}$ where $*$ can be $\alpha$, $rgb$ or $label$ and $j$ is the index of the corresponding volumetric primitive.
The green ones are the primitives for the non-hair region and the orange ones are the primitives for the hair region.
We use the term $\nu_{*}^{j}(\tau_i)$ as the value we sampled from $\nu_{*}^{j}$ at point $\tau_i$.
To get $\nu_{*}^{j}(\tau_i)$, we use trilinear interpolation between the nearest grid points of $\tau_i$ in $\nu_{*}^{j}$.
As in MVP, the aggregated $\alpha$ value up to $\tau_i$ along the ray is computed as below
\begin{align*}
    T(\tau_i) = min(1, \sum_{j=1}^{i}\sum_{k\in j_{k}}\nu_{\alpha}^{j\_k}),
\end{align*}
where $j_k$ is the set for all the indices of the intersected primitives at $\tau_j$.
Supposing that all the $\alpha$ values adds up to a value greater than 1 at $\nu_{\alpha}^{4\_1}$, we will have
\begin{align*}
    T(\tau_1) &= \nu_\alpha^{1\_1} \\
    T(\tau_2) &= T(\tau_1) + \nu_\alpha^{2\_1} + \nu_\alpha^{2\_2} \\
    T(\tau_3) &= T(\tau_2) + \nu_\alpha^{3\_1} \\
    T(\tau_4) &= 1 \\
    T(\tau_5) &= 1,
\end{align*}
where $T(\tau_4)$ and $T(\tau_5)$ will be constant. Thus, we will have the aggregated $rgb$ and $label$ values in MVP~\cite{steve_mvp} as
\begin{align*}
    \mathcal{I}_p^{non-soft} &= \nu_\alpha^{1\_1}\nu_{rgb}^{1\_1} + \nu_\alpha^{2\_1}\nu_{rgb}^{2\_1} + 
    \nu_\alpha^{2\_2}\nu_{rgb}^{2\_2} \\
    &+ \nu_\alpha^{3\_1}\nu_{rgb}^{3\_1} + 
    (1 - T(\tau_3))\nu_{rgb}^{4\_1}\\
    \mathcal{M}_p^{non-soft} &= 0,
\end{align*}
where the above equation will yield 0 gradient with respect to $\nu_*^{4\_2}$, $\nu_*^{4\_3}$ and $\nu_*^{5\_1}$. 
As in NeuWigs~\cite{wang2023neuwigs}, the aggregated label value will be zero in this case as $T(\cdot)$ already saturates at $\nu_\alpha^{4\_1}$ and will not keep aggregating values from $\nu_{*}^{4\_2}$ and $\nu_{*}^{4\_3}$ even though they are at the same point.
This early termination of raymarching is guided by the stochasticity in primitive sorting and is especially problematic if pixel $p$ is on the hair region.
In this case, the label loss will not backpropagate any useful gradients to update $\nu_\alpha^i$ to the correct value.
To fix that problem, we make the saturation point to be better aware of the other intersecting boxes with a new soft blending formulation.
Instead of taking the RGB and label value of the very first box $\tau_4$ intersects, we compute the RGB and label at that point as
\begin{align*}
    \mathcal{I}_p^{soft} &= \nu_\alpha^{1\_1}\nu_{rgb}^{1\_1} + \nu_\alpha^{2\_1}\nu_{rgb}^{2\_1}\\
    &+ \nu_\alpha^{2\_2}\nu_{rgb}^{2\_2} + \nu_\alpha^{3\_1}\nu_{rgb}^{3\_1}\\
    &+ (1 - T(\tau_3)) \frac{\sum_{i=1}^{3}\nu_\alpha^{4\_i}\nu_{rgb}^{4\_i}}{\sum_{i=1}^{3}\nu_\alpha^{4\_i}} \\
    \mathcal{M}_p^{soft} &= (1 - T(\tau_3)) \frac{\sum_{i=2}^{3}\nu_\alpha^{4\_i}}{\sum_{i=1}^{3}\nu_\alpha^{4\_i}}.
\end{align*}
With a soft blending reformulation, the rendering terms $\mathcal{I}_p^{soft}$ and $\mathcal{M}_p^{soft}$ are both aware of all the primitives intersected at the saturation point and the rendering formulation is no longer affected by the stochasticity in the primitive sorting at the same point.
The new formulation with soft blending is more robust to different initialization on the poses of hair and non-hair primitives.
Specifically, we write the gradients of each $\nu_{*}^{4\_1}$ below,
\begin{align*}
    \frac{\partial \mathcal{I}_p^{non-soft}}{\partial \nu_{\alpha}^{4\_1}} &= 0 \\
    \frac{\partial \mathcal{I}_p^{non-soft}}{\partial \nu_{rgb}^{4\_1}} &= 1 - T(\tau_3) \\
    \frac{\partial \mathcal{M}_p^{non-soft}}{\partial \nu_{\alpha}^{4\_1}} &= 0 \\
    \frac{\partial \mathcal{I}_p^{soft}}{\partial \nu_{\alpha}^{4\_1}} &= (1 - T(\tau_3))\frac{\sum_{i=2}^3(\nu_{rgb}^{4\_1}-\nu_{rgb}^{4\_i})\nu_\alpha^{4\_i}}{(\sum_{i=1}^{3}\nu_\alpha^{4\_i})^2} \\
    \frac{\partial \mathcal{I}_p^{soft}}{\partial \nu_{rgb}^{4\_1}} &= (1 - T(\tau_3))\frac{\nu_\alpha^{4\_1}}{\sum_{i=1}^{3}\nu_\alpha^{4\_i}} \\
    \frac{\partial \mathcal{M}_p^{soft}}{\partial \nu_{\alpha}^{4\_1}} &= -(1 - T(\tau_3))\frac{1}{(\sum_{i=1}^{3}\nu_\alpha^{4\_i})^2},
\end{align*}
where you can see the soft blending version has a more meaningful gradient where $\frac{\partial \mathcal{I}_p^{soft}}{\partial \nu_{\alpha}^{4\_1}}$ and $\frac{\partial \mathcal{M}_p^{soft}}{\partial \nu_{\alpha}^{4\_1}}$ are not zero but are jointly determined by all primitives at $\tau_4$. Given the symmetry between $\nu_{*}^{4\_i}$, the gradients of $\nu_{*}^{4\_2}$ and $\nu_{*}^{4\_3}$ are the same to $\nu_{*}^{4\_1}$ except that 
\begin{align*}
    \frac{\partial \mathcal{M}_p^{soft}}{\partial \nu_{\alpha}^{4\_2}} &= (1 - T(\tau_3))\frac{\nu_\alpha^{4\_1}}{(\sum_{i=1}^{3}\nu_\alpha^{4\_i})^2}.
\end{align*}
Given the gradients $\frac{\partial \mathcal{M}_p^{soft}}{\partial \nu_{\alpha}^{4\_i}}$, we will find that the primitives with a zero label will be updated inverse proportional to how $\mathcal{M}_p^{soft}$ changes while primitives with a non-zero label will change proportionally with respect to $\mathcal{M}_p^{soft}$.

\section{Training details}

We formulate the training objective $\mathcal{L}$ as below:
\begin{align*}
    \mathcal{L} = \mathcal{L}_1 + \lambda_{VGG}\mathcal{L}_{VGG} + \lambda_{seg}\mathcal{L}_{seg},
\end{align*}
\noindent where $\lambda_{VGG}$ and $\lambda_{seg}$ are positive values for rebalancing each term in the training objectives.
The first term $\mathcal{L}_1$ measures the difference between the rendered image $\mathcal{\tilde{I}}$ and the ground truth image $I_{gt}$:
\begin{align*}
    \mathcal{L}_1 = ||\mathcal{\tilde{I}} - I_{gt}||_1.
\end{align*}
To enhance the rendering fidelity and achieve better convergence on $\mathcal{L}_1$, we add a second term of perceptual loss as
\begin{align*}
    \mathcal{L}_{VGG} = \sum_{i} ||VGG_i(\mathcal{\tilde{I}}) - VGG_i(I_{gt})||_1,
\end{align*}
\noindent where $VGG_i(\cdot)$ indicates extracting the intermediate feature from the $i$th layer of a pretrained VGG network.
The last term $\mathcal{L}_{seg}$ is segmentation loss, 
\begin{align*}
    \mathcal{L}_{seg} = ||\mathcal{M} - M_{gt}||_1,
\end{align*}
\noindent which is the $L_1$ distance between the rendered mask $\mathcal{M}$ and the ground truth segmentation mask $M_{gt}$.

To mitigate overfitting, we perform data augmentation while training.
In order to mimic the noise pattern in the input point cloud, we randomly jitter each point in the point cloud $\bm{q}$ with Gaussian noise.
We find that data augmentation helps stabilize the training.

\section{More Rendering Results}
\begin{figure*}
    \centering
    \includegraphics[width=\textwidth]{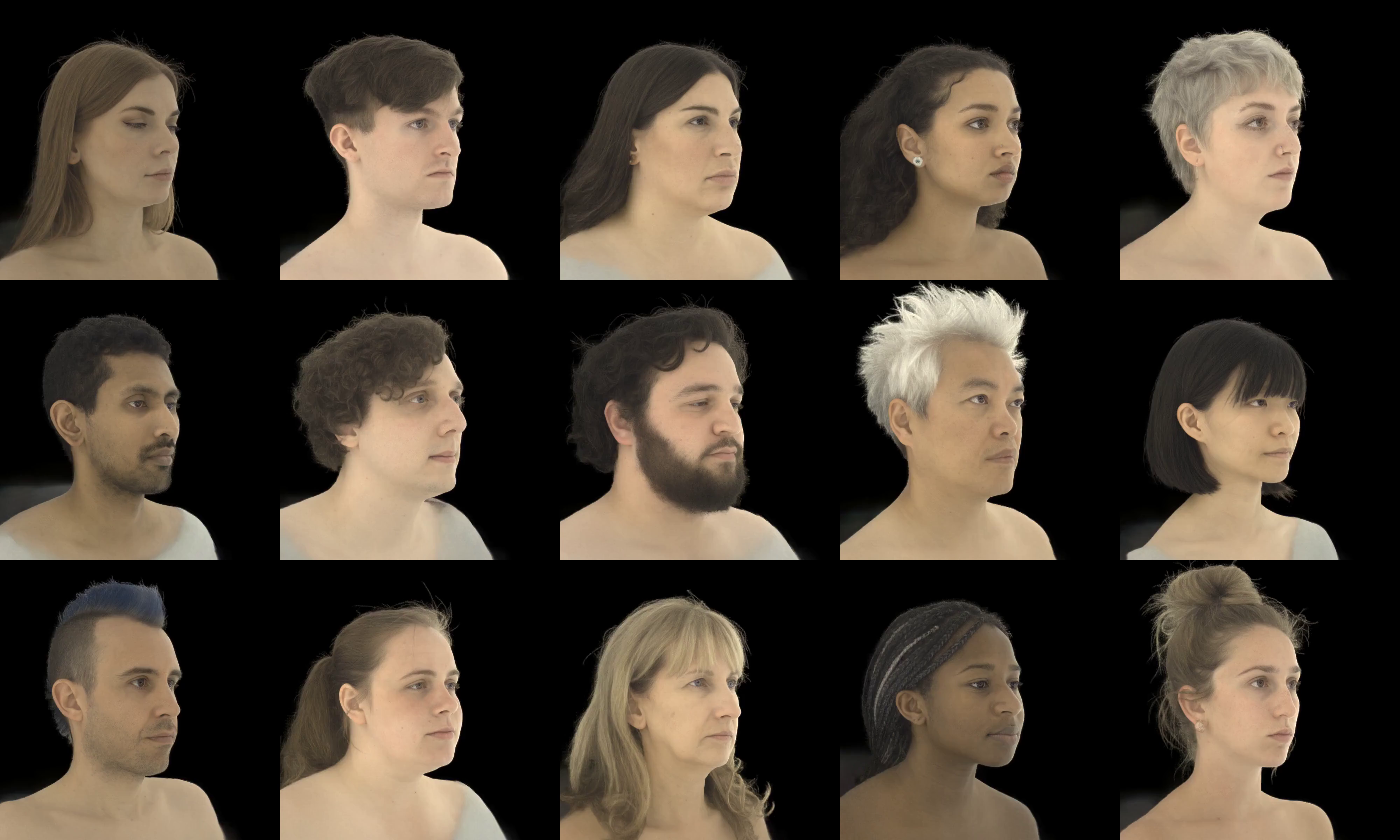}
    \caption{\label{prop_work:supp:trainids}\textbf{Rendering of more hairstyles from the training set.}}
\end{figure*}

\begin{figure*}
    \centering
    \includegraphics[width=\textwidth]{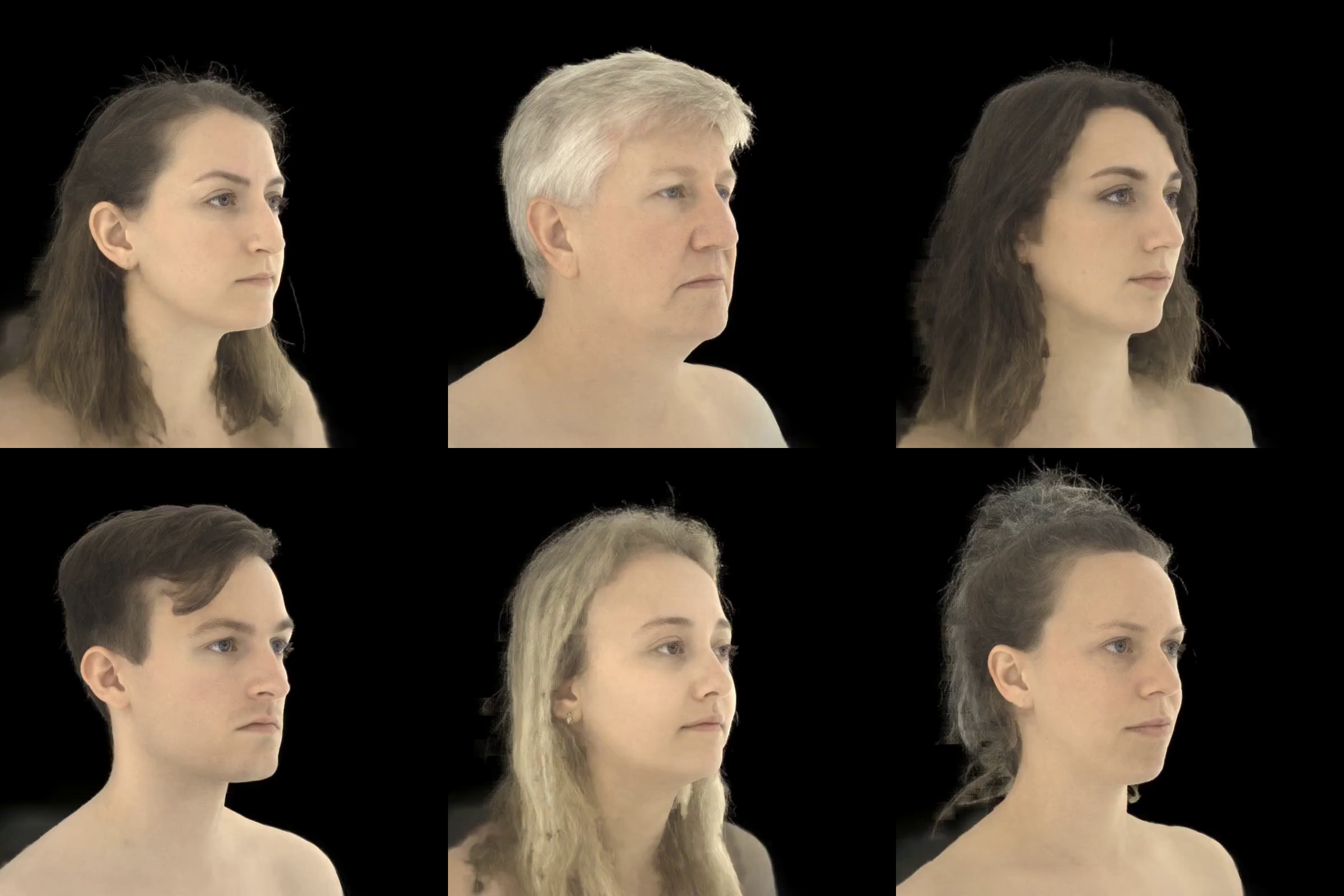}
    \caption{\label{prop_work:supp:testids}\textbf{Rendering of more hairstyles from the test set.}}
\end{figure*}

\begin{figure*}
    \centering
    \includegraphics[width=\textwidth]{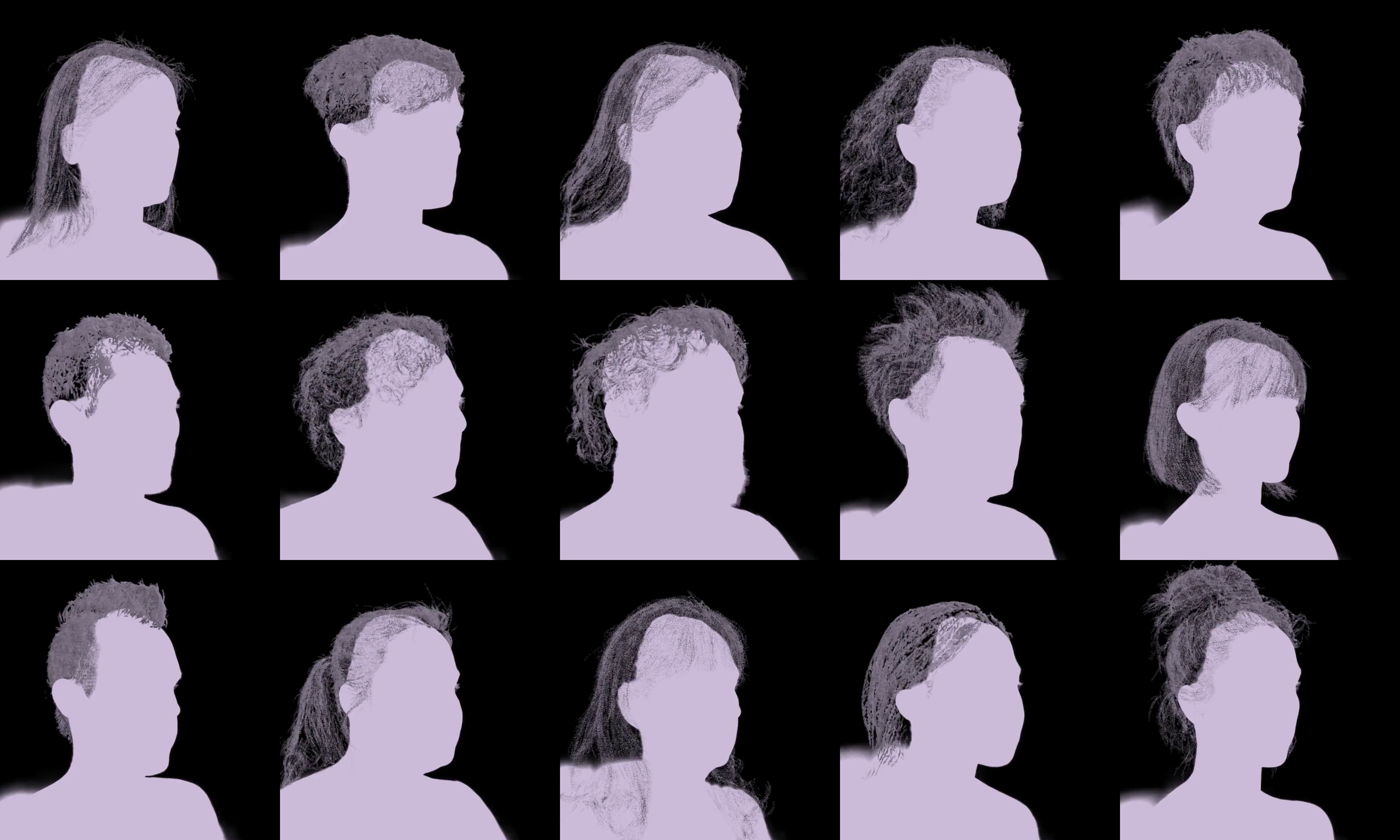}
    \caption{\label{prop_work:supp:trainids_grid}\textbf{Visualization of the sparse point clouds for more hairstyles from the training set.}}
\end{figure*}

\begin{figure*}
    \centering
    \includegraphics[width=\textwidth]{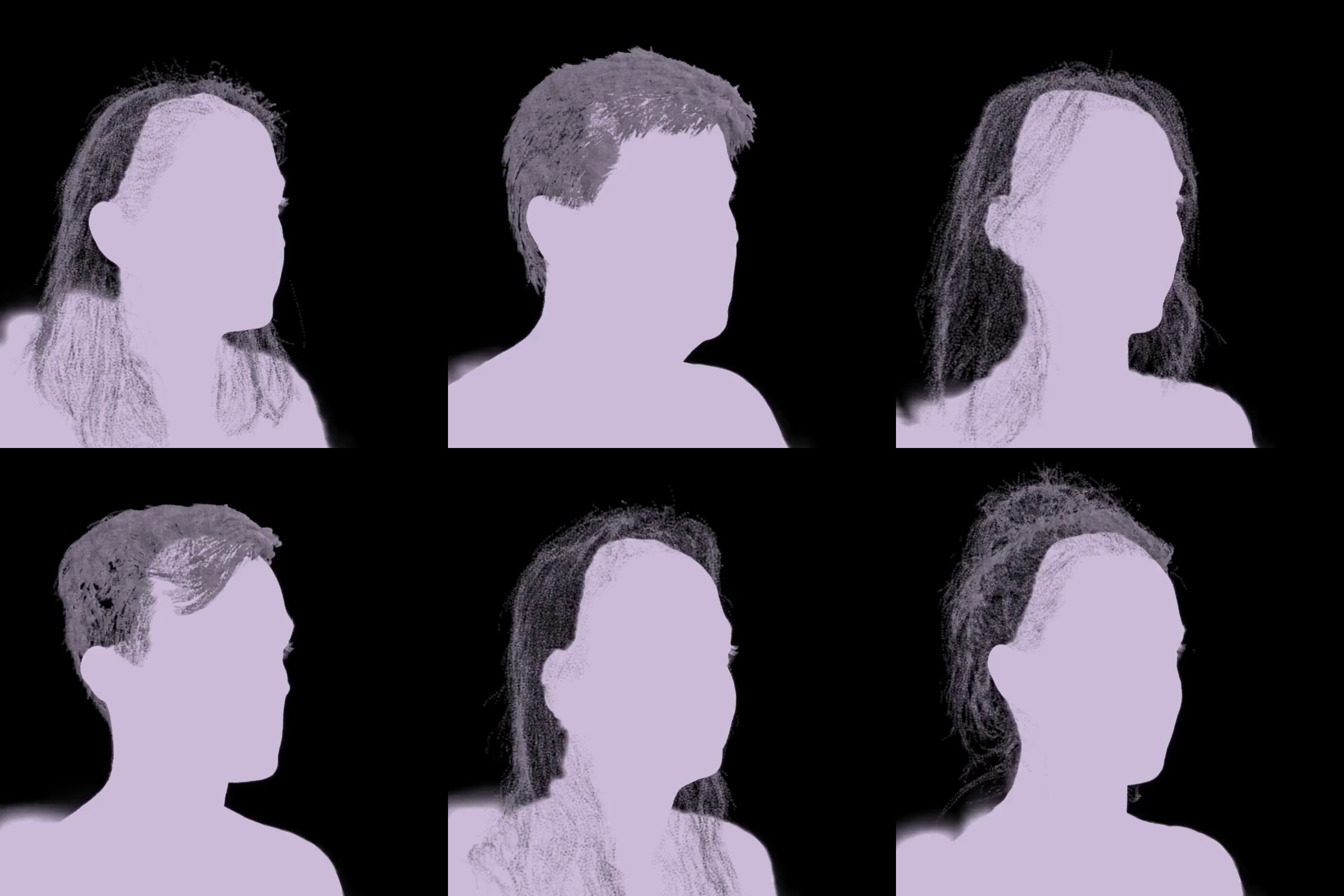}
    \caption{\label{prop_work:supp:testids_grid}\textbf{Visualization of the sparse point clouds for more hairstyles from the test set.}}
\end{figure*}

We show the rendering results in Fig.~\ref{prop_work:supp:trainids} and Fig.~\ref{prop_work:supp:testids} and visualize the input sparse point cloud of more hairstyles in Fig.~\ref{prop_work:supp:trainids_grid} and Fig.~\ref{prop_work:supp:testids_grid}.
As we can see, our method can inpaint the sparse point cloud into a dense radiance field with a photorealistic appearance.
Please refer to the supplemental videos for a free-view rendering of the generated avatars and their corresponding point cloud.

\end{document}